\documentclass{article}

\usepackage[table]{xcolor}

\usepackage{PRIMEarxiv}        % Lexsi Labs arXiv style

\usepackage[T1]{fontenc}
\usepackage[utf8]{inputenc}
\usepackage{microtype}
\usepackage{inconsolata}

\usepackage{amsmath}
\usepackage{amssymb}

\usepackage{booktabs}
\usepackage{multirow}
\usepackage{array}

\usepackage{graphicx}
\usepackage{subcaption}        % loads on top of PRIMEarxiv's caption setup
\graphicspath{{figures/}}
\usepackage[section]{placeins} % keep floats within their section (esp. App. B)

\usepackage{enumitem}
\usepackage{algorithm}
\usepackage{algpseudocode}
\usepackage{url}

\usepackage[numbers]{natbib}

\newcommand{\method}[1]{\textsc{#1}}
\newcommand{\wt}{WikiText-2}

\newcommand{\consensus}{\method{Consensus-2}}
\newcommand{\rev}[1]{#1}
\definecolor{cellgreen}{RGB}{198,239,206}
\definecolor{celllgreen}{RGB}{226,247,229}
\definecolor{cellred}{RGB}{255,199,206}
\definecolor{celllred}{RGB}{255,228,230}

\title{\includegraphics[width=\linewidth]{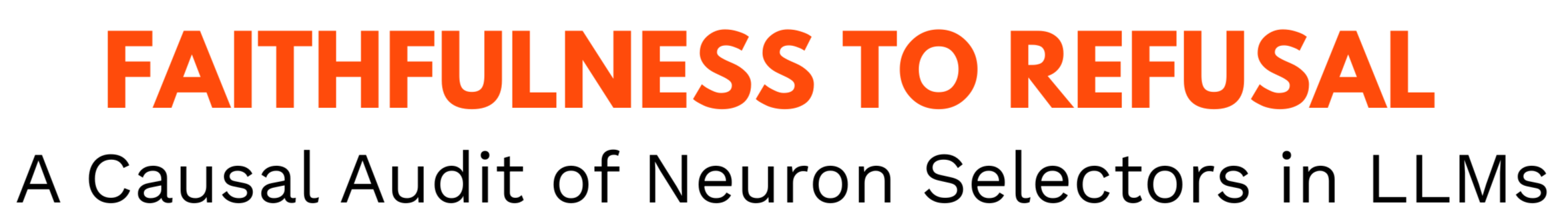}}

\author{
  Ananth Eswar\thanks{Equal contribution; co-first authors.}\hspace{0.6em}\thanks{Work done while at Lexsi Labs.}\hspace{0.1em},
  Pratinav Seth\footnotemark[1], \\
  Utsav Avaiya, Vinay Kumar Sankarapu \\
  \affiliation{Lexsi Labs} \\
  \texttt{pratinav.seth@lexsi.ai}
}

\setabstract{%
\rev{%
Attribution scores increasingly identify which neuron rows of a language model matter for applications such as pruning, interpretability, and editing for safety, yet whether they identify \emph{causally} important rows is rarely tested directly. We address this with two paired audits built on one-shot neuron-row zeroing. We first audit selectors at the language-modeling level: attribution methods substantially
outperform activation- and magnitude-based baselines at identifying dispensable rows across five LLMs. We then adapt the same intervention into a behavior test by driving it with a contrastive harmful-versus-benign signal; the attributed rows are \emph{sufficient} to install refusal on hate and crime while keeping benign over-refusal low and preserving language model fluency, and \emph{specific} in that layer-matched random controls at the same depths fail. Highly rank-stable selectors can be among the least causally valid. Refusal moreover lives in a redundant subspace, where different attribution methods install it through largely disjoint row sets, so the recovered edit is one realization of a sufficient set rather than a unique mechanism.
Together, these findings show that rank-stability proxies miss the kinds of selector failures a direct causal audit can surface.%
}%
}

\setkeywords{neuron attribution, structured pruning, mechanistic interpretability, refusal editing, selector faithfulness, LRP, integrated gradients, LLM safety}

\runningtitle{Faithfulness to Refusal: A Causal Audit of Neuron Selectors}

\begin{document}
\maketitle

% ============================================================
% Abstract is set via \setabstract{} in the preamble and typeset
% inside the PRIMEarxiv title box by \maketitle.
% ============================================================

\section{Introduction}
\label{sec:intro}

Attribution scores now drive three separate workflows: pruning methods
remove parameters by importance score~\cite{sun2024wanda,frantar2023sparsegpt,an2024flap},
mechanistic-interpretability papers identify behavioral circuits via
attribution~\cite{hatefi2025circuit} and safety-editing methods suppress
or install behaviors by ablating attributed
components~\cite{lee2025cast,zou2023representation}. All three rest on
the same \emph{selector-faithfulness assumption}: that the scoring
function (LRP, IG, Wanda, magnitude, or any other) faithfully separates
important from unimportant components. The assumption fails silently.
A pruning recipe wired to an unfaithful selector can still hit a
perplexity (PPL) target after recovery, and a refusal edit can
saturate calibration-set refusal yet generalize unsafely on held-out
red-team prompts, because the standard audit
artifacts (PPL, rank correlation, calibration accuracy) are
blind to the selector's failure mode.

A direct causal test is missing. It would intervene on the selector's
output: zero the rows it ranks dispensable, measure the damage, and do
this under conditions where behavioral change can only come from the
selector, not from surrounding pipeline machinery. The test must also
run at two levels of difficulty, because passing the language-modeling
level (does the selector pick generally important neurons?) does not
entail passing the behavior-specific level (does it pick the neurons
that carry a target behavior?).

\textbf{Approach.}
We use \emph{neuron-row zeroing}: a one-shot, retraining-free ablation
of individual output rows of a transformer's Linear projections (attention
Q/K/V/O or MLP gate/up/down). Unlike DPO, activation steering, or
representation-engineering edits, it adds no fine-tuning step and
no inference-time hooks, so any behavioral change traces to
the rows the selector ranks dispensable rather than to surrounding
pipeline machinery.

At the LM level we run Least-Relevant-First/Most-Relevant-First
(LeRF/MoRF) sweeps for seven selectors (Random, Magnitude, Wanda,
MeanActivation, LRP, IG, and their Borda consensus
\consensus{}) across five base models (LLaMA-3.2-1B/3B,
LLaMA-3.1-8B, Qwen3-8B, Gemma-3-12B). At the behavior level we drive
attribution with a contrastive harmful-versus-benign signal on
matched CAST~\cite{lee2025cast} pairs: single-prompt attribution
would surface prompt-processing neurons rather than refusal-specific
ones; the contrastive margin instead cancels shared content to
isolate the refusal-relevant subspace. We install refusal
on four instruction-tuned models, evaluate on CAST,
SorryBench~\cite{xie2024sorrybench},
OR-Bench-Hard~\cite{cui2024orbench}, and downstream
utility~\cite{hendrycks2021mmlu,cobbe2021gsm8k,zhou2023ifeval}, with
a layer-matched random control to distinguish row identity from
depth allocation.

A central finding cuts across both levels: the most rank-stable
selector is the least causally faithful, and at the behavior level it
even decreases refusal, so rank-stability proxies systematically
miss the failures a direct causal audit can surface.

\textbf{Contributions.}
\begin{enumerate}[nosep,leftmargin=*,label=\textbf{(\arabic*)}]
  \item A causal selector-audit framework adapted from LeRF/MoRF,
    applied to internal neuron rows at two difficulty levels via a
    single one-shot intervention requiring no fine-tuning or
    inference-time modification.
  \item Evidence that attribution-based selectors are causally faithful at the LM
    level: LRP, IG, and \consensus{} identify dispensable neurons
    two to four orders of magnitude more accurately than
    activation, magnitude, and random baselines across
    models, \rev{together with a control suite showing that the rows
    where LRP and IG \emph{disagree} carry stronger causal signal
    than the agreed set, motivating disagreement-aware aggregation
    as a direction for follow-up work}.
  \item Evidence that contrastive attribution installs refusal at the
    behavior level: refusal saturates on hate and crime while
    preserving benign accuracy and language fluency, the effect
    holds on held-out red-team benchmarks, and a layer-matched
    random control fails, locating the effect at row identity
    rather than depth allocation.
  \item Evidence that the underlying subspace is redundant and
    architecture-dependent: LRP and IG share only a small fraction
    of their top rows yet both install refusal, and the
    best-performing attribution method reverses across LLaMA and
    Qwen.
\end{enumerate}

\section{Related Work}
\label{sec:related}

\subsection{Attribution in LLMs: Methods and Faithfulness}
LLM pruning spans a cost-quality spectrum from weight-magnitude and
activation-aware criteria (Wanda~\cite{sun2024wanda},
SparseGPT~\cite{frantar2023sparsegpt}, FLAP~\cite{an2024flap},
LLM-Pruner~\cite{ma2023llmpruner}) to attribution-based criteria
(row-level LRP~\cite{hatefi2024pruning} and model-agnostic relevance
attribution~\cite{sankarapu2024dlbacktrace}). Mechanistic interpretability
identifies circuits via activation patching~\cite{wang2023ioi} and
attribution-based alternatives~\cite{hatefi2025circuit,syed2023attribution}. These approaches all share the
same selector-faithfulness assumption, a premise that becomes more
consequential as interpretability is promoted from a post-hoc
diagnostic to a substrate for alignment~\cite{sengupta2025interpretability}. The
LeRF/MoRF protocol~\cite{samek2017evaluating} and
ROAR~\cite{hooker2019benchmark} test this for input-feature
attributors on classification tasks, and explanation-evaluation
frameworks benchmark the faithfulness, sensitivity, and robustness of
post-hoc attributions~\cite{seth2025xaievals,seth2025xaimetrics}. We adapt the same causal logic to internal neuron rows of autoregressive LLMs,
comparing LRP~\cite{bach2015pixel,montavon2019overview} in its
attention-aware variant AttnLRP~\cite{achtibat2024attnlrp},
IG~\cite{sundararajan2017axiomatic}, their Borda consensus, and
activation/magnitude baselines across five models and two difficulty
levels.

\subsection{Behavior Installation and Evaluation}
Refusal behaviors are conventionally installed via
RLHF~\cite{bai2022hhrlhf} or DPO~\cite{rafailov2023dpo}, updating
most parameters. Inference-time alternatives (contrastive activation
addition~\cite{rimsky2024caa}, CAST~\cite{lee2025cast}, and
representation engineering~\cite{zou2023representation}) steer
hidden states without weight changes but validate only by exhibiting
the steering effect, not by ruling out that any sparse intervention
would suffice. Closer to the weight level, refusal has been traced to
a single residual-stream direction whose rank-one ablation removes
it~\cite{arditi2024refusal}, and pruning a small set of
safety-critical weights can likewise strip
alignment~\cite{wei2024assessing}; both show safety behavior is
sparsely localized. \rev{Concurrent work installs category-specific
refusal directly through circuit-restricted weight
arithmetic~\cite{kasliwal2026ctheta}; our audit is complementary,
asking which selector most faithfully recovers such rows rather than
proposing a single editing recipe. A parallel line localizes safety to sparse
neuron sets in activation space directly: intervening on only ${\sim}5\%$ of neurons,
identified via inference-time activation contrasting and dynamic
activation patching, restores most safety
behavior~\cite{chen2024safetyneurons}, and sparse-autoencoder studies
benchmark feature-selection heuristics (cosine similarity, activation
difference, attribution patching) for recovering causal refusal
features~\cite{yeo2025refusalsae,prakash2025dissecting}. The degree
of sparsity is itself contested: refusal has been cast as a single
residual-stream direction~\cite{arditi2024refusal}, as multiple
functionally independent directions~\cite{wollschlager2025geometry},
and as reducible to a single sufficient MLP
neuron~\cite{kazemi2026singleneuron}. What this line does not do is
compare \emph{selectors} in weight-row space: it asks where refusal
lives, and (for the SAE studies) which feature-selection heuristic
recovers it in activation space, not which weight-row importance
score most faithfully recovers the responsible components under a
common causal-validity metric with layer-matched controls. Our audit
differs on four axes: (i) it compares seven selectors head-to-head
against one validity metric rather than validating a single bespoke
localizer; (ii) it runs layer-matched, rank-randomized, and veto
controls that separate row identity from depth allocation and rank
averaging; (iii) it operates in weight-row space and leaves a
standard checkpoint, so behavioral change is attributable to the
\emph{identity} of the removed rows rather than to an inference-time
hook; and (iv) it needs no aligned/unaligned reference-model pair,
computing importance one-shot from a contrastive signal.} We evaluate LM fluency with
\wt{} perplexity~\cite{merity2017wikitext}, downstream utility via
lm-eval-harness~\cite{eval-harness} on MMLU~\cite{hendrycks2021mmlu},
GSM8K~\cite{cobbe2021gsm8k}, and IFEval~\cite{zhou2023ifeval}, and
red-team transfer with SorryBench~\cite{xie2024sorrybench} and
OR-Bench-Hard~\cite{cui2024orbench}; together these benchmarks separate under-refusal from over-refusal.

% ============================================================
\section{Audit Framework}
\label{sec:method}

\subsection{Intervention: Neuron-Row Masking}
\label{sec:intervention}

Each Linear layer in a transformer block computes $\mathbf{y} =
\mathbf{W}\mathbf{x} + \mathbf{b}$, where $\mathbf{W} \in
\mathbb{R}^{d_\text{out} \times d_\text{in}}$. A \emph{neuron-row}
is one row of $\mathbf{W}$: the weights producing a single output
dimension. \rev{We use \emph{row} (equivalently \emph{neuron row})
throughout; an unqualified ``neuron'' always denotes such a
weight-matrix output row, not an activation unit as in the
safety-neuron literature.} Masking neuron-row $i$ sets $\mathbf{W}[i,:] = \mathbf{0}$
and $\mathbf{b}[i] = 0$, silencing that output dimension. This is the
minimal structural intervention at the weight level: one output
dimension is removed, the rest of the architecture is untouched,
no fine-tuning or recovery is needed, and the result is a standard
checkpoint that loads and runs without modification.

A transformer block contains multiple Linear layers: attention
projections (Q/K/V/O) and MLP projections (gate/up/down) across all
five models. All neuron rows across all blocks are pooled into a
single global ranking. Given a selector's importance scores we mask
the globally \emph{lowest-importance} $k\%$ (LeRF,
least-relevant-first) or the globally \emph{highest-importance}
$k\%$ (MoRF, most-relevant-first), and measure the change in model
behavior: \wt{} perplexity and downstream accuracy at the LM level
(\S\ref{sec:general}), refusal rate paired with benign accuracy and
utility retention at the behavior level (\S\ref{sec:contrastive}).
Embedding layers and the language-model head are excluded because
their rows are vocab-indexed, so masking them destroys token
identity rather than shared computation.

\subsection{Selectors and Validity Gap}
\label{sec:selectors}

We evaluate seven selectors against a single validity metric (full
descriptions and cost table in
Appendix~\ref{app:selectors_detail}). Four are non-attribution
nulls/baselines: \method{Random} (uniform scores, 3-seed average),
\method{Magnitude} (mean absolute row weight, data-free),
\method{Wanda} ($s_i = \sum_j |W_{ij}|\cdot\text{RMS}(X_j)$, our
row-aggregated adaptation of the per-weight criterion of
\cite{sun2024wanda}), and \method{MeanActivation} (mean absolute
output activation over calibration data). Three are attribution-based:
\method{LRP} (AttnLRP \cite{achtibat2024attnlrp} via the
\texttt{lxt} library), \method{IG} (Integrated Gradients
\cite{sundararajan2017axiomatic}, 16-step), and \consensus{}, a
Borda-style rank aggregation of LRP and IG scores
(full formula in Appendix~\ref{app:selectors_detail}). \consensus{} is motivated by error
reduction: LRP propagates relevance layer by layer while IG
integrates the gradient along a path, so the two methods are likely
to fail in different places.

A faithful selector should damage the model little on the rows it
ranks safe (low LeRF PPL) and damage it severely on the rows it
ranks important (high MoRF PPL). We report both quantities and
their \emph{validity gap}, $\text{Gap}(S, k) =
\text{PPL}_\text{MoRF}(S, k) - \text{PPL}_\text{LeRF}(S, k)$;
a random selector produces a gap near zero, while a faithful
selector produces a positive gap whose magnitude scales with the
asymmetry between the two halves. Section~\ref{sec:contrastive} reuses
this intervention at the behavior level by driving the same
selectors with a contrastive harmful-versus-benign signal.

\rev{\textbf{What ``faithful'' means here.}
We call a selector \emph{faithful} in a purely causal, operational
sense: its top-ranked rows are (i) \emph{sufficient}, meaning that
zeroing them produces the intended effect (a large validity gap at
the LM level, installed refusal at the behavior level); and (ii)
\emph{specific}, meaning that the effect is not reproduced by a
layer-matched control that zeroes the same per-layer row counts with
random row identities. We make no claim of \emph{necessity} or
\emph{uniqueness}: because the responsible subspace is redundant
(\S\ref{sec:redundant}), many disjoint row sets are sufficient, and a
faithful selector is one that reliably lands in this sufficient set,
not one that recovers a canonical mechanism. The distinction matters
for how the audit is used. Sufficiency and specificity are what
attribution-guided pruning and safety editing actually require, and
the audit certifies them directly; but circuit-discovery claims that
read a selector's top rows as \emph{the} mechanism need the stronger
necessity property this audit deliberately does not test.}

\subsection{Distribution shift does not explain the audit's signal}
\label{sec:shift}
\rev{Deletion-based faithfulness metrics carry a known confound: removing
components can push a network into states it was never trained on, so
a measured degradation may reflect distribution shift rather than the
removed component's importance, the concern that motivated the
retrain-after-removal protocols of ROAR and Recursive
ROAR~\cite{hooker2019benchmark,madsen2021recursiveroar}. Zeroing
weight rows is a strong perturbation, so the objection applies in
principle. Four features of the design neutralize it. First, the
validity gap is a \emph{differencing} measurement: LeRF and MoRF
apply the same intervention class (equal-sized row zeroing) at the
same sparsity to the same model, so any generic ablation shock is
common to both arms and cancels in
$\text{Gap} = \text{PPL}_\text{MoRF} - \text{PPL}_\text{LeRF}$; only
row-identity-dependent damage survives. Second, \method{Random} is an
empirical shift floor: zeroing arbitrary rows at each rate produces a
near-zero or negative gap (\S\ref{sec:general}), so the intervention
\emph{per se} does not manufacture the asymmetry; it appears
only when the selector carries signal. Third, at the behavior level
the layer-matched controls (\S\ref{sec:exp_b}) hold the
intervention's shape fixed (identical per-layer row counts) and
still collapse to near-baseline refusal, isolating row identity from
any depth-dependent shock profile. Fourth, refusal installation is
certified on \emph{preserved} benign accuracy, downstream utility,
and perplexity, not on degradation alone, so a positive result cannot
be an artifact of the model merely being broken. The audit therefore
measures selector-attributable causal effect, not ablation shock.}

\subsection{Models, Data, and Implementation}
\label{sec:setup}

\textbf{Models.}
The audit covers five open-weight base models across three
architecture families: three LLaMA-3 sizes
(LLaMA-3.2-1B, LLaMA-3.2-3B, LLaMA-3.1-8B)~\cite{dubey2024llama3},
Qwen3-8B~\cite{qwen3technicalreport}, and
Gemma-3-12B~\cite{team2025gemma3}. All five
enter the LeRF/MoRF audit. Stability-validity tests, depth profiles
(\S\ref{sec:stability}), consensus controls and the rank-randomized
null (\S\ref{sec:consensus}) run on LLaMA-3.1-8B and
Qwen3-8B. Jaccard agreement (\S\ref{sec:structural}) covers
LLaMA-3.2-1B, LLaMA-3.1-8B, and Qwen3-8B. Domain sensitivity
(Appendix~\ref{app:domain_full}) is reported on
LLaMA-3.1-8B only. Per-model block, layer, and neuron counts are in
Appendix~\ref{app:selectors_detail} (Table~\ref{tab:models}).

\textbf{Calibration and evaluation.}
Importance is computed on 128 WikiText-2 training samples (max
length 128 tokens) for the data-dependent selectors
(\method{MeanActivation}, \method{Wanda}, \method{LRP}, \method{IG},
\consensus{}); \method{Magnitude} is data-free, and \method{Random}
averages over 3 seeds. The stability-validity sweep uses nested
subsets $C_8 \subset C_{16} \subset C_{32} \cdots \subset C_{128}$ from
the same 128-sample pool, so subset growth changes sample
count rather than sample identity. For domain sensitivity
(LLaMA-3.1-8B) we additionally calibrate on 128 samples from
C4~\cite{raffel2020c4}.
Perplexity is on 256 WikiText-2 test samples
\cite{merity2017wikitext} (sequence length 512); downstream accuracy uses 8 zero-shot
tasks from \texttt{lm-eval-harness} \cite{eval-harness} (ARC-E/C,
HellaSwag, WinoGrande, PIQA, BoolQ, LAMBADA-OpenAI, OpenBookQA).

\textbf{Sweep and implementation.}
The general audit sweeps $0$-$90\%$ in $5\%$ steps for PPL
(19 rates) and $10\%$ steps for downstream (10 rates). All runs use a
single A100~(80\,GB); calibration is $< 15$ GPU-min per model, and
the full sweep (all selectors, all masking rates, PPL and downstream)
takes approximately $3$-$9$ hours per model depending on model size
and selector cost.

% ============================================================
\section{General Selector Audit}
\label{sec:general}

\subsection{Attribution wins on language modeling.}
LRP, IG, and \consensus{} sit far below
non-attribution baselines on LeRF at moderate masking rates,
including the $30\%$ operating point, across all five models
(Figure~\ref{fig:lerf_all}; full per-rate tables in
Appendix~\ref{app:full_lerf}). MoRF shows the matching asymmetry:
when the same selectors target the rows they rank \emph{most}
important, the LM breaks faster (LRP reaches $8.3 \times 10^6$ PPL
at $30\%$). The pattern carries over to downstream utility: on the
larger models, attribution selectors retain substantial mean
\texttt{lm-eval-harness} accuracy at $30\%$ LeRF
($\consensus{}=0.503$ on Gemma-3-12B, $0.392$ on LLaMA-3.1-8B),
while non-attribution baselines collapse to the random floor near
$0.28$ (Table~\ref{tab:audit_summary_full}, column A). Per-model
numerics, validity gap, and mean downstream accuracy are in
Appendix~\ref{app:audit_summary_full}; per-task downstream breakdowns
are in Appendix~\ref{app:full_lerf}.

\begin{figure}[pt]
  \centering
  \includegraphics[width=\textwidth]{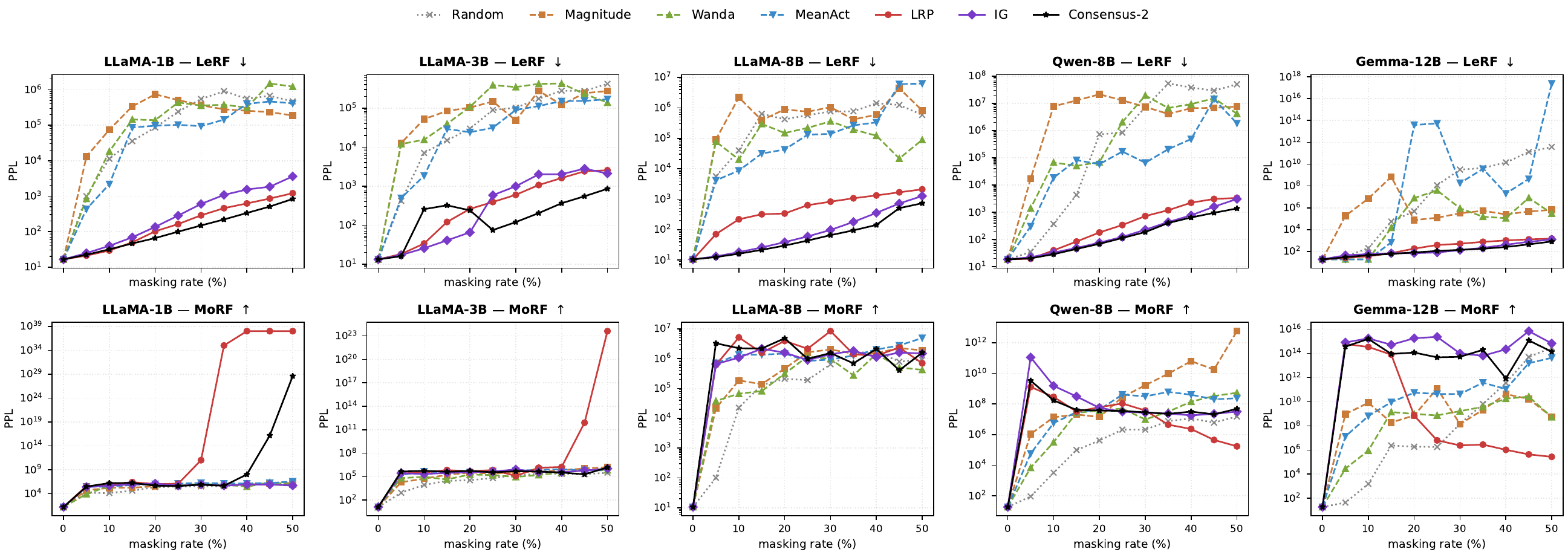}
  \caption{LeRF (lower is better) and MoRF (higher is better) degradation
    curves for all five models and seven selectors (0-50\%, log $y$-axis).
    Attribution selectors (red/black/purple) generally sit below
      non-attribution baselines on LeRF, while MoRF exposes architecture-
      and selector-dependent asymmetries quantified in
    Table~\ref{tab:audit_summary_full}.
    Full 19-rate tables: Appendix~\ref{app:full_lerf}.}
  \label{fig:lerf_all}
\end{figure}

\subsection{Rank stability does not imply causal validity.}
\label{sec:stability}
A common low-cost proxy is rank stability: if a selector
ranks rows similarly across calibration subsets, it is assumed
reliable. We compute importance at
$C \in \{8, 16, 32, 64, 128\}$ and measure stability (Spearman
$\rho$, Jaccard vs.\ $C_{128}$) alongside validity (LeRF PPL).

MeanActivation has near-perfect rank stability and catastrophic
LeRF PPL: the ranking is reproducible across calibration sizes but
does not track causal importance. LRP is the least stable selector
at small $C$ yet produces the largest validity gap; \consensus{} is
the strongest on LeRF
(Table~\ref{tab:stability}). Stability is therefore neither necessary nor sufficient: a selector can be perfectly reproducible and still
rank causally important rows safe. The dissociation replicates on
Qwen3-8B (Appendix Figure~\ref{fig:scatter}).

\begin{table}[pt]
\centering
\footnotesize
\caption{Stability vs.\ validity numerics on the LLaMA-3.1-8B
  reference model. Stability: Spearman $\rho$ and Jaccard$@30\%$ of
  C8 scores/masks vs.\ C128. Validity: LeRF, MoRF, and Gap at
  $30\%$ masking using full C128 scores. Qwen3-8B replication in
  Appendix~\ref{app:stability_full}.}
\label{tab:stability}
\begin{tabular}{l rr rrr}
\toprule
& \multicolumn{2}{c}{Stability (C8$\to$C128)} & \multicolumn{3}{c}{Validity (C128, @30\%)} \\
\cmidrule(lr){2-3}\cmidrule(lr){4-6}
Selector & Spearman & Jacc & LeRF$\downarrow$ & MoRF$\uparrow$ & Gap$\uparrow$ \\
\midrule
\method{MeanAct.}  & .994 & .934 & \cellcolor{celllred} 139{,}874 & \cellcolor{cellred} 930{,}749 & \cellcolor{celllred} 790{,}875 \\ 
\method{Wanda}     & .9998 & .981 & \cellcolor{cellred} 360{,}083 & \cellcolor{celllred} 976{,}078 & \cellcolor{cellred} 615{,}995 \\ 
\method{IG}        & .965 & .756 & \cellcolor{celllgreen} 97.8 & \cellcolor{celllgreen} 1.40e6 & \cellcolor{celllgreen} 1.40e6 \\ 
\method{LRP}       & .913 & .635 & \cellcolor{celllgreen} 830.4 & \cellcolor{cellgreen} 8.32e6 & \cellcolor{cellgreen} 8.32e6 \\ 
\consensus{}       & .945 & .727 & \cellcolor{cellgreen} 66.1 & \cellcolor{celllgreen} 1.54e6 & \cellcolor{celllgreen} 1.54e6 \\ 
\bottomrule
\end{tabular}
\end{table}

\subsection{Consensus needs real cross-method agreement.}
\label{sec:consensus}
The Borda aggregate of LRP and IG produces the lowest LeRF PPL on
four of five models and matches IG on the fifth
(Appendix Figure~\ref{fig:audit_summary}). Four control variants
rule out trivial explanations on LLaMA-3.1-8B and Qwen3-8B
(Table~\ref{tab:consensus_8b}, Appendix
Figure~\ref{fig:consensus_controls}). Layer-matched controls that
copy each method's depth distribution but randomize rows within
layers reach LeRF $459$ (LRP) and $1{,}301$ (IG) on LLaMA-8B against
\consensus{}'s $66.1$, so depth allocation alone does not reproduce
the win. A rank-randomized null that permutes IG ranks before
averaging reaches LeRF $119$, better than LRP alone ($830$) but
worse than real \consensus{} ($66.1$), so the safety advantage is
not produced by the act of averaging two rank lists. A strict
LRP$\cap$IG intersection (rows ranked in the bottom-$k\%$ by
\emph{both} methods) delivers the lowest LeRF in the entire
audit ($10.7$ on LLaMA-8B, $20.9$ on Qwen3-8B), at a small MoRF cost
on LLaMA ($1.39 \times 10^6$ vs.\ $1.54 \times 10^6$) and a
catastrophic one on Qwen ($1.59 \times 10^5$ vs.\ $2.83 \times
10^7$): agreement is a strong but architecture-dependent safety
signal. Finally, \emph{veto} aggregators take rows in one method's
top-$k$ but \emph{not} the other's, testing whether the rows where
the two methods disagree carry causal signal. \method{VETO-LRP}
(LRP's exclusive picks) beats real \consensus{} on both axes on
both models (LLaMA-8B: LeRF $28.0$ / MoRF $3.32 \times
10^6$; Qwen3-8B: LeRF $63.1$ / MoRF $5.89 \times 10^7$), pointing
to disagreement-aware aggregation as a direction for follow-up work.
Full controls in Appendix~\ref{app:consensus_full}.

\subsection{Where selectors agree, and where they don't.}
\label{sec:structural}
At $30\%$ masking, attribution methods (LRP, IG, \consensus{}) agree
with one another at consistent levels across attention and MLP
sublayers (mean pairwise Jaccard $\approx 0.63$-$0.68$ on both,
on LLaMA-3.1-8B and Qwen3-8B). Non-attribution baselines (Magnitude,
Wanda, MeanActivation) cluster strongly on attention but weakly on
MLP on LLaMA-3.1-8B ($\approx 0.72$ vs.\ $0.44$) and only weakly on
both on Qwen3-8B ($\approx 0.49$ vs.\ $0.37$): they pick similar
rows when an obvious weight statistic (attention magnitude, on
LLaMA) drives the ranking, and diverge once that crutch is gone.
Attribution extracts a more consistent signal across both
sublayers and both architectures
(Appendix Figure~\ref{fig:jaccard}). Per-block depth profiles
(Appendix Figure~\ref{fig:depth_profile}) show \method{LRP},
\method{IG}, and \consensus{} concentrating in early-to-mid layers
while \method{Magnitude} and \method{Wanda} climb toward late
layers; this asymmetry is why the layer-matched control is needed
(Table~\ref{tab:consensus_8b}). A false-negative analysis and
a WikiText-vs-C4 corpus check control for depth allocation and
calibration corpus as confounds (Appendix~\ref{app:depth_full} and Appendix~\ref{app:domain_full}).

% ============================================================

\section{Contrastive Refusal Editing}
\label{sec:contrastive}

At the behavior level, the same row-zeroing intervention must pass a
harder test: a faithful selector should identify rows that carry a
specific behavior (safety refusal), not just generally important rows.
Zeroing them should install refusal (sufficiency), and the effect
should not be reproducible by zeroing random rows at the same depths
(specificity). We define the contrastive mask construction
(§\ref{sec:contrastive_method}) and run the test
(§\ref{sec:contrastive_setup} onward) across four instruction-tuned
models, five selectors, and five CAST harm domains. The full grid is $70$ (model, selector, domain) cells.

\subsection{Contrastive Refusal Masks}
\label{sec:contrastive_method}

The intervention is the row-zeroing of §\ref{sec:intervention}. The
target changes: instead of the next-token loss we use a refusal
decision margin at the last prompt position, contrasted across
matched harmful and benign CAST pairs. The margin $m(x) =
\text{mean}_{t\in\mathcal{S}}\,z_t - \text{mean}_{t\in\mathcal{H}}\,z_t$
uses two fixed token sets: $\mathcal{S}$ contains the first tokens of
refusal-onset phrases (e.g., ``I must decline'', ``Unfortunately'')
and $\mathcal{H}$ contains the first tokens of compliance-onset
phrases (e.g., ``Sure'', ``Absolutely''), both drawn from a curated
list of $\le 40$ variants per set tokenized with the model's own
tokenizer. Algorithm~\ref{alg:contrastive}
gives the full construction.

\begin{algorithm}[t]
\caption{Contrastive Refusal Mask Construction}
\label{alg:contrastive}
\footnotesize
\begin{algorithmic}[1]
\Require
  \Statex \hspace{1em} $M$: model under audit
  \Statex \hspace{1em} $D = \{(x^{\text{risk}}, x^{\text{base}})\}$: matched CAST harmful/benign pairs
  \Statex \hspace{1em} $A \in \{\text{LRP},\,\text{IG},\,\text{Consensus-2}\}$: attribution method
  \Statex \hspace{1em} $k$: mask sparsity; \quad $\lambda$: LM-protect strength
  \Statex \hspace{1em} $I_\text{lm}$: per-row LM importance from WikiText next-token attribution
\Ensure top-$k$ refusal mask (row indices to zero)
\Statex
\Comment{\textit{Step 1: compute refusal margin and per-row relevance}}
\For{$x \in D^{\text{risk}} \cup D^{\text{base}}$}
  \State $z_t(x) \gets$ logit of token $t$ at last prompt position
  \State $m(x) \gets \tfrac{1}{|\mathcal{S}|}\!\sum_{t\in\mathcal{S}} z_t(x) - \tfrac{1}{|\mathcal{H}|}\!\sum_{t\in\mathcal{H}} z_t(x)$
    \hfill\Comment{$\mathcal{S}$: refusal tokens; $\mathcal{H}$: compliance tokens}
  \State $R_i(x) \gets A(M,\,x,\,m)$
    \hfill\Comment{signed relevance of row $i$ to $m(x)$}
\EndFor
\Statex
\Comment{\textit{Step 2: aggregate compliance-promoting importance and rank-normalize}}
\For{$D' \in \{D^{\text{risk}},\,D^{\text{base}}\}$}
  \State $I_i^{-,D'} \gets \mathbb{E}_{x \in D'}\!\left[\max\!\left(-R_i(x),\,0\right)\right]$
    \hfill\Comment{compliance-promoting half of relevance}
  \State $\hat{I}_i^{-,D'} \gets r_\ell\!\left(I_i^{-,D'}\right)$
    \hfill\Comment{per-layer percentile rank, $r_\ell(\cdot)\in[0,1]$}
\EndFor
\Statex
\Comment{\textit{Step 3: contrastive score and LM-protect penalty}}
\State $C_\text{comply}[i] \gets \hat{I}_i^{-,\text{risk}} - \hat{I}_i^{-,\text{base}}$
  \hfill\Comment{harm-specific compliance signal, $\in[-1,1]$}
\State $S[i] \gets C_\text{comply}[i] - \lambda\cdot r_\ell\!\left(I_\text{lm}[i]\right)$
  \hfill\Comment{down-weight LM-critical rows}
\State \Return $\operatorname{top-}k(S)$
\end{algorithmic}
\end{algorithm}

Non-attribution variants (\method{MeanActivation} and
\method{Random}) bypass Steps 1-2 and substitute a different score
at Step 3: \method{MeanActivation} uses unsigned activation
magnitude and \method{Random} uses seeded per-layer uniform noise.
\consensus{} is an attribution variant: it runs Steps 1-2
independently for LRP and IG, then Borda-aggregates their per-layer
ranks into a single score.

\textbf{Operating-point selection.}
The sweep is $\lambda \in \{0.2, 0.3, 0.5, 0.6, 0.7, 0.8\}$ and
$k \in \{0.005, 0.01, 0.02, 0.05, 0.075, 0.1\}$, giving $36$
$(\lambda, k)$ cells per attribution run. Two extra $k$ values
($0.015$, $0.15$) are added for LLaMA-3.2-1B medical and legal,
where the standard $k$ values find no feasible cell. The operating
point per (model, selector, domain) is
$(\lambda^\star, k^\star) =
\arg\max_{(\lambda,k)\in\mathcal{V}} \text{CAST-malign}(\lambda,k)$
with feasible set $\mathcal{V} = \{(\lambda,k) : \text{benign} \le
0.10,\ \text{PPL} \le 100\}$. If $\mathcal{V} = \emptyset$, the
benign cap is relaxed in $0.05$ steps until a feasible cell with
non-trivial refusal exists; the relaxed cell is flagged
$^\star$. This fires once in the $70$-cell grid (LLaMA-3.1-8B,
IG, adult). A selector passes the test if its top-$k$ rows raise
refusal on harmful prompts, preserve benign compliance, preserve
general LM capability, and beat a layer-matched random control that
zeroes the same per-layer counts with random row identities.
Of the $70$ cells in the grid, $69$ find a feasible operating
point at the default benign cap; only LLaMA-3.1-8B IG-adult needs
the rescue relaxation (App.~\ref{app:rescue}).
\rev{The reported rates are $\arg\max$ selections over the $36$-cell
$(\lambda,k)$ grid, not isolated spikes: the per-cell sweeps
(App.~\ref{app:lambda_sweep}, Table~\ref{tab:lambda_sweep_8b_lrp_hate})
show neighbouring feasible cells landing within $\sim0.1$-$0.2$ of the
reported optimum.}

\subsection{Audit Setup}
\label{sec:contrastive_setup}

Four instruction-tuned models (the three LLaMA-3 sizes and
Qwen3-8B) and five CAST~\cite{lee2025cast} harm domains form the
grid; LRP, IG, and Random run on all four models, while
\consensus{} and MeanActivation run on LLaMA-3.1-8B only, giving
$70$ cells in total.Calibration uses $64$ matched risk/base pairs
  from CAST's training split; refusal is judged by
  \texttt{ProtectAI/\allowbreak distilroberta-base-rejection-v1}~\cite{protectai2023classifier}
  on CAST's held-out test split: $500$ risk prompts and $500$ matched base prompts. Held-out OOD and utility use
the five benchmarks named in §\ref{sec:intro}; LM fluency is
WikiText-2 PPL as in §\ref{sec:general}. All runs use one A100
(80\,GB); total compute is about six GPU-days. The $70$ edited
checkpoints will be released as standard HuggingFace artifacts upon
publication.

\subsection{Attribution installs refusal; the nulls fail in opposite
directions.}
\label{sec:headline}
\begin{figure}[t]
  \centering
  \includegraphics[width=\textwidth]{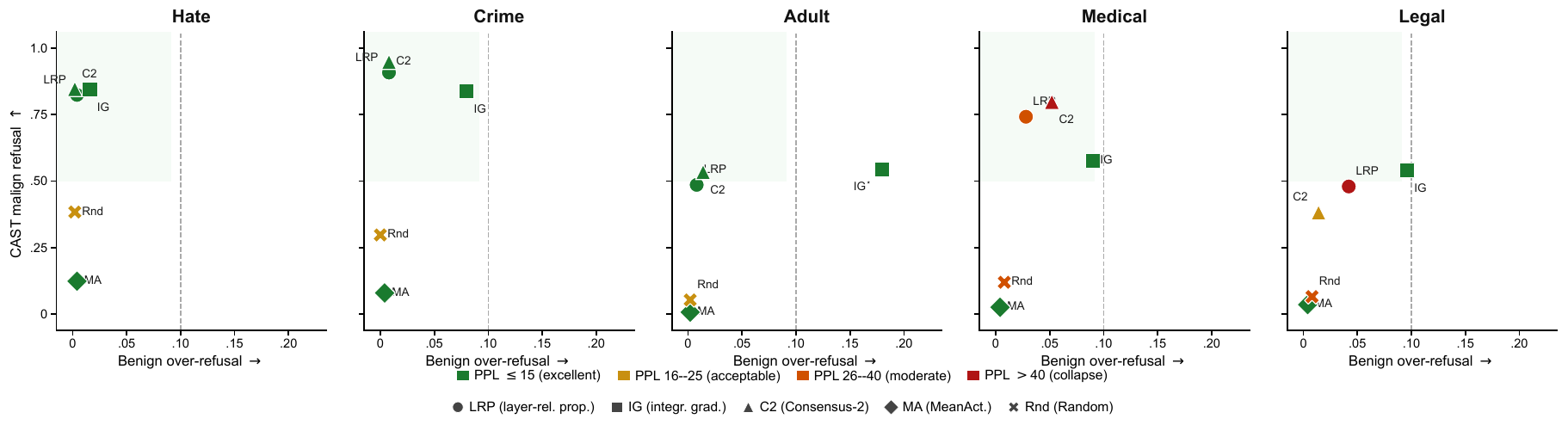}
  \caption{Contrastive refusal editing on LLaMA-3.1-8B-Instruct: one panel per
    CAST harm domain. $x$ = benign over-refusal; $y$ = CAST malign refusal.
    Marker shape encodes selector (see legend); fill color encodes WikiText
    PPL at the operating point. Dashed line marks the benign cap ($0.10$);
    green shading marks the low-benign, high-malign region.
    Attribution selectors (LRP, IG, C2) cluster top-left in hate/crime, with
    high malign, low benign, and excellent PPL. Medical/Legal: LRP and C2 show orange/red
    PPL (sparsity collapse), while IG stays green by using a smaller $k$.
    IG-adult ($^\star$): rescue cell (benign cap relaxed to $0.20$).
    Full numerics: Appendix~\ref{app:utility} (Table~\ref{tab:utility_full}).}
  \label{fig:overrefuse_8b}
\end{figure}

On LLaMA-3.1-8B, baseline malign refusal ranges from $0.024$ on
legal to $0.488$ on hate (Figure~\ref{fig:overrefuse_8b}). LRP, IG,
and \consensus{} lift CAST-malign refusal on hate and crime to
$0.82$-$0.95$, with benign over-refusal at most $0.08$ and PPL
within $+2$-$4$ of the unedited model. The non-attribution baselines fail in opposite
directions.

\method{Random}
reaches only $0.05$-$0.38$ on LLaMA-3.1-8B, a null
result at this scale. \method{MeanActivation} fails in the opposite
direction: it \emph{reduces} malign refusal on every domain (hate
drops $0.488 \to 0.124$). The reason is visible in the signal
density: only $\le 0.01\%$ of \method{MeanActivation}'s top rows
exceed $|C_{\text{comply}}|>0.5$ (where $C_{\text{comply}}$ is the
contrastive compliance score from Algorithm~\ref{alg:contrastive}),
against $1.9$-$8.5\%$ for LRP and IG on the same model and domain,
a roughly $500\times$ gap (Table~\ref{tab:signal_density}). The
top-$k$ set \method{MeanActivation} selects is dominated by
high-activation rows that are neither LM-critical nor
refusal-specific but do include some refusal-promoting
infrastructure; zeroing them removes that infrastructure without
adding refusal-specific signal, so the net effect is a reduction in
refusal.

Refusal is MLP-concentrated. On LLaMA-3.1-8B, $77.6\%$ of LRP's
top-$1\%$ rows are \texttt{gate\_proj} or \texttt{up\_proj}
($69.4\%$ on Qwen3-8B), with attention rows around $15\%$. The
LM-protect penalty separates these from LM-critical rows, which are
\texttt{down\_proj}-heavy. \consensus{} leads or matches LRP on $4$
  of $5$ domains and leads IG on $3$ of $5$ (hate, crime, medical),
  despite an LRP-IG Jaccard of about $5\%$ between top-$1\%$
  behavior masks. This low behavior level overlap contrasts with the
  broader LM-level agreement seen at $30\%$ masking. SorryBench
transfer is strong, with attribution selectors at $0.76$-$0.98$ on
hate and crime. Because the contrastive margin and the refusal
classifier both key on refusal-onset surface forms, this transfer to
$45$ unseen SorryBench categories is what distinguishes installed
refusal from mere refusal-shaped phrasing (we return to this in
Limitations). The cost is on OR-Bench-Hard, where over-refusal
rises from a $0.30$ baseline to $0.57$-$0.88$ at our operating
point. Reporting both benchmarks together is what separates safety
improvements from blanket over-refusal.
These gains also carry a reasoning-utility cost at viable operating
points: hate and crime lose $0.22$-$0.34$ GSM8K accuracy
(Table~\ref{tab:utility_full}).

Signal density also predicts the operating-point $k$. Across all
four models and both attribution methods, the per-domain fraction of
rows with $|C_{\text{comply}}|>0.5$ orders
\emph{hate $>$ crime $>$ adult $>$ medical $>$ legal}
(Table~\ref{tab:signal_density}). The same ordering controls the
sparsity each domain needs: hate and crime converge at
$k=0.01$-$0.02$, legal needs $k=0.05$-$0.15$. The operating point
can be predicted from calibration before any test-set evaluation.

Medical and legal collapse under sparsity, not method. Their signal
density is two to three times lower than hate's, which forces
$k \ge 0.05$-$0.10$. At the high end of that range, about 10\% of rows are zeroed
and general capability collapses (PPL $40$-$52$, GSM8K $\to 0$,
MMLU drops $\sim 45$ percentage points and IFEval by up to $60$ percentage points; Appendix
Figure~\ref{fig:ppl_vs_k}). The same collapse appears for LRP,
\consensus{}, and Random. The LM-protect term $\lambda$ is what
keeps the hate, crime, and adult cells viable; without it the same
top-$k$ rows take PPL with them, while at $\lambda=0.5$ the
operating point reaches CAST-malign $0.814$
(Table~\ref{tab:lambda_sweep_8b_lrp_hate}).

\begin{figure}[t]
  \centering
  \includegraphics[width=0.7\linewidth]{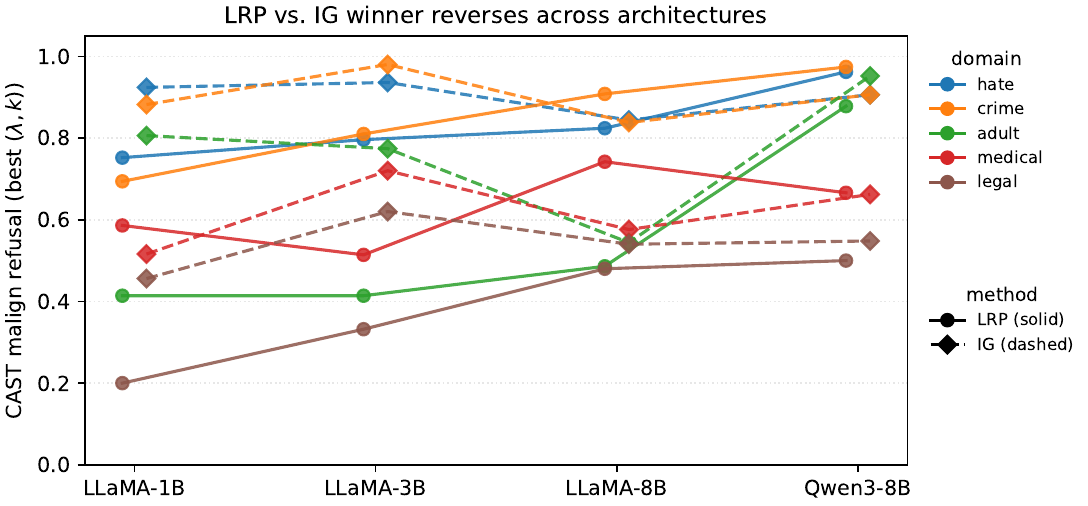}
  \caption{Cross-architecture method reversal: operating-point CAST
    malign refusal per domain (color) for LRP (solid) and IG (dashed)
    across four models. IG dominates on LLaMA-3.2-1B and LLaMA-3.2-3B
    (dashed lines above solid on 9 of 10 cells; LRP wins the medical domain on LLaMA-3.2-1B).
    On LLaMA-3.1-8B and Qwen3-8B the lines cross domain-by-domain, so the
    relative method ranking flips with architecture.}
  \label{fig:reversal}
\end{figure}

\textbf{Winning method reverses across architectures.}
\label{sec:cross}
Extending the audit to LLaMA-3.2-1B, LLaMA-3.2-3B, and Qwen3-8B
shows the winner is architecture-dependent
(Figure~\ref{fig:reversal}). IG wins $9$ of $10$ cells on the two
small LLaMA-3.2 models; LRP wins only the medical domain on
LLaMA-3.2-1B, and with a higher CAST-malign refusal rate
($0.586$ vs.\ $0.516$). On LLaMA-3.1-8B and Qwen3-8B the two
methods split evenly. The flip is coherent rather than noisy: the
five domain ordering within each model is nearly identical.
Per-model operating-point tables are in
Appendix~\ref{app:contrastive_other_models}
(Tables~\ref{tab:overrefuse_1b}, \ref{tab:overrefuse_3b},
\ref{tab:overrefuse_qwen}).

The flip tracks where each method routes relevance relative to
where the model computes the target
(Appendix Figure~\ref{fig:layer_dist}). On LLaMA-3.1-8B LRP weights
late layers and IG weights early layers; IG's early rows amplify
through more downstream transformations and yield higher raw refusal
at matched $k$, but they also bleed onto benign prompts (IG-adult
is the only cell with no feasible operating point), while LRP's
late rows stay clear of the LM pathway. On Qwen3-8B both methods
route to early layers, but LRP picks sharper rows and wins on both
axes. The faithful method on a given architecture is the one whose
routing matches where that architecture computes refusal; there is
no architecture-independent winner.

\subsection{Specificity is row-level, not depth allocation.}
\label{sec:exp_b}

\begin{table}[pt]
\centering
\small
\setlength{\tabcolsep}{4pt}
\caption{Layer-matched faithfulness audit on LLaMA-3.1-8B-LRP at the
  operating point per domain. Real = the published LRP mask;
  Ctrl mean / max = layer-count-matched random masks averaged / max
  over $N=3$ seeds; Gain = real $-$ ctrl mean. Real beats the
  control on every domain by $+0.46$ to $+0.77$. Real values are
  co-evaluated with the controls in the same run for a fair comparison;
  slight differences from Table~\ref{tab:utility_full} \rev{reflect
  re-scoring variance (per-domain deltas $0.010$-$0.080$, median
  $0.018$); only the legal cell exceeds the $n{=}500$ Wilson
  half-width}.}
\label{tab:faithfulness}
\begin{tabular}{l c c rrr r}
\toprule
Domain & $\lambda^\star$ & $k^\star$ & Real & Ctrl mean & Ctrl max & Gain \\
\midrule
hate    & 0.5 & 0.02 & \cellcolor{celllgreen} 0.814 & 0.274 & 0.386 & \cellcolor{celllred} $+$0.540 \\ 
crime   & 0.5 & 0.02 & \cellcolor{cellgreen} 0.926 & 0.264 & 0.360 & \cellcolor{celllgreen} $+$0.662 \\ 
adult   & 0.7 & 0.01 & \cellcolor{cellred} 0.498 & 0.043 & 0.080 & \cellcolor{cellred} $+$0.455 \\ 
medical & 0.5 & 0.10 & \cellcolor{celllgreen} 0.768 & 0.003 & 0.004 & \cellcolor{cellgreen} $+$0.765 \\ 
legal   & 0.5 & 0.10 & \cellcolor{celllred} 0.560 & 0.005 & 0.014 & \cellcolor{celllgreen} $+$0.555 \\ 
\bottomrule
\end{tabular}
\end{table}

The results so far establish sufficiency: zeroing the top-$k$ rows
installs refusal at the operating point. They do not yet establish
that \emph{row identities} matter: attribution might pick the right
\emph{layers} while rows within them remain interchangeable.

To separate the two, for each domain at the LRP operating point
$(\lambda^\star, k^\star)$ we build $N=3$ random masks that zero
the same number of rows in each layer as the published LRP mask, but
with uniformly random row identities. Each control is scored on
CAST-malign (Table~\ref{tab:faithfulness}).

The real LRP mask beats the layer-matched control on every domain
by $+0.46$ (adult) to $+0.77$ (medical). On medical at $k=0.10$ the
control collapses to a mean of $0.003$, so nearly all of the
$0.768$ refusal rate is attributable to which specific rows LRP
picked inside each layer. The effect is row-level: attribution identifies specific rows within
layers, not just the right layers. The subspace is redundant across
methods, but within a given method's mask the row
identities matter.

\subsection{Refusal lives in a redundant subspace.}
\label{sec:redundant}

The LRP$\leftrightarrow$IG top-$1\%$ Jaccard is $3$-$6\%$ at every
(model, domain) cell across all four models
(Table~\ref{tab:jaccard_lrp_ig}). This is about $10\times$ the
LRP-vs-Random overlap ($0.5\%$), still far short of agreement, yet
both methods install refusal. Refusal therefore lives in a redundant
subspace where many sparse masks suffice. The redundancy is not
specific to LLaMA-3.1-8B: the $3$-$6\%$ range holds across all
four models. The $70$ checkpoints are one realization
of this class, not the unique refusal edit.
\rev{This weight-space redundancy mirrors, and is predicted by, the
multi-dimensional structure of refusal in activation space:
\cite{wollschlager2025geometry} find refusal is carried by multiple
functionally independent directions rather than one, and a mechanism
with many independent directions is exactly what admits many disjoint
sufficient masks at the weight level.}

The contrastive subtraction also exposes structure across harm
categories. Two tight clusters appear, \{hate, crime\} with row
Jaccard $0.445$ and \{medical, legal\} with $0.287$, and the
cross-cluster overlap is at most $0.13$ on Qwen3-LRP
(Appendix~\ref{app:transfer}, Figure~\ref{fig:harm_clusters_heat}).
The same two-cluster pattern appears at attenuated magnitudes on
LLaMA-3.1-8B.
\rev{The tight \{hate, crime\} cluster yields a concrete, testable
prediction: a single mask composed over the union of hate and crime
calibration pairs should install refusal on both domains at once,
which we leave to future work.}

% ============================================================
\section{Discussion and Conclusion}
\label{sec:conclusion}

\textbf{Attribution is more causally faithful than rank stability suggests.}
Standard proxies (perplexity, rank correlation, calibration accuracy) cannot detect selector failure: an unfaithful selector passes all three while picking rows that are neither LM-critical nor behavior-specific. The causal audit closes this gap. At the LM level, attribution selectors identify dispensable rows orders of magnitude more accurately than non-attribution baselines, and the gap is consistent across all five models. The rank-stability dissociation is the sharpest finding: \method{MeanActivation} has near-perfect rank stability yet catastrophic causal invalidity, and \method{Wanda} inverts importance ordering on two of five architectures. A selector can be perfectly stable and still be wrong.

\textbf{Contrastive attribution installs refusal selectively.}
At the behavior level, the contrastive signal successfully isolates refusal-specific rows: attribution selectors raise CAST-malign refusal to $0.82$-$0.95$ on hate and crime on LLaMA-3.1-8B, while keeping benign over-refusal at most $0.08$ and perplexity within $+2$-$4$ points. The effect is anatomically grounded (MLP gate/up rows), generalizes to held-out red-team benchmarks, and is row-specific rather than depth-specific. The cost structure is predictable from calibration alone: signal density orders domains hate $>$ crime $>$ adult $>$ medical $>$ legal across most models, and the required sparsity follows the same ordering. Medical and legal collapse under the sparsity needed to install refusal, a limitation of the signal rather than the method. The harm-category clustering (hate/crime vs. medical/legal) suggests refusal is organized at the row level around harm type, not just around the model's general safety behavior.

\textbf{Three structural findings challenge common assumptions.}
Beyond the headline attribution-vs-baseline gap, three results cut
against common practice. First, the best-performing method reverses
across architectures: IG dominates on small LLaMA models, LRP leads
on Qwen3-8B, and neither is architecture-independent; the flip
tracks each method's layer-routing relative to where the architecture
computes refusal (Appendix Figure~\ref{fig:layer_dist}), so
practitioners must validate on their target architecture. Second,
refusal lives in a redundant subspace: LRP and IG select largely
disjoint rows (Jaccard overlap of $3$-$6\%$, meaning only $3$-$6$
out of every $100$ top-ranked rows are shared) yet both install
refusal (Table~\ref{tab:jaccard_lrp_ig}). Third, rows
where LRP and IG disagree carry stronger causal signal than the
agreed set: VETO-LRP, which keeps only the rows LRP ranks important
but IG does not, outperforms real \consensus{} on both LeRF and MoRF
on both LLaMA-3.1-8B and Qwen3-8B (Appendix~\ref{app:consensus_full});
layer-matched and rank-randomized controls confirm this is not an
artifact of depth allocation or rank averaging
(Table~\ref{tab:consensus_8b}), pointing to disagreement-aware
aggregation as a promising avenue for future work.

\textbf{What the audit establishes and what it opens.}
The audit demonstrates that attribution-based selection is causally
faithful at both levels and provides a reusable four-step protocol:
report sufficiency, specificity, utility cost, and cross-architecture
replication. For practitioners, this means attribution-guided pruning
and safety editing can be validated directly rather than relying on
rank stability or calibration accuracy as proxies. The audit does not
demonstrate necessity: many row sets suffice, and the edits
are one realization of a redundant subspace. Rank-stability proxies
systematically miss the failure modes a direct causal audit surfaces,
and architecture-dependence means there is no universal selector,
only validated ones. Detailed implications for attribution-guided
pruning, behavior localization, and circuit discovery are in
Appendix~\ref{app:extended_discussion}.

% ============================================================
\section*{Limitations}

\rev{\textbf{Scope.} The audit covers five dense decoder-only models
across LLaMA-3, Qwen3, and Gemma-3 at the language-modeling level and
four instruction-tuned variants at the behavior level, with the
stability, consensus, and agreement analyses concentrated on
LLaMA-3.1-8B and Qwen3-8B and the layer-matched faithfulness control
reported for LLaMA-3.1-8B-LRP. Mixture-of-experts and encoder-decoder
architectures, non-gradient attributors such as gradient$\times$input
or SHAP, direction- and SAE-based selectors (a distinct inference-time
intervention class), and non-English inputs fall outside the present
scope and are natural extensions.

\textbf{Evaluation.} Selector runs are single evaluations at fixed
operating points; the 70-cell grid and cross-architecture replication
provide the robustness check in place of repeated seeds (Random, the
one seeded selector, varies negligibly). We report Wilson $95\%$
sampling intervals for the refusal-editing rates
(Table~\ref{tab:wilson}), under which every headline
attribution-vs-control separation is disjoint. Refusal is scored by a
single classifier under greedy decoding; a multi-judge or
sampling-based protocol~\cite{chen2025saferluckier,schwinn2026coinflip}
would sharpen absolute rates near the feasibility boundary, though the
order-of-magnitude effects ($0.024 \to 0.9{+}$ against controls at
$0.003$) are robust to judge choice. One confound is specific to this
setup: the mask is selected via a margin over refusal-onset tokens and
the judge is sensitive to refusal surface forms, so a model that only
learned refusal-shaped openings would satisfy both. Transfer to
SorryBench across 45 unseen harm categories makes pure surface mimicry
unlikely; a template-diverse judge is the definitive check.

\textbf{Artifacts and deployment.} The 70 edited checkpoints are
diagnostic artifacts showing what row-zeroing installs at the reported
operating points; each is released for research use under a
research-use agreement, with a model card recording its operating
point, its rescue-cell flag where applicable, over-refusal and utility
figures, and a ``not for deployment'' designation. A production safety edit would use
milder operating points, broader red-team evaluation, and standard
post-training recovery. All models, datasets, and benchmarks are used
under their published terms; no personally identifiable information is
present, and no verbatim harmful prompt text is reproduced.}

% ============================================================
\section*{Ethics Statement}

This work develops methods for auditing and editing the safety
behavior of language models, and is safety-positive in intent: the
causal audit exposes selector failures that standard metrics miss and
helps practitioners verify that safety edits generalize beyond a
calibration set. The contrastive construction is sign-symmetric and
could in principle remove refusal rather than install it, so we
release only the refusal-installing checkpoints and withhold any
removal-optimized masks or code. The mechanism builds on published
ablation and steering methods and introduces no new vulnerability, and
refusal-removal tooling is already publicly available, so the release
adds negligible marginal risk. All experiments use public models and
datasets, involve no human participants, and collect no new data.

% ============================================================
\bibliographystyle{unsrtnat}
\bibliography{references}

% ============================================================
% APPENDIX - Full results for all experiments
% ============================================================
\appendix

\clearpage
\section*{Appendix}

% ============================================================
\section{Selector Descriptions and Full 30\% Numerics}
\label{app:selectors_detail}

\textbf{Selector descriptions.}
The four non-attribution baselines cover the cost spectrum from
data-free to one forward pass: \method{Random} assigns uniform
$U(0,1)$ scores averaged over 3 seeds; \method{Magnitude} uses
$s_i = \text{mean}(|W[i,:]|)$ with no calibration data;
\method{Wanda} adapts the per-weight criterion of
\cite{sun2024wanda} (originally $|W_{ij}| \cdot \|X_j\|_2$ for
individual weights) to the neuron-row level by summing across the
input dimension: $s_i = \sum_j |W_{ij}|\cdot\text{RMS}(X_j)$, where
$\text{RMS}(X_j)$ is the RMS of input feature $j$ across calibration
data (row-level aggregation is necessary because our intervention
zeroes whole output rows rather than individual weights);
\method{MeanActivation} averages the absolute output activation
across calibration data.

The three attribution-based selectors require one or more backward
passes. \method{LRP} uses AttnLRP~\cite{achtibat2024attnlrp} via
the \texttt{lxt} library to propagate relevance from the next-token
loss back through attention and MLP layers in a single
backward-like pass. \method{IG}~\cite{sundararajan2017axiomatic}
integrates the gradient along a linear path from a zero baseline
using 16 interpolation steps. \consensus{} aggregates LRP and IG
via a global Borda rank average: for each method
$M \in \{\text{LRP}, \text{IG}\}$, let
$\tilde{r}_M[i] = \text{rank}_M(i)/N$ be the normalized rank of
row $i$ among all $N$ prunable neuron-rows (lower rank = higher
importance); the \consensus{} score is
$s_{\text{C2}}[i] = \tfrac{1}{2}(\tilde{r}_\text{LRP}[i] +
\tilde{r}_\text{IG}[i])$, and rows with the lowest combined rank
are selected for masking. The motivation is error reduction: LRP
uses a conservation-based layer-by-layer propagation while IG uses
a path-integral approach, so the two methods have incompatible
failure modes and their agreement identifies rows that are robustly
unimportant under both paradigms. Scoring functions and
computational costs are summarized in Table~\ref{tab:selectors_full}.

\textbf{Model architecture details.}
Table~\ref{tab:models} gives the block, layer, and neuron counts for
each model. The five models span a roughly $9\times$ range in
prunable neuron count, from LLaMA-3.2-1B ($377\text{k}$) to
Gemma-3-12B ($3.5\text{M}$), which partly explains why attribution
selectors retain higher downstream accuracy on larger models at
matched sparsity rates.

\begin{table}[ht]
\centering
\small
\setlength{\tabcolsep}{3pt}
\caption{Models used in the general selector audit. ``Layers''
  counts all prunable Linear layers; ``Neurons'' counts all
  neuron-rows across those layers, computed from each model's
  HuggingFace config.}
\label{tab:models}
\begin{tabular}{l rrr}
\toprule
Model & Blocks & Layers & Neurons \\
\midrule
LLaMA-3.2-1B  & \cellcolor{cellred} 16 & \cellcolor{cellred} 112 & \cellcolor{cellred} 376{,}832 \\ 
LLaMA-3.2-3B  & \cellcolor{celllred} 28 & \cellcolor{celllred} 196 & \cellcolor{celllred} 774{,}144 \\ 
LLaMA-3.1-8B  & \cellcolor{celllgreen} 32 & \cellcolor{celllgreen} 224 & \cellcolor{celllgreen} 1{,}376{,}256 \\ 
Qwen3-8B   & \cellcolor{celllgreen} 32 & \cellcolor{celllgreen} 224 & \cellcolor{celllgreen} 1{,}245{,}184 \\ 
Gemma-3-12B & \cellcolor{cellgreen} 38 & \cellcolor{cellgreen} 499 & \cellcolor{cellgreen} 3{,}514{,}368 \\ 
\bottomrule
\end{tabular}
\end{table}

\begin{table}[ht]
\centering
\small
\setlength{\tabcolsep}{3pt}
\caption{Selectors evaluated. Cost: \textit{triv.}=no data;
  \textit{low}=forward only; \textit{med.}=1 backward/trace;
  \textit{high}=16$\times$ backward.}
\label{tab:selectors_full}
\begin{tabular}{l l l}
\toprule
Selector & Score & Cost \\
\midrule
\method{Random}      & $U(0,1)$, 3-seed avg. & \cellcolor{cellgreen} \textit{triv.} \\ 
\method{Magnitude}   & $\text{mean}(|W[i,:]|)$ & \cellcolor{cellgreen} \textit{triv.} \\ 
\method{Wanda}       & $\sum_j |W_{ij}| \cdot \text{RMS}(X_j)$ & \cellcolor{celllgreen} \textit{low} \\ 
\method{MeanAct.}    & $\mathbb{E}_{x,t}[|h_i(x,t)|]$ & \cellcolor{celllgreen} \textit{low} \\ 
\method{LRP}         & AttnLRP via the \texttt{lxt} library & \cellcolor{celllred} \textit{med.} \\ 
\method{IG}          & 16-step path integral & \cellcolor{celllred} \textit{high} \\ 
\consensus{}         & $\tfrac{1}{2}(\tilde{r}_\text{LRP}[i] + \tilde{r}_\text{IG}[i])$ & \cellcolor{celllred} \textit{high} \\ 
\bottomrule
\end{tabular}
\end{table}

\textbf{30\% masking numerics.}
\label{app:audit_summary_full}
Figure~\ref{fig:audit_summary} visualizes the LeRF and MoRF PPL at
$30\%$ masking as a heatmap. Table~\ref{tab:audit_summary_full}
gives the full numerics including the validity gap
(Gap $= \text{MoRF} - \text{LeRF}$) and mean downstream accuracy
across 8 \texttt{lm-eval-harness} tasks.

Key patterns: attribution selectors (LRP, IG, \consensus{}) generally sit two to four orders of magnitude below non-attribution baselines on LeRF across all five models; MoRF shows the matching asymmetry; \method{Random} produces a negative gap on 4 of 5 models; \method{Wanda} inverts on LLaMA-3.2-3B and Qwen3-8B. Downstream accuracy at $30\%$ LeRF is highest for \consensus{} on Gemma-3-12B ($0.503$) and LLaMA-3.1-8B ($0.392$), against a non-attribution floor near $0.30$.

\rev{\textbf{Reading the extreme MoRF values.} At high masking rates the
MoRF perplexities in the per-rate tables saturate at a numeric
ceiling (${\sim}7.16\times10^{38}$ for LLaMA-3.2-1B,
${\sim}1.36\times10^{38}$ for LLaMA-3.2-3B) set by float overflow of
the exponentiated loss once the model is fully destroyed; the exact
value is a representation limit, not a measurement. Entries at or
near this ceiling should be read \emph{ordinally} (``model
collapsed''), and the quantitative MoRF comparisons we draw in the
main text (e.g.\ LRP's ${\sim}10^{11}$ spike near $30\%$) are taken
from the pre-saturation regime where the values are informative.}

\begin{table}[ht]
\centering
\small
\setlength{\tabcolsep}{4pt}
\caption{Full numerics for the selector faithfulness audit at $30\%$
  masking. L = LeRF PPL$\downarrow$ (lower = safer choices),
  M = MoRF PPL$\uparrow$ (higher = better necessity identification),
  G = Gap$\uparrow$ ($= \text{M} - \text{L}$), A = mean downstream
  accuracy across 8 lm-eval-harness tasks ($\uparrow$). Full rate
  sweeps in Appendix~\ref{app:full_lerf}.}
\label{tab:audit_summary_full}
\begin{tabular}{l rrrr}
\toprule
Selector & L & M & G & A \\
\midrule
\multicolumn{5}{l}{\textit{LLaMA-3.2-1B}} \\
\midrule
\method{Random}    & \cellcolor{cellred} 548318 & \cellcolor{cellred} 344336 & \cellcolor{cellred} $-$203982 & \cellcolor{celllgreen} .304 \\ 
\method{Magnitude} & \cellcolor{celllred} 364252 & \cellcolor{celllred} 690316 & \cellcolor{celllred} 326065 & \cellcolor{cellred} .280 \\ 
\method{Wanda}     & \cellcolor{celllred} 359248 & \cellcolor{celllgreen} 1.01e6 & \cellcolor{celllred} 650003 & \cellcolor{celllred} .286 \\ 
\method{MeanAct.}  & \cellcolor{celllgreen} 93828 & \cellcolor{celllgreen} 1.76e6 & \cellcolor{celllgreen} 1.67e6 & \cellcolor{celllred} .284 \\ 
\method{LRP}       & \cellcolor{celllgreen} 288 & \cellcolor{cellgreen} 1.04e11 & \cellcolor{cellgreen} 1.04e11 & \cellcolor{celllgreen} .292 \\ 
\method{IG}        & \cellcolor{celllgreen} 603 & \cellcolor{celllgreen} 821049 & \cellcolor{celllgreen} 820446 & \cellcolor{cellgreen} .309 \\ 
\consensus{}       & \cellcolor{cellgreen} 149 & \cellcolor{celllred} 771675 & \cellcolor{celllgreen} 771526 & \cellcolor{celllgreen} .303 \\ 
\midrule
\multicolumn{5}{l}{\textit{LLaMA-3.2-3B}} \\
\midrule
\method{Random}    & \cellcolor{celllred} 101662 & \cellcolor{celllred} 149592 & \cellcolor{celllred} 47930 & \cellcolor{celllgreen} .299 \\ 
\method{Magnitude} & \cellcolor{celllgreen} 48289 & \cellcolor{celllgreen} 401191 & \cellcolor{celllgreen} 352902 & \cellcolor{celllred} .278 \\ 
\method{Wanda}     & \cellcolor{cellred} 347227 & \cellcolor{cellred} 93170 & \cellcolor{cellred} $-$254057 & \cellcolor{celllgreen} .286 \\ 
\method{MeanAct.}  & \cellcolor{celllred} 89751 & \cellcolor{celllgreen} 588584 & \cellcolor{celllgreen} 498834 & \cellcolor{celllred} .278 \\ 
\method{LRP}       & \cellcolor{celllgreen} 599 & \cellcolor{celllred} 126315 & \cellcolor{celllred} 125716 & \cellcolor{celllred} .285 \\ 
\method{IG}        & \cellcolor{celllgreen} 993 & \cellcolor{cellgreen} 852031 & \cellcolor{cellgreen} 851037 & \cellcolor{celllgreen} .286 \\ 
\consensus{}       & \cellcolor{cellgreen} 120 & \cellcolor{celllgreen} 499225 & \cellcolor{celllgreen} 499105 & \cellcolor{cellgreen} .341 \\ 
\midrule
\multicolumn{5}{l}{\textit{LLaMA-3.1-8B}} \\
\midrule
\method{Random}    & \cellcolor{celllred} 762286 & \cellcolor{cellred} 645432 & \cellcolor{cellred} $-$116854 & \cellcolor{celllgreen} .316 \\ 
\method{Magnitude} & \cellcolor{cellred} 1.05e6 & \cellcolor{celllgreen} 2.03e6 & \cellcolor{celllgreen} 976135 & \cellcolor{celllgreen} .308 \\ 
\method{Wanda}     & \cellcolor{celllred} 360083 & \cellcolor{celllred} 976078 & \cellcolor{celllred} 615995 & \cellcolor{cellred} .285 \\ 
\method{MeanAct.}  & \cellcolor{celllgreen} 139874 & \cellcolor{celllred} 930749 & \cellcolor{celllred} 790875 & \cellcolor{celllred} .299 \\ 
\method{LRP}       & \cellcolor{celllgreen} 830 & \cellcolor{cellgreen} 8.32e6 & \cellcolor{cellgreen} 8.32e6 & \cellcolor{celllred} .292 \\ 
\method{IG}        & \cellcolor{celllgreen} 98 & \cellcolor{celllgreen} 1.40e6 & \cellcolor{celllgreen} 1.40e6 & \cellcolor{celllgreen} .383 \\ 
\consensus{}       & \cellcolor{cellgreen} 66 & \cellcolor{celllgreen} 1.54e6 & \cellcolor{celllgreen} 1.54e6 & \cellcolor{cellgreen} .392 \\ 
\midrule
\multicolumn{5}{l}{\textit{Qwen3-8B}} \\
\midrule
\method{Random}    & \cellcolor{celllred} 7.19e6 & \cellcolor{cellred} 2.09e6 & \cellcolor{celllred} $-$5.10e6 & \cellcolor{celllgreen} .297 \\ 
\method{Magnitude} & \cellcolor{celllred} 7.38e6 & \cellcolor{cellgreen} 1.66e9 & \cellcolor{cellgreen} 1.65e9 & \cellcolor{celllred} .288 \\ 
\method{Wanda}     & \cellcolor{cellred} 1.95e7 & \cellcolor{celllred} 9.67e6 & \cellcolor{cellred} $-$9.78e6 & \cellcolor{celllred} .292 \\ 
\method{MeanAct.}  & \cellcolor{celllgreen} 6.59e4 & \cellcolor{celllgreen} 3.11e8 & \cellcolor{celllgreen} 3.11e8 & \cellcolor{cellred} .279 \\ 
\method{LRP}       & \cellcolor{celllgreen} 724 & \cellcolor{celllgreen} 3.74e7 & \cellcolor{celllgreen} 3.74e7 & \cellcolor{celllgreen} .332 \\ 
\method{IG}        & \cellcolor{celllgreen} 229 & \cellcolor{celllred} 2.68e7 & \cellcolor{celllred} 2.68e7 & \cellcolor{cellgreen} .355 \\ 
\consensus{}       & \cellcolor{cellgreen} 187 & \cellcolor{celllgreen} 2.83e7 & \cellcolor{celllgreen} 2.83e7 & \cellcolor{celllgreen} .348 \\ 
\midrule
\multicolumn{5}{l}{\textit{Gemma-3-12B}} \\
\midrule
\method{Random}    & \cellcolor{cellred} 3.16e9 & \cellcolor{celllred} 1.48e8 & \cellcolor{cellred} $-$3.01e9 & \cellcolor{cellred} .284 \\ 
\method{Magnitude} & \cellcolor{celllgreen} 316743 & \cellcolor{celllred} 1.40e8 & \cellcolor{celllred} 1.40e8 & \cellcolor{celllred} .289 \\ 
\method{Wanda}     & \cellcolor{celllred} 9.07e5 & \cellcolor{celllgreen} 1.66e9 & \cellcolor{celllgreen} 1.65e9 & \cellcolor{celllgreen} .292 \\ 
\method{MeanAct.}  & \cellcolor{celllred} 1.88e8 & \cellcolor{celllgreen} 4.27e10 & \cellcolor{celllgreen} 4.26e10 & \cellcolor{celllred} .287 \\ 
\method{LRP}       & \cellcolor{celllgreen} 514 & \cellcolor{cellred} 2.40e6 & \cellcolor{celllred} 2.40e6 & \cellcolor{celllgreen} .329 \\ 
\method{IG}        & \cellcolor{cellgreen} 135 & \cellcolor{cellgreen} 1.00e14 & \cellcolor{cellgreen} 1.00e14 & \cellcolor{celllgreen} .462 \\ 
\consensus{}       & \cellcolor{celllgreen} 141 & \cellcolor{celllgreen} 5.26e13 & \cellcolor{celllgreen} 5.26e13 & \cellcolor{cellgreen} .503 \\ 
\bottomrule
\end{tabular}
\end{table}

% ============================================================
\clearpage
\section{Full LeRF / MoRF / Downstream Sweeps}
\label{app:full_lerf}

Tables~\ref{tab:full_lerf_1b}-\ref{tab:full_lerf_gemma} give complete LeRF and MoRF PPL at all 19 masking rates for all five models; each table has a LeRF section (upper half) and a MoRF section (lower half).
Tables~\ref{tab:downstream_full_8b}-\ref{tab:downstream_avg_all} give downstream task accuracy: Table~\ref{tab:downstream_full_8b} at representative masking rates for LLaMA-8B, Table~\ref{tab:downstream_30pct} as a 30\% snapshot across four models, and Table~\ref{tab:downstream_avg_all} as mean accuracy across all rates and all five models.

\textbf{Notable per-model findings.}
\method{LRP} on LLaMA-3.2-1B achieves MoRF$=1.04\times10^{11}$ at $30\%$, collapsing to $1.30\times10^{31}$ at $35\%$, the strongest necessity identification in the dataset, $4$ orders of magnitude above the LLaMA-8B value.
On LLaMA-3.2-3B, \method{IG} outperforms \method{LRP} on MoRF at $30\%$ ($852\text{k}$ vs.\ $126\text{k}$), an early signal of architecture-dependent method ordering.
On Qwen3-8B, \method{Random} also inverts (MoRF $2.09\times10^6 <$ LeRF $7.19\times10^6$), not just \method{Wanda}.
On Gemma-3-12B, \method{IG} achieves the highest 30\% MoRF in the dataset ($1.00\times10^{14}$); \method{Wanda} and \method{MeanAct.} at $5\%$ LeRF still equal dense PPL ($19.0$), confirming the bottom-end of their rankings is correctly identifying low-importance rows even though both selectors collapse rapidly past $10\%$.

\begin{figure}[ht]
  \centering
  \includegraphics[width=\textwidth]{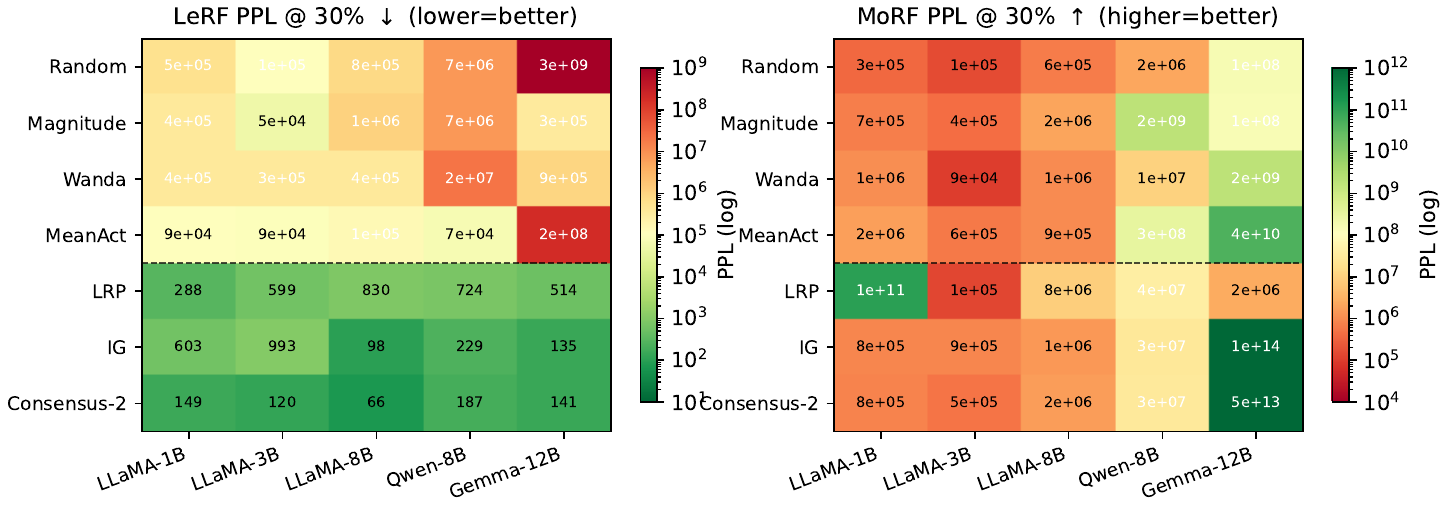}
  \caption{Selector faithfulness audit at $30\%$ masking as a
    heatmap: seven selectors $\times$ five models. Left: LeRF PPL
    (lower greener = ``safe'' set is genuinely safe). Right: MoRF
    PPL (higher greener = ``important'' set is genuinely important).
    Dashed line separates the four non-attribution selectors (top)
    from the three attribution selectors (bottom). Compact companion
    to Figure~\ref{fig:lerf_all}.}
  \label{fig:audit_summary}
\end{figure}

\textbf{30\% snapshot heatmap (across all 5 models, 7 selectors).}
Figure~\ref{fig:audit_summary} summarizes the LeRF and MoRF PPL at $30\%$ masking as a color-coded matrix, with attribution selectors visually separated from non-attribution baselines by a dashed rule. The heatmap is a compact companion to Figure~\ref{fig:lerf_all}: where Figure~\ref{fig:lerf_all} shows the full $0$-$50\%$ degradation trajectory, this figure collapses the $30\%$ slice into a single matrix that surfaces the per-cell ranking and downstream-accuracy structure at a glance.

% ============================================================
\textbf{Downstream benchmark accuracy.}
\label{app:downstream}

\method{IG} and \consensus{} maintain near-dense accuracy at $10\%$
masking ($0.580$-$0.584$ vs.\ $0.671$ dense on LLaMA-8B). LAMBADA
is the most sensitive discriminator: non-attribution baselines drop
to $0$ at $10\%$ while \method{IG}/\consensus{} retain
$0.528$-$0.586$, and \method{LRP} retains $0.086$. \method{LRP} is
weaker on utility than \method{IG}/\consensus{} at low rates
despite stronger MoRF, suggesting LRP's mask spans more capability-relevant
rows.

Table~\ref{tab:downstream_full_8b} gives per-task accuracy at
representative masking rates for LLaMA-8B; Table~\ref{tab:downstream_30pct}
gives the $30\%$ snapshot across four models; Table~\ref{tab:downstream_avg_all}
gives mean accuracy across all rates and all five models, showing
that attribution selectors maintain higher accuracy than baselines
at low-to-moderate rates ($10$-$30\%$) before all methods converge
to the collapsed floor ($\sim$0.28) at high rates ($50$-$90\%$).

\clearpage
% ---- LLaMA-1B ----
\begin{table*}[ht]
\centering
\tiny
\setlength{\tabcolsep}{2pt}
\caption{Full LeRF / MoRF PPL: LLaMA-3.2-1B. 0-90\% in 5\% steps.
  LeRF$\downarrow$ = least-relevant-first (lower = safer choices);
  MoRF$\uparrow$ = most-relevant-first (higher = better necessity identification). Companion plot: Fig~\ref{fig:sweep_1b}.}
\label{tab:full_lerf_1b}
\begin{tabular}{l *{19}{r}}
\toprule
Selector & 0\% & 5\% & 10\% & 15\% & 20\% & 25\% & 30\% & 35\% & 40\% & 45\% & 50\% & 55\% & 60\% & 65\% & 70\% & 75\% & 80\% & 85\% & 90\% \\
\midrule
\multicolumn{20}{l}{\textit{\textbf{LeRF} (least-relevant-first masking; lower PPL = safer choices $\downarrow$)}} \\
\midrule
\method{Random}    & 16.5 & \cellcolor{celllred} 993 & \cellcolor{celllred} 11.4k & \cellcolor{celllgreen} 35.8k & \cellcolor{celllgreen} 86.6k & \cellcolor{celllred} 240k & \cellcolor{cellred} 548k & \cellcolor{cellred} 914k & \cellcolor{cellred} 556k & \cellcolor{celllred} 675k & \cellcolor{celllred} 469k & \cellcolor{cellred} 392k & \cellcolor{celllred} 572k & \cellcolor{celllred} 608k & \cellcolor{celllred} 1.18e6 & \cellcolor{celllred} 1.30e6 & \cellcolor{celllred} 1.29e7 & \cellcolor{celllred} 2.09e9 & \cellcolor{celllred} 6.44e15 \\ 
\method{Magnitude} & 16.5 & \cellcolor{cellred} 13.1k & \cellcolor{cellred} 74.7k & \cellcolor{cellred} 340k & \cellcolor{cellred} 743k & \cellcolor{cellred} 513k & \cellcolor{celllred} 364k & \cellcolor{celllred} 281k & \cellcolor{celllgreen} 256k & \cellcolor{celllgreen} 234k & \cellcolor{celllgreen} 188k & \cellcolor{celllgreen} 159k & \cellcolor{cellred} 787k & \cellcolor{cellred} 1.91e6 & \cellcolor{cellred} 1.97e6 & \cellcolor{cellred} 1.53e6 & \cellcolor{cellred} 2.05e10 & \cellcolor{cellred} 3.15e10 & \cellcolor{cellred} 3.39e16 \\ 
\method{Wanda}     & 16.5 & \cellcolor{celllred} 844 & \cellcolor{celllred} 18.3k & \cellcolor{celllred} 145k & \cellcolor{celllred} 139k & \cellcolor{celllred} 438k & \cellcolor{celllred} 359k & \cellcolor{celllred} 378k & \cellcolor{celllred} 314k & \cellcolor{cellred} 1.50e6 & \cellcolor{cellred} 1.23e6 & \cellcolor{celllred} 357k & \cellcolor{celllred} 375k & \cellcolor{celllgreen} 244k & \cellcolor{celllgreen} 190k & \cellcolor{celllgreen} 287k & \cellcolor{celllred} 1.34e6 & \cellcolor{celllred} 4.00e6 & \cellcolor{celllred} 1.52e7 \\ 
\method{MeanAct.}  & 16.5 & \cellcolor{celllgreen} 430 & \cellcolor{celllgreen} 2180 & \cellcolor{celllred} 86.2k & \cellcolor{celllred} 96.9k & \cellcolor{celllgreen} 103k & \cellcolor{celllgreen} 93.8k & \cellcolor{celllgreen} 144k & \cellcolor{celllred} 399k & \cellcolor{celllred} 467k & \cellcolor{celllred} 413k & \cellcolor{celllred} 349k & \cellcolor{celllgreen} 292k & \cellcolor{celllred} 546k & \cellcolor{celllred} 540k & \cellcolor{celllred} 313k & \cellcolor{celllgreen} 300k & \cellcolor{celllgreen} 555k & \cellcolor{celllgreen} 9.46e6 \\ 
\method{LRP}       & 16.5 & \cellcolor{cellgreen} 21.1 & \cellcolor{cellgreen} 29.3 & \cellcolor{celllgreen} 49.9 & \cellcolor{celllgreen} 102 & \cellcolor{celllgreen} 163 & \cellcolor{celllgreen} 288 & \cellcolor{celllgreen} 458 & \cellcolor{celllgreen} 623 & \cellcolor{celllgreen} 856 & \cellcolor{celllgreen} 1209 & \cellcolor{cellgreen} 1571 & \cellcolor{cellgreen} 2247 & \cellcolor{cellgreen} 3635 & \cellcolor{cellgreen} 5366 & \cellcolor{cellgreen} 13.5k & \cellcolor{cellgreen} 12.2k & \cellcolor{cellgreen} 6039 & \cellcolor{cellgreen} 13.9k \\ 
\method{IG}        & 16.5 & \cellcolor{celllgreen} 24.8 & \cellcolor{celllgreen} 39.7 & \cellcolor{celllgreen} 69.0 & \cellcolor{celllgreen} 136 & \cellcolor{celllgreen} 283 & \cellcolor{celllgreen} 603 & \cellcolor{celllgreen} 1084 & \cellcolor{celllgreen} 1541 & \cellcolor{celllgreen} 1845 & \cellcolor{celllgreen} 3610 & \cellcolor{celllgreen} 4696 & \cellcolor{celllgreen} 4847 & \cellcolor{celllgreen} 5379 & \cellcolor{celllgreen} 7130 & \cellcolor{celllgreen} 18.7k & \cellcolor{celllgreen} 31.4k & \cellcolor{celllgreen} 163k & \cellcolor{celllgreen} 7.34e6 \\ 
\consensus{}       & 16.5 & \cellcolor{celllgreen} 22.3 & \cellcolor{celllgreen} 31.7 & \cellcolor{cellgreen} 46.3 & \cellcolor{cellgreen} 65.7 & \cellcolor{cellgreen} 99.6 & \cellcolor{cellgreen} 149 & \cellcolor{cellgreen} 220 & \cellcolor{cellgreen} 335 & \cellcolor{cellgreen} 509 & \cellcolor{cellgreen} 834 & \cellcolor{celllgreen} 1646 & \cellcolor{celllgreen} 5662 & \cellcolor{celllgreen} 8490 & \cellcolor{celllgreen} 10.4k & \cellcolor{celllgreen} 15.5k & \cellcolor{celllgreen} 12.7k & \cellcolor{celllgreen} 16.1k & \cellcolor{celllgreen} 109k \\ 
\midrule
\multicolumn{20}{l}{\textit{\textbf{MoRF} (most-relevant-first masking; higher PPL = better necessity identification $\uparrow$)}} \\
\midrule
\method{Random}    & 16.5 & \cellcolor{celllred} 15.1k & \cellcolor{cellred} 13.9k & \cellcolor{cellred} 44.3k & \cellcolor{cellred} 237k & \cellcolor{cellred} 306k & \cellcolor{cellred} 344k & \cellcolor{cellred} 393k & \cellcolor{celllred} 752k & \cellcolor{celllred} 1.22e6 & \cellcolor{celllred} 664k & \cellcolor{cellred} 992k & \cellcolor{cellred} 689k & \cellcolor{cellred} 1.09e6 & \cellcolor{celllred} 2.43e6 & \cellcolor{celllred} 2.10e6 & \cellcolor{celllred} 2.55e7 & \cellcolor{celllred} 8.46e9 & \cellcolor{celllred} 1.72e16 \\ 
\method{Magnitude} & 16.5 & \cellcolor{celllred} 52.1k & \cellcolor{celllred} 170k & \cellcolor{celllred} 183k & \cellcolor{celllred} 367k & \cellcolor{celllgreen} 591k & \cellcolor{celllred} 690k & \cellcolor{celllgreen} 594k & \cellcolor{celllred} 840k & \cellcolor{celllred} 1.54e6 & \cellcolor{celllgreen} 1.71e6 & \cellcolor{celllred} 1.87e6 & \cellcolor{celllred} 2.15e6 & \cellcolor{celllred} 1.68e6 & \cellcolor{cellred} 1.26e6 & \cellcolor{cellred} 1.81e6 & \cellcolor{cellred} 3.31e6 & \cellcolor{cellred} 2.06e6 & \cellcolor{cellred} 789k \\ 
\method{Wanda}     & 16.5 & \cellcolor{cellred} 8314 & \cellcolor{celllgreen} 442k & \cellcolor{celllgreen} 1.33e6 & \cellcolor{celllgreen} 1.02e6 & \cellcolor{celllgreen} 710k & \cellcolor{celllgreen} 1.01e6 & \cellcolor{celllgreen} 802k & \cellcolor{cellred} 295k & \cellcolor{celllgreen} 2.31e6 & \cellcolor{celllred} 841k & \cellcolor{celllgreen} 3.00e6 & \cellcolor{celllgreen} 8.67e6 & \cellcolor{celllgreen} 4.03e6 & \cellcolor{celllgreen} 4.82e6 & \cellcolor{celllred} 8.62e7 & \cellcolor{celllred} 1.17e11 & \cellcolor{celllred} 1.73e22 & \cellcolor{celllred} 3.47e35 \\ 
\method{MeanAct.}  & 16.5 & \cellcolor{celllgreen} 298k & \cellcolor{celllred} 300k & \cellcolor{celllred} 676k & \cellcolor{celllgreen} 1.04e6 & \cellcolor{cellgreen} 1.28e6 & \cellcolor{celllgreen} 1.76e6 & \cellcolor{celllgreen} 1.35e6 & \cellcolor{celllgreen} 1.34e6 & \cellcolor{celllgreen} 1.72e6 & \cellcolor{celllgreen} 3.56e6 & \cellcolor{celllgreen} 2.55e6 & \cellcolor{celllred} 3.07e6 & \cellcolor{celllred} 1.15e6 & \cellcolor{celllred} 3.37e6 & \cellcolor{celllgreen} 1.11e13 & \cellcolor{celllgreen} 1.29e28 & \cellcolor{celllgreen} 1.99e38 & \cellcolor{celllgreen} 6.57e38 \\ 
\method{LRP}       & 16.5 & \cellcolor{cellgreen} 387k & \cellcolor{celllgreen} 800k & \cellcolor{cellgreen} 2.66e6 & \cellcolor{celllgreen} 1.01e6 & \cellcolor{celllgreen} 1.20e6 & \cellcolor{cellgreen} 1.04e11 & \cellcolor{cellgreen} 1.30e31 & \cellcolor{cellgreen} 1.07e38 & \cellcolor{cellgreen} 5.03e38 & \cellcolor{cellgreen} 6.98e38 & \cellcolor{cellgreen} 7.16e38 & \cellcolor{cellgreen} 7.16e38 & \cellcolor{cellgreen} 7.16e38 & \cellcolor{cellgreen} 7.16e38 & \cellcolor{cellgreen} 7.16e38 & \cellcolor{cellgreen} 7.16e38 & \cellcolor{cellgreen} 7.16e38 & \cellcolor{cellgreen} 7.16e38 \\ 
\method{IG}        & 16.5 & \cellcolor{celllgreen} 260k & \cellcolor{celllgreen} 477k & \cellcolor{celllgreen} 906k & \cellcolor{cellgreen} 1.33e6 & \cellcolor{celllred} 470k & \cellcolor{celllgreen} 821k & \cellcolor{celllred} 499k & \cellcolor{celllgreen} 843k & \cellcolor{cellred} 807k & \cellcolor{cellred} 528k & \cellcolor{celllred} 1.66e6 & \cellcolor{celllgreen} 2.95e9 & \cellcolor{celllgreen} 9.95e16 & \cellcolor{celllgreen} 1.43e33 & \cellcolor{cellgreen} 7.16e38 & \cellcolor{cellgreen} 7.16e38 & \cellcolor{cellgreen} 7.16e38 & \cellcolor{cellgreen} 7.16e38 \\ 
\consensus{}       & 16.5 & \cellcolor{celllgreen} 303k & \cellcolor{cellgreen} 1.68e6 & \cellcolor{celllgreen} 1.81e6 & \cellcolor{celllred} 473k & \cellcolor{celllred} 405k & \cellcolor{celllred} 772k & \cellcolor{celllred} 453k & \cellcolor{celllgreen} 1.06e8 & \cellcolor{celllgreen} 1.53e16 & \cellcolor{celllgreen} 4.13e28 & \cellcolor{celllgreen} 1.53e38 & \cellcolor{celllgreen} 5.59e38 & \cellcolor{cellgreen} 7.16e38 & \cellcolor{cellgreen} 7.16e38 & \cellcolor{cellgreen} 7.16e38 & \cellcolor{cellgreen} 7.16e38 & \cellcolor{cellgreen} 7.16e38 & \cellcolor{cellgreen} 7.16e38 \\ 
\bottomrule
\end{tabular}
\end{table*}

\begin{figure}[pb]
  \centering
  \includegraphics[width=\linewidth]{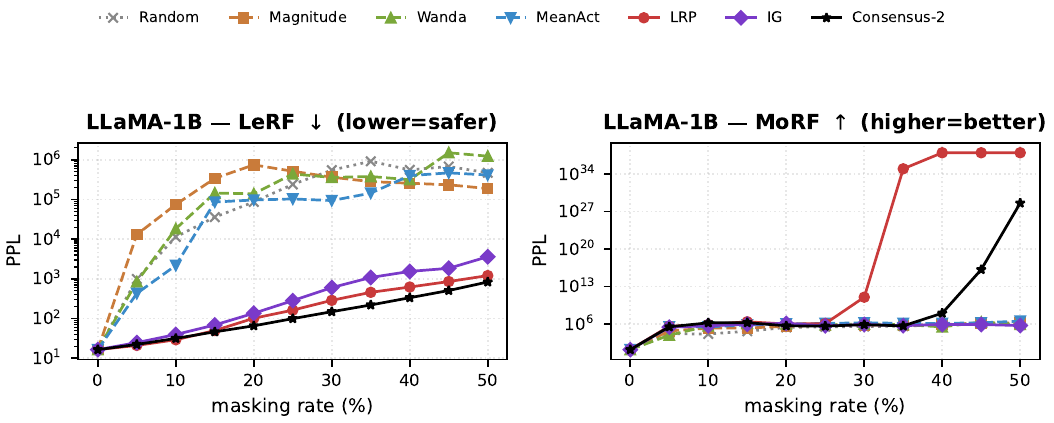}
  \caption{\textbf{LLaMA-3.2-1B: LeRF / MoRF degradation (companion to Table~\ref{tab:full_lerf_1b}).} Per-rate PPL for all seven selectors over $0$--$50\%$ masking (log $y$; the informative range before all selectors saturate at model collapse): LeRF (left, lower~$=$~safer choices) and MoRF (right, higher~$=$~better necessity identification). Attribution selectors (LRP, IG, \consensus{}) sit orders of magnitude below the non-attribution baselines on LeRF at every rate; LRP's MoRF spikes to ${\sim}10^{11}$ near $30\%$, the strongest necessity signal in the audit.}
  \label{fig:sweep_1b}
\end{figure}

% ---- LLaMA-3B ----
\begin{table*}[ht]
\centering
\tiny
\setlength{\tabcolsep}{2pt}
\caption{Full LeRF / MoRF PPL: LLaMA-3.2-3B. Same format as Table~\ref{tab:full_lerf_1b}. Companion plot: Fig~\ref{fig:sweep_3b}.}
\label{tab:full_lerf_3b}
\begin{tabular}{l *{19}{r}}
\toprule
Selector & 0\% & 5\% & 10\% & 15\% & 20\% & 25\% & 30\% & 35\% & 40\% & 45\% & 50\% & 55\% & 60\% & 65\% & 70\% & 75\% & 80\% & 85\% & 90\% \\
\midrule
\multicolumn{20}{l}{\textit{\textbf{LeRF} $\downarrow$}} \\
\midrule
\method{Random}    & 13.3 & \cellcolor{celllgreen} 419 & \cellcolor{celllred} 7085 & \cellcolor{celllgreen} 14.9k & \cellcolor{celllred} 29.9k & \cellcolor{celllred} 89.5k & \cellcolor{celllred} 102k & \cellcolor{celllred} 174k & \cellcolor{celllred} 279k & \cellcolor{cellred} 275k & \cellcolor{cellred} 420k & \cellcolor{cellred} 735k & \cellcolor{celllred} 529k & \cellcolor{cellred} 682k & \cellcolor{celllred} 407k & \cellcolor{cellred} 466k & \cellcolor{celllgreen} 346k & \cellcolor{celllgreen} 557k & \cellcolor{celllred} 4.27e6 \\ 
\method{Magnitude} & 13.3 & \cellcolor{cellred} 12.7k & \cellcolor{cellred} 52.3k & \cellcolor{cellred} 84.6k & \cellcolor{celllred} 103k & \cellcolor{celllred} 148k & \cellcolor{celllgreen} 48.3k & \cellcolor{celllred} 277k & \cellcolor{celllgreen} 122k & \cellcolor{celllred} 242k & \cellcolor{celllred} 274k & \cellcolor{celllred} 223k & \cellcolor{celllred} 225k & \cellcolor{celllred} 290k & \cellcolor{celllgreen} 292k & \cellcolor{celllgreen} 163k & \cellcolor{cellred} 777k & \cellcolor{cellred} 7.32e8 & \cellcolor{cellred} 2.38e18 \\ 
\method{Wanda}     & 13.3 & \cellcolor{celllred} 12.0k & \cellcolor{celllred} 15.5k & \cellcolor{celllred} 39.4k & \cellcolor{cellred} 108k & \cellcolor{cellred} 387k & \cellcolor{cellred} 347k & \cellcolor{cellred} 418k & \cellcolor{cellred} 427k & \cellcolor{celllred} 229k & \cellcolor{celllgreen} 139k & \cellcolor{celllred} 159k & \cellcolor{cellred} 700k & \cellcolor{celllred} 392k & \cellcolor{celllred} 313k & \cellcolor{celllred} 337k & \cellcolor{celllred} 657k & \cellcolor{celllred} 4.83e6 & \cellcolor{celllred} 8.80e7 \\ 
\method{MeanAct.}  & 13.3 & \cellcolor{celllred} 495 & \cellcolor{celllgreen} 1855 & \cellcolor{celllred} 29.2k & \cellcolor{celllgreen} 24.1k & \cellcolor{celllgreen} 31.4k & \cellcolor{celllred} 89.8k & \cellcolor{celllgreen} 113k & \cellcolor{celllred} 151k & \cellcolor{celllgreen} 153k & \cellcolor{celllred} 169k & \cellcolor{celllgreen} 156k & \cellcolor{celllgreen} 147k & \cellcolor{celllgreen} 207k & \cellcolor{cellred} 415k & \cellcolor{celllred} 219k & \cellcolor{celllred} 349k & \cellcolor{celllred} 660k & \cellcolor{celllgreen} 482k \\ 
\method{LRP}       & 13.3 & \cellcolor{celllgreen} 18.2 & \cellcolor{celllgreen} 33.9 & \cellcolor{celllgreen} 121 & \cellcolor{celllgreen} 261 & \cellcolor{celllgreen} 393 & \cellcolor{celllgreen} 599 & \cellcolor{celllgreen} 1069 & \cellcolor{celllgreen} 1617 & \cellcolor{celllgreen} 2405 & \cellcolor{celllgreen} 2557 & \cellcolor{celllgreen} 2267 & \cellcolor{cellgreen} 2321 & \cellcolor{cellgreen} 2402 & \cellcolor{cellgreen} 3380 & \cellcolor{cellgreen} 4194 & \cellcolor{cellgreen} 5016 & \cellcolor{celllgreen} 14.7k & \cellcolor{cellgreen} 32.1k \\ 
\method{IG}        & 13.3 & \cellcolor{celllgreen} 17.2 & \cellcolor{cellgreen} 25.6 & \cellcolor{cellgreen} 40.8 & \cellcolor{cellgreen} 65.5 & \cellcolor{celllgreen} 587 & \cellcolor{celllgreen} 993 & \cellcolor{celllgreen} 2005 & \cellcolor{celllgreen} 2027 & \cellcolor{celllgreen} 2809 & \cellcolor{celllgreen} 2108 & \cellcolor{celllgreen} 2582 & \cellcolor{celllgreen} 3210 & \cellcolor{celllgreen} 7736 & \cellcolor{celllgreen} 16.8k & \cellcolor{celllgreen} 12.6k & \cellcolor{celllgreen} 47.9k & \cellcolor{celllgreen} 46.6k & \cellcolor{celllgreen} 186k \\ 
\consensus{}       & 13.3 & \cellcolor{cellgreen} 16.0 & \cellcolor{celllgreen} 257 & \cellcolor{celllgreen} 323 & \cellcolor{celllgreen} 240 & \cellcolor{cellgreen} 74.9 & \cellcolor{cellgreen} 120 & \cellcolor{cellgreen} 204 & \cellcolor{cellgreen} 365 & \cellcolor{cellgreen} 551 & \cellcolor{cellgreen} 857 & \cellcolor{cellgreen} 1776 & \cellcolor{celllgreen} 3430 & \cellcolor{celllgreen} 6962 & \cellcolor{celllgreen} 12.9k & \cellcolor{celllgreen} 35.8k & \cellcolor{celllgreen} 10.4k & \cellcolor{cellgreen} 11.8k & \cellcolor{celllgreen} 48.3k \\ 
\midrule
\multicolumn{20}{l}{\textit{\textbf{MoRF} $\uparrow$}} \\
\midrule
\method{Random}    & 13.3 & \cellcolor{cellred} 882 & \cellcolor{cellred} 9042 & \cellcolor{cellred} 28.9k & \cellcolor{cellred} 35.8k & \cellcolor{cellred} 66.5k & \cellcolor{celllred} 150k & \cellcolor{celllred} 229k & \cellcolor{cellred} 225k & \cellcolor{celllred} 203k & \cellcolor{cellred} 325k & \cellcolor{celllred} 344k & \cellcolor{celllred} 381k & \cellcolor{celllred} 418k & \cellcolor{celllred} 460k & \cellcolor{celllred} 430k & \cellcolor{cellred} 372k & \cellcolor{celllred} 1.06e6 & \cellcolor{celllred} 2.60e6 \\ 
\method{Magnitude} & 13.3 & \cellcolor{celllred} 19.6k & \cellcolor{celllred} 55.8k & \cellcolor{celllred} 174k & \cellcolor{celllred} 342k & \cellcolor{celllred} 252k & \cellcolor{celllgreen} 401k & \cellcolor{celllgreen} 423k & \cellcolor{celllgreen} 680k & \cellcolor{celllgreen} 1.12e6 & \cellcolor{celllgreen} 1.64e6 & \cellcolor{celllgreen} 1.69e6 & \cellcolor{celllgreen} 1.50e6 & \cellcolor{celllgreen} 1.26e6 & \cellcolor{celllgreen} 757k & \cellcolor{celllred} 964k & \cellcolor{celllred} 927k & \cellcolor{cellred} 653k & \cellcolor{cellred} 208k \\ 
\method{Wanda}     & 13.3 & \cellcolor{celllred} 66.4k & \cellcolor{celllred} 94.1k & \cellcolor{celllred} 44.6k & \cellcolor{celllred} 192k & \cellcolor{celllred} 161k & \cellcolor{cellred} 93.2k & \cellcolor{cellred} 163k & \cellcolor{celllred} 348k & \cellcolor{celllgreen} 709k & \cellcolor{celllred} 673k & \cellcolor{celllgreen} 634k & \cellcolor{celllred} 372k & \cellcolor{cellred} 148k & \cellcolor{celllred} 330k & \cellcolor{cellred} 209k & \cellcolor{celllred} 8.83e10 & \cellcolor{celllred} 8.89e13 & \cellcolor{celllred} 8.49e29 \\ 
\method{MeanAct.}  & 13.3 & \cellcolor{celllgreen} 164k & \cellcolor{celllgreen} 459k & \cellcolor{celllgreen} 303k & \cellcolor{celllgreen} 358k & \cellcolor{celllgreen} 501k & \cellcolor{celllgreen} 589k & \cellcolor{celllgreen} 755k & \cellcolor{celllgreen} 878k & \cellcolor{celllgreen} 665k & \cellcolor{celllred} 687k & \cellcolor{celllred} 588k & \cellcolor{celllgreen} 711k & \cellcolor{celllred} 609k & \cellcolor{cellred} 174k & \cellcolor{celllgreen} 4.48e11 & \cellcolor{celllgreen} 1.36e25 & \cellcolor{celllgreen} 4.08e36 & \cellcolor{celllgreen} 2.80e37 \\ 
\method{LRP}       & 13.3 & \cellcolor{celllgreen} 259k & \cellcolor{celllgreen} 353k & \cellcolor{cellgreen} 697k & \cellcolor{celllgreen} 483k & \cellcolor{cellgreen} 713k & \cellcolor{celllred} 126k & \cellcolor{cellgreen} 1.39e6 & \cellcolor{cellgreen} 1.75e6 & \cellcolor{cellgreen} 7.98e11 & \cellcolor{cellgreen} 4.04e23 & \cellcolor{cellgreen} 2.32e27 & \cellcolor{cellgreen} 2.66e27 & \cellcolor{cellgreen} 5.53e27 & \cellcolor{celllgreen} 9.31e27 & \cellcolor{celllgreen} 1.40e28 & \cellcolor{celllgreen} 6.45e28 & \cellcolor{celllgreen} 2.29e29 & \cellcolor{celllgreen} 1.08e30 \\ 
\method{IG}        & 13.3 & \cellcolor{celllgreen} 290k & \cellcolor{celllgreen} 215k & \cellcolor{celllgreen} 315k & \cellcolor{celllgreen} 380k & \cellcolor{celllgreen} 449k & \cellcolor{cellgreen} 852k & \cellcolor{celllred} 399k & \cellcolor{celllgreen} 371k & \cellcolor{celllred} 606k & \cellcolor{celllgreen} 1.07e6 & \cellcolor{cellred} 263k & \cellcolor{cellred} 367k & \cellcolor{celllgreen} 3.05e9 & \cellcolor{celllgreen} 8.19e18 & \cellcolor{celllgreen} 4.02e27 & \cellcolor{cellgreen} 1.36e38 & \cellcolor{cellgreen} 1.36e38 & \cellcolor{cellgreen} 1.36e38 \\ 
\consensus{}       & 13.3 & \cellcolor{cellgreen} 455k & \cellcolor{cellgreen} 559k & \cellcolor{celllgreen} 423k & \cellcolor{cellgreen} 545k & \cellcolor{celllgreen} 373k & \cellcolor{celllgreen} 499k & \cellcolor{celllgreen} 455k & \cellcolor{celllred} 344k & \cellcolor{cellred} 200k & \cellcolor{celllgreen} 1.51e6 & \cellcolor{celllgreen} 951k & \cellcolor{celllgreen} 4.29e7 & \cellcolor{celllgreen} 1.70e24 & \cellcolor{cellgreen} 5.96e37 & \cellcolor{cellgreen} 1.35e38 & \cellcolor{cellgreen} 1.36e38 & \cellcolor{cellgreen} 1.36e38 & \cellcolor{cellgreen} 1.36e38 \\ 
\bottomrule
\end{tabular}
\end{table*}

\begin{figure}[ht]
  \centering
  \includegraphics[width=\linewidth]{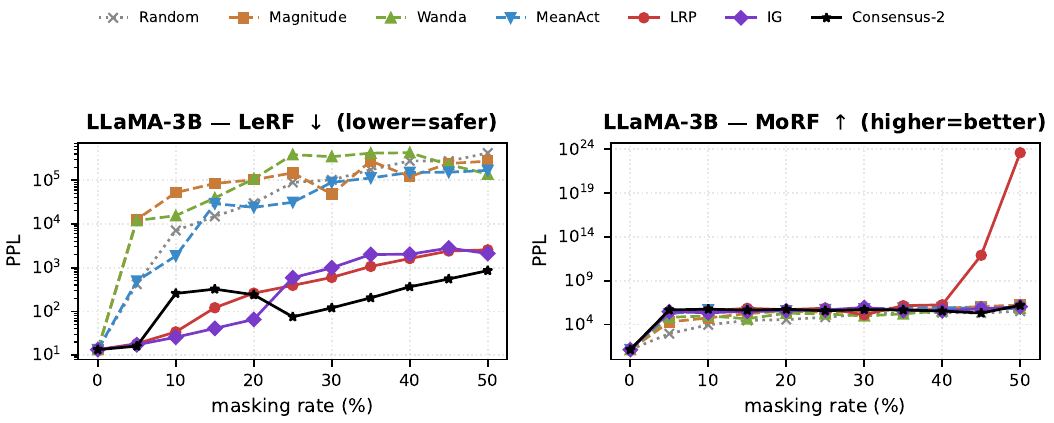}
  \caption{\textbf{LLaMA-3.2-3B: LeRF / MoRF degradation (companion to Table~\ref{tab:full_lerf_3b}).} Per-rate PPL for all seven selectors over $0$--$50\%$ masking (log $y$; the informative range before all selectors saturate at model collapse): LeRF (left, lower~$=$~safer choices) and MoRF (right, higher~$=$~better necessity identification). Attribution stays well below the baselines on LeRF throughout; on MoRF, IG overtakes LRP around $30\%$ ($852$k vs.\ $126$k), an early sign of architecture-dependent method ordering.}
  \label{fig:sweep_3b}
\end{figure}
\clearpage
% ---- LLaMA-8B ----
\begin{table*}[ht]
\centering
\tiny
\setlength{\tabcolsep}{2pt}
\caption{Full LeRF / MoRF PPL: LLaMA-3.1-8B. Same format as Table~\ref{tab:full_lerf_1b}. Companion plot: Fig~\ref{fig:sweep_8b}.}
\label{tab:full_lerf_8b}
\begin{tabular}{l *{19}{r}}
\toprule
Selector & 0\% & 5\% & 10\% & 15\% & 20\% & 25\% & 30\% & 35\% & 40\% & 45\% & 50\% & 55\% & 60\% & 65\% & 70\% & 75\% & 80\% & 85\% & 90\% \\
\midrule
\multicolumn{20}{l}{\textit{\textbf{LeRF} $\downarrow$}} \\
\midrule
\method{Random}    & 10.6 & \cellcolor{celllred} 5647 & \cellcolor{celllred} 40.1k & \cellcolor{cellred} 635k & \cellcolor{celllred} 420k & \cellcolor{celllred} 571k & \cellcolor{celllred} 762k & \cellcolor{cellred} 768k & \cellcolor{cellred} 1.41e6 & \cellcolor{celllred} 1.24e6 & \cellcolor{celllred} 584k & \cellcolor{cellred} 2.01e6 & \cellcolor{celllred} 880k & \cellcolor{cellred} 1.64e6 & \cellcolor{cellred} 2.90e6 & \cellcolor{cellred} 4.88e6 & \cellcolor{celllred} 3.48e6 & \cellcolor{celllred} 5.19e6 & \cellcolor{celllred} 3.17e6 \\ 
\method{Magnitude} & 10.6 & \cellcolor{cellred} 91.8k & \cellcolor{cellred} 2.19e6 & \cellcolor{celllred} 393k & \cellcolor{cellred} 892k & \cellcolor{cellred} 744k & \cellcolor{cellred} 1.05e6 & \cellcolor{celllred} 415k & \cellcolor{celllred} 594k & \cellcolor{celllred} 4.37e6 & \cellcolor{celllred} 816k & \cellcolor{celllred} 92.1k & \cellcolor{cellred} 1.00e6 & \cellcolor{celllred} 551k & \cellcolor{celllgreen} 55.7k & \cellcolor{celllred} 318k & \cellcolor{celllgreen} 64.2k & \cellcolor{celllgreen} 3.76e6 & \cellcolor{celllgreen} 328k \\ 
\method{Wanda}     & 10.6 & \cellcolor{celllred} 77.8k & \cellcolor{celllred} 20.0k & \cellcolor{celllred} 296k & \cellcolor{celllred} 149k & \cellcolor{celllred} 219k & \cellcolor{celllred} 360k & \cellcolor{celllgreen} 193k & \cellcolor{celllgreen} 122k & \cellcolor{celllgreen} 21.7k & \cellcolor{celllgreen} 89.1k & \cellcolor{celllgreen} 63.4k & \cellcolor{celllgreen} 39.1k & \cellcolor{celllred} 640k & \cellcolor{celllred} 403k & \cellcolor{celllred} 686k & \cellcolor{celllred} 818k & \cellcolor{celllred} 4.82e6 & \cellcolor{cellred} 2.94e7 \\ 
\method{MeanAct.}  & 10.6 & \cellcolor{celllgreen} 4175 & \cellcolor{celllgreen} 8689 & \cellcolor{celllgreen} 31.6k & \cellcolor{celllgreen} 42.8k & \cellcolor{celllgreen} 131k & \cellcolor{celllgreen} 140k & \cellcolor{celllred} 266k & \cellcolor{celllred} 332k & \cellcolor{cellred} 5.99e6 & \cellcolor{cellred} 6.28e6 & \cellcolor{celllred} 1.61e6 & \cellcolor{celllred} 675k & \cellcolor{celllgreen} 192k & \cellcolor{celllred} 721k & \cellcolor{celllgreen} 315k & \cellcolor{celllgreen} 253k & \cellcolor{celllgreen} 475k & \cellcolor{celllgreen} 3.01e6 \\ 
\method{LRP}       & 10.6 & \cellcolor{celllgreen} 71.3 & \cellcolor{celllgreen} 219 & \cellcolor{celllgreen} 321 & \cellcolor{celllgreen} 340 & \cellcolor{celllgreen} 637 & \cellcolor{celllgreen} 830 & \cellcolor{celllgreen} 1075 & \cellcolor{celllgreen} 1332 & \cellcolor{celllgreen} 1685 & \cellcolor{celllgreen} 2108 & \cellcolor{celllgreen} 2817 & \cellcolor{celllgreen} 5451 & \cellcolor{celllgreen} 22.1k & \cellcolor{celllgreen} 8581 & \cellcolor{celllgreen} 6923 & \cellcolor{cellred} 2.75e7 & \cellcolor{cellred} 1.74e7 & \cellcolor{celllred} 4.28e6 \\ 
\method{IG}        & 10.6 & \cellcolor{celllgreen} 13.3 & \cellcolor{celllgreen} 18.3 & \cellcolor{celllgreen} 26.1 & \cellcolor{celllgreen} 39.0 & \cellcolor{celllgreen} 60.0 & \cellcolor{celllgreen} 97.8 & \cellcolor{celllgreen} 180 & \cellcolor{celllgreen} 356 & \cellcolor{celllgreen} 724 & \cellcolor{celllgreen} 1275 & \cellcolor{celllgreen} 1966 & \cellcolor{celllgreen} 2945 & \cellcolor{cellgreen} 2824 & \cellcolor{cellgreen} 3928 & \cellcolor{cellgreen} 4311 & \cellcolor{cellgreen} 6140 & \cellcolor{cellgreen} 10.2k & \cellcolor{celllgreen} 37.2k \\ 
\consensus{}       & 10.6 & \cellcolor{cellgreen} 12.6 & \cellcolor{cellgreen} 16.2 & \cellcolor{cellgreen} 21.8 & \cellcolor{cellgreen} 30.0 & \cellcolor{cellgreen} 43.8 & \cellcolor{cellgreen} 66.1 & \cellcolor{cellgreen} 95.1 & \cellcolor{cellgreen} 143 & \cellcolor{cellgreen} 509 & \cellcolor{cellgreen} 738 & \cellcolor{cellgreen} 1364 & \cellcolor{cellgreen} 1996 & \cellcolor{celllgreen} 3194 & \cellcolor{celllgreen} 17.0k & \cellcolor{celllgreen} 187k & \cellcolor{celllgreen} 219k & \cellcolor{celllgreen} 52.8k & \cellcolor{cellgreen} 9477 \\ 
\midrule
\multicolumn{20}{l}{\textit{\textbf{MoRF} $\uparrow$}} \\
\midrule
\method{Random}    & 10.6 & \cellcolor{cellred} 101 & \cellcolor{cellred} 22.6k & \cellcolor{celllred} 131k & \cellcolor{cellred} 210k & \cellcolor{cellred} 190k & \cellcolor{cellred} 645k & \cellcolor{celllgreen} 1.50e6 & \cellcolor{celllgreen} 1.79e6 & \cellcolor{celllred} 785k & \cellcolor{celllred} 927k & \cellcolor{celllgreen} 812k & \cellcolor{celllgreen} 779k & \cellcolor{celllgreen} 561k & \cellcolor{celllgreen} 4.30e6 & \cellcolor{cellred} 396k & \cellcolor{celllgreen} 904k & \cellcolor{cellgreen} 2.07e6 & \cellcolor{cellgreen} 1.70e6 \\ 
\method{Magnitude} & 10.6 & \cellcolor{celllred} 21.8k & \cellcolor{celllred} 185k & \cellcolor{celllred} 139k & \cellcolor{celllred} 461k & \cellcolor{celllgreen} 1.62e6 & \cellcolor{celllgreen} 2.03e6 & \cellcolor{celllgreen} 1.46e6 & \cellcolor{celllgreen} 1.53e6 & \cellcolor{celllgreen} 2.24e6 & \cellcolor{celllgreen} 1.90e6 & \cellcolor{celllgreen} 1.16e6 & \cellcolor{celllgreen} 982k & \cellcolor{celllgreen} 725k & \cellcolor{celllgreen} 624k & \cellcolor{celllgreen} 509k & \cellcolor{cellgreen} 1.24e6 & \cellcolor{celllgreen} 511k & \cellcolor{celllred} 445k \\ 
\method{Wanda}     & 10.6 & \cellcolor{celllred} 38.1k & \cellcolor{celllred} 66.6k & \cellcolor{cellred} 81.0k & \cellcolor{celllred} 316k & \cellcolor{celllgreen} 1.17e6 & \cellcolor{celllred} 976k & \cellcolor{cellred} 273k & \cellcolor{celllred} 1.44e6 & \cellcolor{celllred} 501k & \cellcolor{cellred} 421k & \cellcolor{cellred} 270k & \cellcolor{cellred} 264k & \cellcolor{cellred} 448k & \cellcolor{celllgreen} 1.31e6 & \cellcolor{celllgreen} 967k & \cellcolor{celllgreen} 954k & \cellcolor{celllgreen} 726k & \cellcolor{cellred} 84.8k \\ 
\method{MeanAct.}  & 10.6 & \cellcolor{celllgreen} 715k & \cellcolor{celllgreen} 1.35e6 & \cellcolor{celllgreen} 1.35e6 & \cellcolor{celllgreen} 1.47e6 & \cellcolor{celllred} 838k & \cellcolor{celllred} 931k & \cellcolor{celllred} 1.26e6 & \cellcolor{celllgreen} 2.01e6 & \cellcolor{cellgreen} 2.75e6 & \cellcolor{cellgreen} 4.90e6 & \cellcolor{cellgreen} 2.84e6 & \cellcolor{cellgreen} 2.55e6 & \cellcolor{cellgreen} 1.79e6 & \cellcolor{cellgreen} 6.64e6 & \cellcolor{cellgreen} 1.14e6 & \cellcolor{celllgreen} 693k & \cellcolor{cellred} 132k & \cellcolor{celllred} 452k \\ 
\method{LRP}       & 10.6 & \cellcolor{celllgreen} 611k & \cellcolor{cellgreen} 5.12e6 & \cellcolor{celllgreen} 1.56e6 & \cellcolor{celllgreen} 3.92e6 & \cellcolor{cellgreen} 2.14e6 & \cellcolor{cellgreen} 8.32e6 & \cellcolor{celllgreen} 1.40e6 & \cellcolor{celllred} 1.25e6 & \cellcolor{celllgreen} 2.35e6 & \cellcolor{celllred} 696k & \cellcolor{celllred} 515k & \cellcolor{celllred} 502k & \cellcolor{celllred} 502k & \cellcolor{celllred} 502k & \cellcolor{celllgreen} 502k & \cellcolor{celllred} 502k & \cellcolor{celllgreen} 502k & \cellcolor{celllgreen} 502k \\ 
\method{IG}        & 10.6 & \cellcolor{celllgreen} 653k & \cellcolor{celllgreen} 1.08e6 & \cellcolor{celllgreen} 2.14e6 & \cellcolor{celllgreen} 1.58e6 & \cellcolor{celllred} 893k & \cellcolor{celllgreen} 1.40e6 & \cellcolor{cellgreen} 1.82e6 & \cellcolor{cellred} 1.12e6 & \cellcolor{celllgreen} 1.59e6 & \cellcolor{celllgreen} 1.49e6 & \cellcolor{celllgreen} 1.12e6 & \cellcolor{celllgreen} 939k & \cellcolor{celllgreen} 517k & \cellcolor{celllred} 502k & \cellcolor{celllgreen} 502k & \cellcolor{celllred} 502k & \cellcolor{celllgreen} 502k & \cellcolor{celllgreen} 502k \\ 
\consensus{}       & 10.6 & \cellcolor{cellgreen} 3.26e6 & \cellcolor{celllgreen} 2.22e6 & \cellcolor{cellgreen} 2.18e6 & \cellcolor{cellgreen} 4.80e6 & \cellcolor{celllgreen} 979k & \cellcolor{celllgreen} 1.54e6 & \cellcolor{celllred} 691k & \cellcolor{cellgreen} 2.12e6 & \cellcolor{cellred} 404k & \cellcolor{celllgreen} 1.58e6 & \cellcolor{celllred} 722k & \cellcolor{celllred} 493k & \cellcolor{celllred} 499k & \cellcolor{celllred} 502k & \cellcolor{celllgreen} 502k & \cellcolor{celllred} 502k & \cellcolor{celllgreen} 502k & \cellcolor{celllgreen} 502k \\ 
\bottomrule
\end{tabular}
\end{table*}

\begin{figure}[pb]
  \centering
  \includegraphics[width=\linewidth]{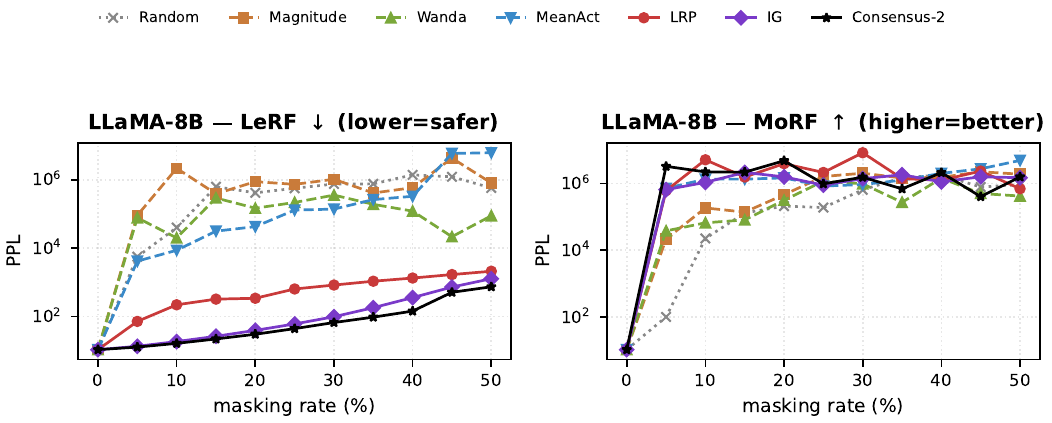}
  \caption{\textbf{LLaMA-3.1-8B: LeRF / MoRF degradation (companion to Table~\ref{tab:full_lerf_8b}).} Per-rate PPL for all seven selectors over $0$--$50\%$ masking (log $y$; the informative range before all selectors saturate at model collapse): LeRF (left, lower~$=$~safer choices) and MoRF (right, higher~$=$~better necessity identification). LRP, IG, and \consensus{} are much safer than the baselines on LeRF through the main low-to-moderate masking range, with a matching MoRF asymmetry at representative rates.}
  \label{fig:sweep_8b}
\end{figure}

% ---- Qwen-8B ----
\begin{table*}[ht]
\centering
\tiny
\setlength{\tabcolsep}{2pt}
\caption{Full LeRF / MoRF PPL: Qwen3-8B. Same format as Table~\ref{tab:full_lerf_1b}. Companion plot: Fig~\ref{fig:sweep_qwen}.}
\label{tab:full_lerf_qwen}
\begin{tabular}{l *{19}{r}}
\toprule
Selector & 0\% & 5\% & 10\% & 15\% & 20\% & 25\% & 30\% & 35\% & 40\% & 45\% & 50\% & 55\% & 60\% & 65\% & 70\% & 75\% & 80\% & 85\% & 90\% \\
\midrule
\multicolumn{20}{l}{\textit{\textbf{LeRF} $\downarrow$}} \\
\midrule
\method{Random}    & 18.6 & \cellcolor{celllgreen} 35.0 & \cellcolor{celllgreen} 373 & \cellcolor{celllgreen} 4354 & \cellcolor{celllred} 731k & \cellcolor{celllred} 839k & \cellcolor{celllred} 7.19e6 & \cellcolor{cellred} 5.35e7 & \cellcolor{cellred} 3.82e7 & \cellcolor{cellred} 2.91e7 & \cellcolor{cellred} 5.03e7 & \cellcolor{cellred} 3.87e7 & \cellcolor{cellred} 9.53e7 & \cellcolor{cellred} 6.19e7 & \cellcolor{cellred} 7.24e7 & \cellcolor{cellred} 8.69e7 & \cellcolor{cellred} 5.81e7 & \cellcolor{cellred} 6.13e7 & \cellcolor{cellred} 6.04e7 \\ 
\method{Magnitude} & 18.6 & \cellcolor{cellred} 17.2k & \cellcolor{cellred} 7.76e6 & \cellcolor{cellred} 1.28e7 & \cellcolor{cellred} 2.14e7 & \cellcolor{cellred} 1.29e7 & \cellcolor{celllred} 7.38e6 & \cellcolor{celllred} 4.09e6 & \cellcolor{celllred} 6.63e6 & \cellcolor{celllgreen} 6.84e6 & \cellcolor{celllred} 7.64e6 & \cellcolor{celllred} 1.21e7 & \cellcolor{celllred} 5.95e6 & \cellcolor{celllred} 6.82e6 & \cellcolor{celllred} 6.51e6 & \cellcolor{celllred} 1.02e7 & \cellcolor{celllred} 9.31e6 & \cellcolor{celllred} 6.75e6 & \cellcolor{celllred} 535k \\ 
\method{Wanda}     & 18.6 & \cellcolor{celllred} 1398 & \cellcolor{celllred} 67.5k & \cellcolor{celllred} 50.2k & \cellcolor{celllred} 69.6k & \cellcolor{celllred} 2.05e6 & \cellcolor{cellred} 1.95e7 & \cellcolor{celllred} 6.74e6 & \cellcolor{celllred} 9.33e6 & \cellcolor{celllred} 1.53e7 & \cellcolor{celllred} 4.17e6 & \cellcolor{celllgreen} 1.62e6 & \cellcolor{celllred} 2.26e7 & \cellcolor{celllred} 1.24e7 & \cellcolor{celllred} 993k & \cellcolor{celllred} 551k & \cellcolor{celllgreen} 709k & \cellcolor{celllred} 1.97e6 & \cellcolor{celllgreen} 134k \\ 
\method{MeanAct.}  & 18.6 & \cellcolor{celllred} 304 & \cellcolor{celllred} 18.6k & \cellcolor{celllred} 81.7k & \cellcolor{celllgreen} 58.3k & \cellcolor{celllgreen} 170k & \cellcolor{celllgreen} 65.9k & \cellcolor{celllgreen} 207k & \cellcolor{celllgreen} 487k & \cellcolor{celllred} 1.43e7 & \cellcolor{celllgreen} 1.84e6 & \cellcolor{celllred} 3.99e6 & \cellcolor{celllgreen} 558k & \cellcolor{celllgreen} 151k & \cellcolor{celllgreen} 61.4k & \cellcolor{celllgreen} 130k & \cellcolor{celllred} 1.57e6 & \cellcolor{cellgreen} 21.3k & \cellcolor{celllgreen} 100k \\ 
\method{LRP}       & 18.6 & \cellcolor{cellgreen} 19.6 & \cellcolor{celllgreen} 39.9 & \cellcolor{celllgreen} 84.0 & \cellcolor{celllgreen} 181 & \cellcolor{celllgreen} 340 & \cellcolor{celllgreen} 724 & \cellcolor{celllgreen} 1189 & \cellcolor{celllgreen} 2243 & \cellcolor{celllgreen} 3072 & \cellcolor{celllgreen} 3356 & \cellcolor{celllgreen} 6262 & \cellcolor{celllgreen} 16.7k & \cellcolor{celllgreen} 17.9k & \cellcolor{celllgreen} 71.1k & \cellcolor{celllgreen} 193k & \cellcolor{celllgreen} 88.5k & \cellcolor{celllgreen} 124k & \cellcolor{cellgreen} 24.7k \\ 
\method{IG}        & 18.6 & \cellcolor{celllgreen} 22.3 & \cellcolor{celllgreen} 33.0 & \cellcolor{celllgreen} 48.9 & \cellcolor{celllgreen} 76.5 & \cellcolor{celllgreen} 124 & \cellcolor{celllgreen} 229 & \cellcolor{celllgreen} 436 & \cellcolor{celllgreen} 772 & \cellcolor{celllgreen} 1601 & \cellcolor{celllgreen} 3074 & \cellcolor{cellgreen} 5918 & \cellcolor{cellgreen} 11.2k & \cellcolor{celllgreen} 17.0k & \cellcolor{celllgreen} 23.8k & \cellcolor{cellgreen} 31.6k & \cellcolor{celllgreen} 224k & \cellcolor{celllgreen} 1.32e6 & \cellcolor{celllred} 603k \\ 
\consensus{}       & 18.6 & \cellcolor{celllgreen} 20.3 & \cellcolor{cellgreen} 28.8 & \cellcolor{cellgreen} 44.9 & \cellcolor{cellgreen} 69.1 & \cellcolor{cellgreen} 113 & \cellcolor{cellgreen} 187 & \cellcolor{cellgreen} 403 & \cellcolor{cellgreen} 642 & \cellcolor{cellgreen} 953 & \cellcolor{cellgreen} 1378 & \cellcolor{celllgreen} 8616 & \cellcolor{celllgreen} 13.6k & \cellcolor{cellgreen} 9766 & \cellcolor{cellgreen} 14.7k & \cellcolor{celllgreen} 116k & \cellcolor{cellgreen} 36.6k & \cellcolor{celllgreen} 104k & \cellcolor{celllgreen} 117k \\ 
\midrule
\multicolumn{20}{l}{\textit{\textbf{MoRF} $\uparrow$}} \\
\midrule
\method{Random}    & 18.6 & \cellcolor{cellred} 90.8 & \cellcolor{cellred} 3240 & \cellcolor{cellred} 97.8k & \cellcolor{cellred} 413k & \cellcolor{cellred} 2.11e6 & \cellcolor{cellred} 2.09e6 & \cellcolor{celllred} 7.51e6 & \cellcolor{celllred} 1.14e7 & \cellcolor{celllred} 6.01e6 & \cellcolor{celllred} 1.52e7 & \cellcolor{celllred} 2.47e7 & \cellcolor{celllred} 1.91e7 & \cellcolor{celllred} 4.74e7 & \cellcolor{celllred} 7.07e7 & \cellcolor{celllred} 2.66e7 & \cellcolor{celllred} 5.98e7 & \cellcolor{celllgreen} 8.94e7 & \cellcolor{celllred} 2.67e7 \\ 
\method{Magnitude} & 18.6 & \cellcolor{celllgreen} 1.09e6 & \cellcolor{celllgreen} 1.35e7 & \cellcolor{celllred} 2.04e7 & \cellcolor{celllred} 1.40e7 & \cellcolor{celllgreen} 2.80e8 & \cellcolor{cellgreen} 1.66e9 & \cellcolor{cellgreen} 9.40e9 & \cellcolor{cellgreen} 6.20e10 & \cellcolor{cellgreen} 1.78e10 & \cellcolor{cellgreen} 5.67e12 & \cellcolor{cellgreen} 3.17e11 & \cellcolor{cellgreen} 1.51e12 & \cellcolor{cellgreen} 4.54e10 & \cellcolor{cellgreen} 8.80e17 & \cellcolor{cellgreen} 9.80e23 & \cellcolor{cellgreen} 9.07e16 & \cellcolor{cellgreen} 3.44e11 & \cellcolor{cellgreen} 4.48e10 \\ 
\method{Wanda}     & 18.6 & \cellcolor{celllred} 6857 & \cellcolor{celllred} 316k & \cellcolor{celllred} 2.65e7 & \cellcolor{celllred} 3.68e7 & \cellcolor{celllgreen} 5.01e7 & \cellcolor{celllred} 9.67e6 & \cellcolor{celllgreen} 3.14e7 & \cellcolor{celllgreen} 1.40e8 & \cellcolor{celllgreen} 3.30e8 & \cellcolor{celllgreen} 5.36e8 & \cellcolor{celllgreen} 1.07e8 & \cellcolor{celllgreen} 2.51e8 & \cellcolor{celllgreen} 2.08e8 & \cellcolor{celllgreen} 1.88e8 & \cellcolor{celllgreen} 2.66e8 & \cellcolor{celllgreen} 6.29e7 & \cellcolor{celllred} 4.48e7 & \cellcolor{celllred} 8.66e6 \\ 
\method{MeanAct.}  & 18.6 & \cellcolor{celllred} 57.6k & \cellcolor{celllred} 5.64e6 & \cellcolor{celllgreen} 2.91e7 & \cellcolor{celllgreen} 4.25e7 & \cellcolor{cellgreen} 4.07e8 & \cellcolor{celllgreen} 3.11e8 & \cellcolor{celllgreen} 6.08e8 & \cellcolor{celllgreen} 3.98e8 & \cellcolor{celllgreen} 2.01e8 & \cellcolor{celllgreen} 2.36e8 & \cellcolor{celllgreen} 7.82e7 & \cellcolor{celllgreen} 9.72e7 & \cellcolor{celllgreen} 7.57e7 & \cellcolor{celllgreen} 1.47e8 & \cellcolor{celllgreen} 1.14e8 & \cellcolor{celllgreen} 6.74e7 & \cellcolor{celllgreen} 7.18e7 & \cellcolor{celllgreen} 4.10e7 \\ 
\method{LRP}       & 18.6 & \cellcolor{celllgreen} 1.29e9 & \cellcolor{celllgreen} 2.87e8 & \cellcolor{celllgreen} 3.16e7 & \cellcolor{cellgreen} 6.12e7 & \cellcolor{celllgreen} 1.07e8 & \cellcolor{celllgreen} 3.74e7 & \cellcolor{cellred} 4.43e6 & \cellcolor{cellred} 2.31e6 & \cellcolor{cellred} 445k & \cellcolor{cellred} 171k & \cellcolor{cellred} 152k & \cellcolor{cellred} 151k & \cellcolor{cellred} 151k & \cellcolor{cellred} 151k & \cellcolor{cellred} 152k & \cellcolor{cellred} 151k & \cellcolor{cellred} 151k & \cellcolor{cellred} 152k \\ 
\method{IG}        & 18.6 & \cellcolor{cellgreen} 1.10e11 & \cellcolor{cellgreen} 1.52e9 & \cellcolor{cellgreen} 3.08e8 & \cellcolor{celllgreen} 5.51e7 & \cellcolor{celllred} 3.13e7 & \cellcolor{celllred} 2.68e7 & \cellcolor{celllgreen} 2.36e7 & \cellcolor{celllred} 1.80e7 & \cellcolor{celllgreen} 2.24e7 & \cellcolor{celllred} 3.19e7 & \cellcolor{celllred} 2.78e7 & \cellcolor{celllred} 4.10e7 & \cellcolor{celllred} 3.07e7 & \cellcolor{celllred} 4.06e7 & \cellcolor{celllred} 4.01e7 & \cellcolor{celllred} 3.93e7 & \cellcolor{celllred} 4.59e7 & \cellcolor{celllgreen} 3.48e7 \\ 
\consensus{}       & 18.6 & \cellcolor{celllgreen} 3.32e9 & \cellcolor{celllgreen} 1.71e8 & \cellcolor{celllgreen} 4.07e7 & \cellcolor{celllgreen} 3.71e7 & \cellcolor{celllred} 3.48e7 & \cellcolor{celllgreen} 2.83e7 & \cellcolor{celllred} 2.25e7 & \cellcolor{celllgreen} 3.16e7 & \cellcolor{celllred} 2.11e7 & \cellcolor{celllgreen} 4.80e7 & \cellcolor{celllgreen} 3.57e7 & \cellcolor{celllgreen} 4.83e7 & \cellcolor{celllgreen} 1.59e8 & \cellcolor{celllgreen} 1.59e8 & \cellcolor{celllgreen} 1.59e8 & \cellcolor{celllgreen} 1.59e8 & \cellcolor{celllgreen} 1.59e8 & \cellcolor{celllgreen} 1.59e8 \\ 
\bottomrule
\end{tabular}
\end{table*}

\begin{figure}[hb]
  \centering
  \includegraphics[width=\linewidth]{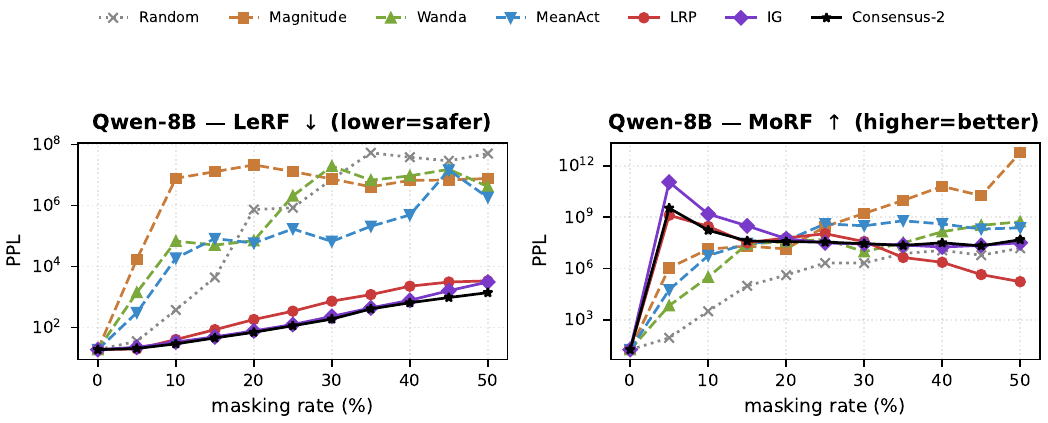}
  \caption{\textbf{Qwen3-8B: LeRF / MoRF degradation (companion to Table~\ref{tab:full_lerf_qwen}).} Per-rate PPL for all seven selectors over $0$--$50\%$ masking (log $y$; the informative range before all selectors saturate at model collapse): LeRF (left, lower~$=$~safer choices) and MoRF (right, higher~$=$~better necessity identification). Attribution selectors degrade gracefully on LeRF while the baselines collapse early; here even \method{Random} inverts (MoRF $<$ LeRF), not just \method{Wanda}.}
  \label{fig:sweep_qwen}
\end{figure}
\clearpage
% ---- Gemma-12B ----
\begin{table*}[ht]
\centering
\tiny
\setlength{\tabcolsep}{2pt}
\caption{Full LeRF / MoRF PPL: Gemma-3-12B. Same format as Table~\ref{tab:full_lerf_1b}. Companion plot: Fig~\ref{fig:sweep_gemma}.}
\label{tab:full_lerf_gemma}
\begin{tabular}{l *{19}{r}}
\toprule
Selector & 0\% & 5\% & 10\% & 15\% & 20\% & 25\% & 30\% & 35\% & 40\% & 45\% & 50\% & 55\% & 60\% & 65\% & 70\% & 75\% & 80\% & 85\% & 90\% \\
\midrule
\multicolumn{20}{l}{\textit{\textbf{LeRF} $\downarrow$}} \\
\midrule
\method{Random}    & 19.0 & \cellcolor{celllred} 34.5 & \cellcolor{celllred} 204 & \cellcolor{celllred} 55.2k & \cellcolor{celllred} 477k & \cellcolor{celllred} 1.18e8 & \cellcolor{cellred} 3.16e9 & \cellcolor{cellred} 4.38e9 & \cellcolor{cellred} 1.43e10 & \cellcolor{cellred} 1.31e11 & \cellcolor{celllred} 3.77e11 & \cellcolor{celllred} 7.32e11 & \cellcolor{celllred} 5.86e12 & \cellcolor{cellred} 4.12e13 & \cellcolor{cellred} 4.38e14 & \cellcolor{cellred} 9.86e14 & \cellcolor{cellred} 3.19e15 & \cellcolor{cellred} 1.17e15 & \cellcolor{cellred} 1.03e16 \\ 
\method{Magnitude} & 19.0 & \cellcolor{cellred} 186k & \cellcolor{cellred} 7.05e6 & \cellcolor{cellred} 6.91e8 & \cellcolor{celllgreen} 74.8k & \cellcolor{celllgreen} 124k & \cellcolor{celllgreen} 317k & \cellcolor{celllred} 534k & \cellcolor{celllred} 253k & \cellcolor{celllgreen} 456k & \cellcolor{celllred} 649k & \cellcolor{celllgreen} 416k & \cellcolor{celllgreen} 372k & \cellcolor{celllgreen} 313k & \cellcolor{celllred} 422k & \cellcolor{celllgreen} 223k & \cellcolor{celllgreen} 215k & \cellcolor{celllgreen} 210k & \cellcolor{celllgreen} 656k \\ 
\method{Wanda}     & 19.0 & \cellcolor{cellgreen} 19.0 & \cellcolor{cellgreen} 19.0 & \cellcolor{celllred} 15.1k & \cellcolor{celllred} 7.88e6 & \cellcolor{celllred} 3.94e7 & \cellcolor{celllred} 907k & \cellcolor{celllgreen} 154k & \cellcolor{celllgreen} 113k & \cellcolor{celllred} 8.19e6 & \cellcolor{celllgreen} 305k & \cellcolor{celllred} 3.27e7 & \cellcolor{celllred} 1.64e6 & \cellcolor{celllred} 638k & \cellcolor{celllgreen} 355k & \cellcolor{celllred} 616k & \cellcolor{celllred} 232k & \cellcolor{celllred} 252k & \cellcolor{celllred} 1.14e6 \\ 
\method{MeanAct.}  & 19.0 & \cellcolor{cellgreen} 19.0 & \cellcolor{cellgreen} 19.0 & \cellcolor{celllgreen} 680 & \cellcolor{cellred} 3.87e13 & \cellcolor{cellred} 5.35e13 & \cellcolor{celllred} 1.88e8 & \cellcolor{celllred} 3.68e9 & \cellcolor{celllred} 2.06e7 & \cellcolor{celllred} 4.39e8 & \cellcolor{cellred} 2.43e17 & \cellcolor{cellred} 3.36e18 & \cellcolor{cellred} 1.78e13 & \cellcolor{celllred} 8.45e6 & \cellcolor{celllred} 3.39e9 & \cellcolor{celllgreen} 491k & \cellcolor{celllred} 1.46e6 & \cellcolor{celllred} 778k & \cellcolor{celllgreen} 131k \\ 
\method{LRP}       & 19.0 & \cellcolor{celllgreen} 26.1 & \cellcolor{celllgreen} 36.3 & \cellcolor{celllgreen} 71.0 & \cellcolor{celllgreen} 183 & \cellcolor{celllgreen} 393 & \cellcolor{celllgreen} 514 & \cellcolor{celllgreen} 764 & \cellcolor{celllgreen} 1034 & \cellcolor{celllgreen} 1286 & \cellcolor{celllgreen} 1522 & \cellcolor{celllgreen} 1912 & \cellcolor{celllgreen} 3534 & \cellcolor{celllgreen} 4712 & \cellcolor{celllgreen} 7847 & \cellcolor{celllgreen} 10.2k & \cellcolor{celllgreen} 96.8k & \cellcolor{cellgreen} 11.1k & \cellcolor{cellgreen} 6007 \\ 
\method{IG}        & 19.0 & \cellcolor{celllred} 45.1 & \cellcolor{celllred} 57.1 & \cellcolor{celllgreen} 69.0 & \cellcolor{cellgreen} 70.8 & \cellcolor{cellgreen} 81.4 & \cellcolor{cellgreen} 135 & \cellcolor{celllgreen} 215 & \cellcolor{celllgreen} 418 & \cellcolor{celllgreen} 776 & \cellcolor{celllgreen} 1256 & \cellcolor{celllgreen} 1982 & \cellcolor{celllgreen} 3480 & \cellcolor{celllgreen} 6117 & \cellcolor{celllgreen} 30.7k & \cellcolor{celllred} 2.67e6 & \cellcolor{celllgreen} 115k & \cellcolor{celllgreen} 160k & \cellcolor{celllred} 4.85e11 \\ 
\consensus{}       & 19.0 & \cellcolor{celllgreen} 30.2 & \cellcolor{celllgreen} 46.1 & \cellcolor{cellgreen} 60.2 & \cellcolor{celllgreen} 81.5 & \cellcolor{celllgreen} 112 & \cellcolor{celllgreen} 141 & \cellcolor{cellgreen} 178 & \cellcolor{cellgreen} 259 & \cellcolor{cellgreen} 449 & \cellcolor{cellgreen} 823 & \cellcolor{cellgreen} 1024 & \cellcolor{cellgreen} 2640 & \cellcolor{cellgreen} 2713 & \cellcolor{cellgreen} 4288 & \cellcolor{cellgreen} 6602 & \cellcolor{cellgreen} 8825 & \cellcolor{celllgreen} 46.9k & \cellcolor{celllgreen} 26.1k \\ 
\midrule
\multicolumn{20}{l}{\textit{\textbf{MoRF} $\uparrow$}} \\
\midrule
\method{Random}    & 19.0 & \cellcolor{cellred} 44.2 & \cellcolor{cellred} 1567 & \cellcolor{cellred} 2.38e6 & \cellcolor{cellred} 1.86e6 & \cellcolor{cellred} 1.80e6 & \cellcolor{celllred} 1.48e8 & \cellcolor{celllgreen} 6.43e9 & \cellcolor{celllgreen} 3.58e11 & \cellcolor{celllgreen} 5.07e13 & \cellcolor{celllgreen} 3.04e14 & \cellcolor{celllred} 2.16e12 & \cellcolor{celllred} 1.78e12 & \cellcolor{celllgreen} 1.41e14 & \cellcolor{celllgreen} 1.98e14 & \cellcolor{cellgreen} 3.81e15 & \cellcolor{celllgreen} 3.95e14 & \cellcolor{celllgreen} 2.23e15 & \cellcolor{celllgreen} 1.39e15 \\ 
\method{Magnitude} & 19.0 & \cellcolor{celllgreen} 9.43e8 & \cellcolor{celllgreen} 8.21e9 & \cellcolor{celllred} 1.91e8 & \cellcolor{celllred} 7.30e8 & \cellcolor{celllgreen} 1.24e11 & \cellcolor{celllred} 1.40e8 & \cellcolor{celllred} 2.13e9 & \cellcolor{celllred} 4.28e10 & \cellcolor{celllred} 1.70e10 & \cellcolor{celllred} 5.52e8 & \cellcolor{celllred} 3.35e9 & \cellcolor{celllred} 1.29e11 & \cellcolor{celllred} 1.38e7 & \cellcolor{celllred} 1.56e8 & \cellcolor{celllred} 4.41e11 & \cellcolor{celllred} 4.73e8 & \cellcolor{celllgreen} 3.10e17 & \cellcolor{celllgreen} 2.91e23 \\ 
\method{Wanda}     & 19.0 & \cellcolor{celllred} 28.7k & \cellcolor{celllred} 895k & \cellcolor{celllred} 1.41e9 & \cellcolor{celllgreen} 9.66e8 & \cellcolor{celllred} 7.38e8 & \cellcolor{celllgreen} 1.66e9 & \cellcolor{celllred} 3.76e9 & \cellcolor{celllred} 1.79e10 & \cellcolor{celllred} 2.63e10 & \cellcolor{celllred} 5.34e8 & \cellcolor{celllgreen} 1.07e13 & \cellcolor{celllgreen} 1.69e14 & \cellcolor{celllred} 1.41e12 & \cellcolor{celllred} 3.28e10 & \cellcolor{celllred} 2.67e12 & \cellcolor{celllred} 1.14e11 & \cellcolor{celllred} 2.17e14 & \cellcolor{cellgreen} 4.56e241 \\ 
\method{MeanAct.}  & 19.0 & \cellcolor{celllred} 1.34e7 & \cellcolor{celllred} 6.50e8 & \cellcolor{celllgreen} 9.27e9 & \cellcolor{celllgreen} 5.64e10 & \cellcolor{celllgreen} 4.38e10 & \cellcolor{celllgreen} 4.27e10 & \cellcolor{celllgreen} 3.82e11 & \cellcolor{celllgreen} 1.10e11 & \cellcolor{celllgreen} 1.56e13 & \cellcolor{celllgreen} 4.33e13 & \cellcolor{celllgreen} 1.34e15 & \cellcolor{celllgreen} 4.04e14 & \cellcolor{cellgreen} 4.08e22 & \cellcolor{celllgreen} 5.60e13 & \cellcolor{celllgreen} 3.33e14 & \cellcolor{cellgreen} 3.87e20 & \cellcolor{cellgreen} 4.56e241 & \cellcolor{cellgreen} 4.56e241 \\ 
\method{LRP}       & 19.0 & \cellcolor{celllgreen} 6.03e14 & \cellcolor{celllgreen} 3.43e14 & \cellcolor{celllgreen} 7.96e13 & \cellcolor{celllred} 7.08e8 & \cellcolor{celllred} 6.30e6 & \cellcolor{cellred} 2.40e6 & \cellcolor{cellred} 2.81e6 & \cellcolor{cellred} 1.04e6 & \cellcolor{cellred} 434k & \cellcolor{cellred} 272k & \cellcolor{cellred} 294k & \cellcolor{cellred} 253k & \cellcolor{cellred} 374k & \cellcolor{cellred} 188k & \cellcolor{cellred} 193k & \cellcolor{cellred} 184k & \cellcolor{cellred} 193k & \cellcolor{cellred} 1.56e6 \\ 
\method{IG}        & 19.0 & \cellcolor{cellgreen} 8.15e14 & \cellcolor{cellgreen} 1.76e15 & \cellcolor{cellgreen} 5.24e14 & \cellcolor{cellgreen} 1.81e15 & \cellcolor{cellgreen} 2.37e15 & \cellcolor{cellgreen} 1.00e14 & \cellcolor{celllgreen} 6.39e13 & \cellcolor{cellgreen} 2.20e14 & \cellcolor{cellgreen} 7.02e15 & \cellcolor{cellgreen} 6.79e14 & \cellcolor{cellgreen} 3.71e15 & \cellcolor{cellgreen} 1.65e15 & \cellcolor{celllgreen} 2.47e14 & \cellcolor{cellgreen} 6.89e14 & \cellcolor{celllgreen} 8.49e14 & \cellcolor{celllgreen} 1.10e15 & \cellcolor{celllred} 2.27e14 & \cellcolor{celllred} 1.44e14 \\ 
\consensus{}       & 19.0 & \cellcolor{celllgreen} 3.68e14 & \cellcolor{celllgreen} 1.60e15 & \cellcolor{celllgreen} 9.06e13 & \cellcolor{celllgreen} 1.18e14 & \cellcolor{celllgreen} 4.75e13 & \cellcolor{celllgreen} 5.26e13 & \cellcolor{cellgreen} 1.99e14 & \cellcolor{celllgreen} 8.50e11 & \cellcolor{celllgreen} 1.19e15 & \cellcolor{celllgreen} 1.39e14 & \cellcolor{celllgreen} 6.05e14 & \cellcolor{celllgreen} 1.81e14 & \cellcolor{celllgreen} 4.48e13 & \cellcolor{celllgreen} 5.96e14 & \cellcolor{celllgreen} 1.38e14 & \cellcolor{celllgreen} 1.90e14 & \cellcolor{celllgreen} 4.38e14 & \cellcolor{celllred} 8.02e12 \\ 
\bottomrule
\end{tabular}
\end{table*}

\begin{figure}[pb]
  \centering
  \includegraphics[width=\linewidth]{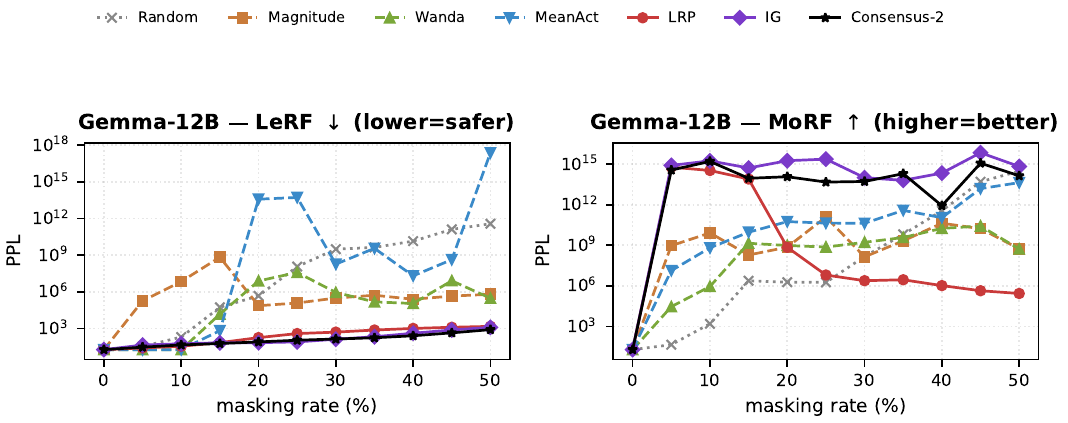}
  \caption{\textbf{Gemma-3-12B: LeRF / MoRF degradation (companion to Table~\ref{tab:full_lerf_gemma}).} Per-rate PPL for all
    seven selectors over $0$--$50\%$ masking (log $y$; the informative range before all selectors saturate at model collapse):
    LeRF (left, lower~$=$~safer choices) and MoRF (right, higher~$=$~better necessity identification). Attribution selectors
    dominate the moderate-rate LeRF regime; IG reaches the highest $30\%$ MoRF in the audit (${\sim}10^{14}$), while
    \method{Wanda}/\method{MeanAct.} still match dense PPL at $5\%$.}
  \label{fig:sweep_gemma}
\end{figure}

\begin{table*}[ht]
\centering
\scriptsize
\caption{Downstream accuracy proportions at LeRF masking rates 0/10/20/30/50/70/90\%
  (LLaMA-8B, 0-shot). Representative selectors shown.}
\label{tab:downstream_full_8b}
\begin{tabular}{l l cccccccc r}
\toprule
Selector & Rate & ARC-E & ARC-C & HS & WG & PIQA & BoolQ & LAM & OBQA & Avg \\
\midrule
Dense & 0\% & .822 & .532 & .544 & .746 & .798 & .824 & .766 & .338 & .671 \\ 
\midrule
\multirow{6}{*}{\method{Random}}
  & 10\% & \cellcolor{celllgreen} .246 & \cellcolor{cellgreen} .232 & \cellcolor{celllgreen} .260 & \cellcolor{celllred} .498 & \cellcolor{celllred} .514 & \cellcolor{celllgreen} .630 & \cellcolor{cellgreen} .000 & \cellcolor{cellred} .126 & \cellcolor{celllgreen} .313 \\ 
  & 20\% & \cellcolor{cellred} .234 & \cellcolor{celllred} .208 & \cellcolor{celllred} .250 & \cellcolor{celllgreen} .500 & \cellcolor{celllred} .510 & \cellcolor{cellgreen} .632 & \cellcolor{cellgreen} .000 & \cellcolor{celllred} .150 & \cellcolor{celllred} .310 \\ 
  & 30\% & \cellcolor{celllred} .242 & \cellcolor{celllgreen} .226 & \cellcolor{celllgreen} .260 & \cellcolor{celllgreen} .504 & \cellcolor{cellred} .508 & \cellcolor{celllred} .620 & \cellcolor{cellgreen} .000 & \cellcolor{cellgreen} .166 & \cellcolor{celllgreen} .316 \\ 
  & 50\% & \cellcolor{celllred} .242 & \cellcolor{celllred} .212 & \cellcolor{cellgreen} .262 & \cellcolor{cellred} .492 & \cellcolor{celllgreen} .534 & \cellcolor{celllred} .468 & \cellcolor{cellgreen} .000 & \cellcolor{celllred} .148 & \cellcolor{celllred} .295 \\ 
  & 70\% & \cellcolor{cellgreen} .272 & \cellcolor{celllgreen} .224 & \cellcolor{celllred} .254 & \cellcolor{celllred} .498 & \cellcolor{cellgreen} .536 & \cellcolor{celllgreen} .624 & \cellcolor{cellgreen} .000 & \cellcolor{celllgreen} .158 & \cellcolor{cellgreen} .321 \\ 
  & 90\% & \cellcolor{celllgreen} .250 & \cellcolor{cellred} .204 & \cellcolor{cellred} .240 & \cellcolor{cellgreen} .544 & \cellcolor{celllgreen} .526 & \cellcolor{cellred} .428 & \cellcolor{cellgreen} .000 & \cellcolor{celllgreen} .158 & \cellcolor{cellred} .294 \\ 
\midrule
\multirow{6}{*}{\method{LRP}}
  & 10\% & \cellcolor{cellgreen} .450 & \cellcolor{celllred} .172 & \cellcolor{celllgreen} .298 & \cellcolor{celllred} .496 & \cellcolor{cellgreen} .592 & \cellcolor{celllgreen} .370 & \cellcolor{cellgreen} .086 & \cellcolor{celllgreen} .136 & \cellcolor{cellgreen} .325 \\ 
  & 20\% & \cellcolor{celllgreen} .318 & \cellcolor{celllred} .168 & \cellcolor{cellgreen} .306 & \cellcolor{celllgreen} .506 & \cellcolor{celllgreen} .568 & \cellcolor{celllgreen} .370 & \cellcolor{celllgreen} .042 & \cellcolor{celllgreen} .130 & \cellcolor{celllgreen} .301 \\ 
  & 30\% & \cellcolor{celllgreen} .308 & \cellcolor{cellred} .160 & \cellcolor{celllgreen} .292 & \cellcolor{celllgreen} .500 & \cellcolor{celllgreen} .564 & \cellcolor{cellgreen} .378 & \cellcolor{celllgreen} .008 & \cellcolor{celllred} .124 & \cellcolor{celllgreen} .292 \\ 
  & 50\% & \cellcolor{celllred} .260 & \cellcolor{celllgreen} .176 & \cellcolor{celllred} .280 & \cellcolor{cellgreen} .520 & \cellcolor{celllred} .520 & \cellcolor{celllgreen} .370 & \cellcolor{celllred} .000 & \cellcolor{celllred} .118 & \cellcolor{celllred} .281 \\ 
  & 70\% & \cellcolor{celllred} .252 & \cellcolor{celllgreen} .182 & \cellcolor{celllred} .272 & \cellcolor{celllred} .496 & \cellcolor{celllred} .522 & \cellcolor{celllgreen} .370 & \cellcolor{celllred} .000 & \cellcolor{cellred} .112 & \cellcolor{cellred} .276 \\ 
  & 90\% & \cellcolor{cellred} .244 & \cellcolor{cellgreen} .244 & \cellcolor{cellred} .264 & \cellcolor{cellred} .482 & \cellcolor{cellred} .502 & \cellcolor{celllgreen} .370 & \cellcolor{celllred} .000 & \cellcolor{cellgreen} .186 & \cellcolor{celllred} .286 \\ 
\midrule
\multirow{6}{*}{\method{IG}}
  & 10\% & \cellcolor{cellgreen} .710 & \cellcolor{cellgreen} .422 & \cellcolor{cellgreen} .486 & \cellcolor{cellgreen} .708 & \cellcolor{cellgreen} .754 & \cellcolor{cellgreen} .730 & \cellcolor{cellgreen} .586 & \cellcolor{cellgreen} .276 & \cellcolor{cellgreen} .584 \\ 
  & 20\% & \cellcolor{celllgreen} .576 & \cellcolor{celllgreen} .324 & \cellcolor{celllgreen} .424 & \cellcolor{celllgreen} .632 & \cellcolor{celllgreen} .690 & \cellcolor{celllgreen} .664 & \cellcolor{celllgreen} .330 & \cellcolor{celllgreen} .204 & \cellcolor{celllgreen} .480 \\ 
  & 30\% & \cellcolor{celllgreen} .446 & \cellcolor{celllgreen} .254 & \cellcolor{celllgreen} .342 & \cellcolor{celllgreen} .584 & \cellcolor{celllgreen} .608 & \cellcolor{celllgreen} .594 & \cellcolor{celllgreen} .056 & \cellcolor{celllgreen} .176 & \cellcolor{celllgreen} .383 \\ 
  & 50\% & \cellcolor{celllred} .272 & \cellcolor{cellred} .212 & \cellcolor{celllred} .292 & \cellcolor{celllred} .472 & \cellcolor{cellred} .510 & \cellcolor{celllred} .422 & \cellcolor{celllred} .000 & \cellcolor{celllred} .140 & \cellcolor{celllred} .290 \\ 
  & 70\% & \cellcolor{celllred} .248 & \cellcolor{celllred} .224 & \cellcolor{celllred} .272 & \cellcolor{celllred} .496 & \cellcolor{celllred} .544 & \cellcolor{cellred} .372 & \cellcolor{celllred} .000 & \cellcolor{cellred} .132 & \cellcolor{celllred} .286 \\ 
  & 90\% & \cellcolor{cellred} .230 & \cellcolor{celllred} .214 & \cellcolor{cellred} .270 & \cellcolor{cellred} .454 & \cellcolor{celllred} .516 & \cellcolor{celllred} .400 & \cellcolor{celllred} .000 & \cellcolor{celllred} .146 & \cellcolor{cellred} .279 \\ 
\midrule
\multirow{6}{*}{\consensus{}}
  & 10\% & \cellcolor{cellgreen} .724 & \cellcolor{cellgreen} .416 & \cellcolor{cellgreen} .494 & \cellcolor{cellgreen} .696 & \cellcolor{cellgreen} .766 & \cellcolor{cellgreen} .740 & \cellcolor{cellgreen} .528 & \cellcolor{cellgreen} .274 & \cellcolor{cellgreen} .580 \\ 
  & 20\% & \cellcolor{celllgreen} .598 & \cellcolor{celllgreen} .276 & \cellcolor{celllgreen} .408 & \cellcolor{celllgreen} .602 & \cellcolor{celllgreen} .674 & \cellcolor{celllgreen} .392 & \cellcolor{celllgreen} .296 & \cellcolor{celllgreen} .214 & \cellcolor{celllgreen} .432 \\ 
  & 30\% & \cellcolor{celllgreen} .472 & \cellcolor{celllgreen} .244 & \cellcolor{celllgreen} .348 & \cellcolor{celllgreen} .560 & \cellcolor{celllgreen} .634 & \cellcolor{celllgreen} .632 & \cellcolor{celllgreen} .092 & \cellcolor{celllgreen} .156 & \cellcolor{celllgreen} .392 \\ 
  & 50\% & \cellcolor{celllred} .260 & \cellcolor{cellred} .180 & \cellcolor{celllred} .282 & \cellcolor{celllred} .522 & \cellcolor{celllred} .528 & \cellcolor{cellred} .370 & \cellcolor{celllred} .000 & \cellcolor{celllred} .130 & \cellcolor{celllred} .284 \\ 
  & 70\% & \cellcolor{celllred} .244 & \cellcolor{celllred} .186 & \cellcolor{celllred} .280 & \cellcolor{celllred} .506 & \cellcolor{cellred} .508 & \cellcolor{celllred} .372 & \cellcolor{celllred} .000 & \cellcolor{cellred} .110 & \cellcolor{celllred} .276 \\ 
  & 90\% & \cellcolor{cellred} .236 & \cellcolor{celllred} .186 & \cellcolor{celllred} .280 & \cellcolor{cellred} .486 & \cellcolor{celllred} .520 & \cellcolor{celllred} .380 & \cellcolor{celllred} .000 & \cellcolor{celllred} .120 & \cellcolor{celllred} .276 \\ 
\bottomrule
\end{tabular}
\end{table*}

%  (Full per-task downstream accuracy table for LLaMA-3B follows the same
%   format as Table~\ref{tab:downstream_full_8b} and will be released with
%   the anonymised artifacts upon de-anonymisation.)

\begin{table*}[ht]
\centering
\scriptsize
\caption{Per-task downstream accuracy proportions at LeRF 30\% masking (4 of 5
  models; LLaMA-3B in the released artifacts). All 7 selectors. Dense
  baseline shown for reference.}
\label{tab:downstream_30pct}
\begin{tabular}{l l cccccccc r}
\toprule
Model & Selector & ARC-E & ARC-C & HS & WG & PIQA & BoolQ & LAM & OBQA & Avg \\
\midrule
\multirow{8}{*}{LLaMA-1B}
  & Dense          & \cellcolor{cellgreen} .634 & \cellcolor{cellgreen} .300 & \cellcolor{cellgreen} .460 & \cellcolor{cellgreen} .584 & \cellcolor{cellgreen} .754 & \cellcolor{cellgreen} .650 & \cellcolor{cellgreen} .608 & \cellcolor{cellgreen} .278 & \cellcolor{cellgreen} .533 \\ 
  & \method{Random}    & \cellcolor{celllred} .232 & \cellcolor{celllgreen} .224 & \cellcolor{cellred} .246 & \cellcolor{celllgreen} .504 & \cellcolor{celllgreen} .560 & \cellcolor{celllgreen} .512 & \cellcolor{celllgreen} .000 & \cellcolor{celllred} .152 & \cellcolor{celllgreen} .304 \\ 
  & \method{Magnitude} & \cellcolor{celllred} .240 & \cellcolor{celllgreen} .234 & \cellcolor{celllred} .258 & \cellcolor{cellred} .452 & \cellcolor{celllred} .524 & \cellcolor{celllgreen} .370 & \cellcolor{celllgreen} .000 & \cellcolor{celllgreen} .162 & \cellcolor{cellred} .280 \\ 
  & \method{Wanda}     & \cellcolor{celllred} .232 & \cellcolor{celllgreen} .228 & \cellcolor{celllred} .256 & \cellcolor{celllgreen} .510 & \cellcolor{cellred} .516 & \cellcolor{celllgreen} .370 & \cellcolor{celllgreen} .000 & \cellcolor{celllgreen} .174 & \cellcolor{celllred} .286 \\ 
  & \method{MeanAct.}  & \cellcolor{celllred} .252 & \cellcolor{celllred} .206 & \cellcolor{celllred} .274 & \cellcolor{celllred} .500 & \cellcolor{celllred} .532 & \cellcolor{celllgreen} .370 & \cellcolor{celllgreen} .000 & \cellcolor{celllred} .138 & \cellcolor{celllred} .284 \\ 
  & \method{LRP}       & \cellcolor{celllgreen} .310 & \cellcolor{cellred} .162 & \cellcolor{celllgreen} .300 & \cellcolor{celllred} .488 & \cellcolor{celllred} .548 & \cellcolor{celllgreen} .370 & \cellcolor{celllgreen} .020 & \cellcolor{celllred} .138 & \cellcolor{celllred} .292 \\ 
  & \method{IG}        & \cellcolor{celllgreen} .314 & \cellcolor{celllred} .208 & \cellcolor{celllgreen} .292 & \cellcolor{celllgreen} .516 & \cellcolor{celllgreen} .558 & \cellcolor{celllgreen} .426 & \cellcolor{celllgreen} .000 & \cellcolor{celllgreen} .162 & \cellcolor{celllgreen} .309 \\ 
  & \consensus{}       & \cellcolor{celllgreen} .336 & \cellcolor{celllred} .184 & \cellcolor{celllgreen} .308 & \cellcolor{celllred} .480 & \cellcolor{celllgreen} .590 & \cellcolor{celllgreen} .370 & \cellcolor{celllgreen} .006 & \cellcolor{celllred} .150 & \cellcolor{celllgreen} .303 \\ 
\midrule
\multirow{8}{*}{LLaMA-8B}
  & Dense          & \cellcolor{cellgreen} .822 & \cellcolor{cellgreen} .532 & \cellcolor{cellgreen} .544 & \cellcolor{cellgreen} .746 & \cellcolor{cellgreen} .798 & \cellcolor{cellgreen} .824 & \cellcolor{cellgreen} .766 & \cellcolor{cellgreen} .338 & \cellcolor{cellgreen} .671 \\ 
  & \method{Random}    & \cellcolor{celllred} .242 & \cellcolor{celllgreen} .226 & \cellcolor{celllred} .260 & \cellcolor{celllred} .504 & \cellcolor{celllred} .508 & \cellcolor{celllgreen} .620 & \cellcolor{celllred} .000 & \cellcolor{celllgreen} .166 & \cellcolor{celllgreen} .316 \\ 
  & \method{Magnitude} & \cellcolor{celllred} .230 & \cellcolor{celllred} .210 & \cellcolor{cellred} .250 & \cellcolor{celllred} .504 & \cellcolor{cellred} .500 & \cellcolor{celllgreen} .604 & \cellcolor{celllred} .000 & \cellcolor{celllgreen} .168 & \cellcolor{celllred} .308 \\ 
  & \method{Wanda}     & \cellcolor{celllred} .238 & \cellcolor{celllred} .206 & \cellcolor{celllred} .256 & \cellcolor{celllred} .514 & \cellcolor{celllred} .536 & \cellcolor{cellred} .370 & \cellcolor{celllred} .000 & \cellcolor{celllred} .164 & \cellcolor{cellred} .285 \\ 
  & \method{MeanAct.}  & \cellcolor{cellred} .222 & \cellcolor{celllred} .202 & \cellcolor{celllred} .252 & \cellcolor{celllgreen} .522 & \cellcolor{celllred} .526 & \cellcolor{celllred} .516 & \cellcolor{celllred} .000 & \cellcolor{celllred} .154 & \cellcolor{celllred} .299 \\ 
  & \method{LRP}       & \cellcolor{celllgreen} .308 & \cellcolor{cellred} .160 & \cellcolor{celllgreen} .292 & \cellcolor{cellred} .500 & \cellcolor{celllgreen} .564 & \cellcolor{celllred} .378 & \cellcolor{celllgreen} .008 & \cellcolor{cellred} .124 & \cellcolor{celllred} .292 \\ 
  & \method{IG}        & \cellcolor{celllgreen} .446 & \cellcolor{celllgreen} .254 & \cellcolor{celllgreen} .342 & \cellcolor{celllgreen} .584 & \cellcolor{celllgreen} .608 & \cellcolor{celllred} .594 & \cellcolor{celllgreen} .056 & \cellcolor{celllgreen} .176 & \cellcolor{celllgreen} .383 \\ 
  & \consensus{}       & \cellcolor{celllgreen} .472 & \cellcolor{celllgreen} .244 & \cellcolor{celllgreen} .348 & \cellcolor{celllgreen} .560 & \cellcolor{celllgreen} .634 & \cellcolor{celllgreen} .632 & \cellcolor{celllgreen} .092 & \cellcolor{celllred} .156 & \cellcolor{celllgreen} .392 \\ 
\midrule
\multirow{8}{*}{Qwen-8B}
  & Dense          & \cellcolor{cellgreen} .842 & \cellcolor{cellgreen} .542 & \cellcolor{cellgreen} .524 & \cellcolor{cellgreen} .672 & \cellcolor{cellgreen} .774 & \cellcolor{cellgreen} .864 & \cellcolor{cellgreen} .658 & \cellcolor{cellgreen} .316 & \cellcolor{cellgreen} .649 \\ 
  & \method{Random}    & \cellcolor{celllred} .244 & \cellcolor{celllgreen} .222 & \cellcolor{cellred} .250 & \cellcolor{celllred} .492 & \cellcolor{celllred} .544 & \cellcolor{celllred} .466 & \cellcolor{celllred} .000 & \cellcolor{celllgreen} .162 & \cellcolor{celllred} .297 \\ 
  & \method{Magnitude} & \cellcolor{celllred} .244 & \cellcolor{celllgreen} .220 & \cellcolor{celllred} .262 & \cellcolor{celllred} .494 & \cellcolor{celllred} .492 & \cellcolor{celllred} .438 & \cellcolor{celllred} .000 & \cellcolor{celllgreen} .158 & \cellcolor{celllred} .288 \\ 
  & \method{Wanda}     & \cellcolor{cellred} .228 & \cellcolor{celllgreen} .240 & \cellcolor{celllred} .268 & \cellcolor{celllgreen} .532 & \cellcolor{celllred} .534 & \cellcolor{celllred} .380 & \cellcolor{celllred} .000 & \cellcolor{celllgreen} .156 & \cellcolor{celllred} .292 \\ 
  & \method{MeanAct.}  & \cellcolor{celllred} .236 & \cellcolor{celllred} .216 & \cellcolor{celllred} .276 & \cellcolor{celllgreen} .498 & \cellcolor{cellred} .490 & \cellcolor{cellred} .370 & \cellcolor{celllred} .000 & \cellcolor{celllred} .144 & \cellcolor{cellred} .279 \\ 
  & \method{LRP}       & \cellcolor{celllgreen} .304 & \cellcolor{cellred} .172 & \cellcolor{celllgreen} .320 & \cellcolor{celllred} .488 & \cellcolor{celllgreen} .558 & \cellcolor{celllgreen} .572 & \cellcolor{celllgreen} .114 & \cellcolor{cellred} .130 & \cellcolor{celllgreen} .332 \\ 
  & \method{IG}        & \cellcolor{celllgreen} .352 & \cellcolor{celllgreen} .220 & \cellcolor{celllgreen} .306 & \cellcolor{celllgreen} .536 & \cellcolor{celllgreen} .622 & \cellcolor{celllgreen} .630 & \cellcolor{celllgreen} .044 & \cellcolor{celllred} .134 & \cellcolor{celllgreen} .355 \\ 
  & \consensus{}       & \cellcolor{celllgreen} .356 & \cellcolor{celllred} .208 & \cellcolor{celllgreen} .308 & \cellcolor{cellred} .486 & \cellcolor{celllgreen} .588 & \cellcolor{celllgreen} .636 & \cellcolor{celllgreen} .068 & \cellcolor{celllred} .134 & \cellcolor{celllgreen} .348 \\ 
\midrule
\multirow{8}{*}{Gemma-12B}
  & Dense          & \cellcolor{cellgreen} .832 & \cellcolor{cellgreen} .618 & \cellcolor{cellgreen} .568 & \cellcolor{cellgreen} .768 & \cellcolor{cellgreen} .796 & \cellcolor{cellgreen} .872 & \cellcolor{cellgreen} .670 & \cellcolor{cellgreen} .412 & \cellcolor{cellgreen} .692 \\ 
  & \method{Random}    & \cellcolor{celllred} .250 & \cellcolor{celllred} .200 & \cellcolor{celllred} .268 & \cellcolor{celllred} .508 & \cellcolor{celllred} .524 & \cellcolor{celllred} .374 & \cellcolor{celllred} .000 & \cellcolor{celllred} .144 & \cellcolor{cellred} .284 \\ 
  & \method{Magnitude} & \cellcolor{cellred} .248 & \cellcolor{celllred} .214 & \cellcolor{celllred} .260 & \cellcolor{celllred} .520 & \cellcolor{celllred} .554 & \cellcolor{celllred} .372 & \cellcolor{celllred} .000 & \cellcolor{celllred} .148 & \cellcolor{celllred} .289 \\ 
  & \method{Wanda}     & \cellcolor{celllred} .260 & \cellcolor{celllgreen} .240 & \cellcolor{cellred} .258 & \cellcolor{cellred} .500 & \cellcolor{cellred} .522 & \cellcolor{cellred} .362 & \cellcolor{celllred} .000 & \cellcolor{celllgreen} .194 & \cellcolor{celllred} .292 \\ 
  & \method{MeanAct.}  & \cellcolor{celllred} .258 & \cellcolor{celllred} .200 & \cellcolor{celllred} .272 & \cellcolor{celllred} .504 & \cellcolor{celllred} .538 & \cellcolor{celllred} .370 & \cellcolor{celllred} .000 & \cellcolor{celllred} .152 & \cellcolor{celllred} .287 \\ 
  & \method{LRP}       & \cellcolor{celllgreen} .344 & \cellcolor{celllred} .214 & \cellcolor{celllgreen} .326 & \cellcolor{celllgreen} .522 & \cellcolor{celllgreen} .560 & \cellcolor{celllgreen} .430 & \cellcolor{celllgreen} .090 & \cellcolor{cellred} .142 & \cellcolor{celllgreen} .329 \\ 
  & \method{IG}        & \cellcolor{celllgreen} .566 & \cellcolor{celllgreen} .318 & \cellcolor{celllgreen} .364 & \cellcolor{celllgreen} .632 & \cellcolor{celllgreen} .646 & \cellcolor{celllgreen} .856 & \cellcolor{celllgreen} .096 & \cellcolor{celllgreen} .220 & \cellcolor{celllgreen} .462 \\ 
  & \consensus{}       & \cellcolor{celllgreen} .586 & \cellcolor{celllgreen} .316 & \cellcolor{celllgreen} .374 & \cellcolor{celllgreen} .636 & \cellcolor{celllgreen} .630 & \cellcolor{celllgreen} .800 & \cellcolor{celllgreen} .442 & \cellcolor{celllgreen} .240 & \cellcolor{celllgreen} .503 \\ 
\bottomrule
\end{tabular}
\end{table*}

\begin{table*}[ht]
\centering
\scriptsize
\setlength{\tabcolsep}{3pt}
\caption{Mean downstream accuracy across 8 lm-eval-harness tasks at
  LeRF masking rates 0-90\% (all 5 models, all 7 selectors). Each
  cell is the average over ARC-E, ARC-C, HellaSwag, WinoGrande, PIQA,
  BoolQ, LAMBADA-OpenAI, and OpenBookQA. Dense baseline (0\%) ranges
  from .533 (LLaMA-1B) to .692 (Gemma-12B). Attribution selectors
  (LRP, IG, \consensus{}) maintain higher accuracy than baselines at
  low-to-moderate rates (10-30\%) before converging to the collapsed
  floor ($\sim$0.28) at high rates (50-90\%). Per-task breakdown for
  LLaMA-8B in Table~\ref{tab:downstream_full_8b}; 30\% snapshot across
  models in Table~\ref{tab:downstream_30pct}.}
\label{tab:downstream_avg_all}
\begin{tabular}{l l *{10}{r}}
\toprule
Model & Selector & 0\% & 10\% & 20\% & 30\% & 40\% & 50\% & 60\% & 70\% & 80\% & 90\% \\
\midrule
\multirow{7}{*}{LLaMA-1B}
  & \method{Random}    & \cellcolor{cellgreen} .533 & \cellcolor{celllred} .286 & \cellcolor{cellred} .280 & \cellcolor{celllgreen} .304 & \cellcolor{celllgreen} .291 & \cellcolor{celllgreen} .287 & \cellcolor{celllgreen} .281 & \cellcolor{cellgreen} .288 & \cellcolor{celllgreen} .278 & \cellcolor{celllgreen} .281 \\ 
  & \method{Magnitude} & \cellcolor{cellgreen} .533 & \cellcolor{celllred} .286 & \cellcolor{celllred} .288 & \cellcolor{cellred} .280 & \cellcolor{celllred} .281 & \cellcolor{celllgreen} .288 & \cellcolor{celllgreen} .282 & \cellcolor{celllgreen} .285 & \cellcolor{celllgreen} .285 & \cellcolor{celllgreen} .281 \\ 
  & \method{Wanda}     & \cellcolor{cellgreen} .533 & \cellcolor{celllgreen} .289 & \cellcolor{celllgreen} .315 & \cellcolor{celllred} .286 & \cellcolor{celllgreen} .283 & \cellcolor{celllred} .281 & \cellcolor{celllgreen} .285 & \cellcolor{celllred} .279 & \cellcolor{cellgreen} .288 & \cellcolor{cellgreen} .289 \\ 
  & \method{MeanAct.}  & \cellcolor{cellgreen} .533 & \cellcolor{cellred} .278 & \cellcolor{celllred} .281 & \cellcolor{celllred} .284 & \cellcolor{celllgreen} .284 & \cellcolor{cellgreen} .303 & \cellcolor{cellgreen} .289 & \cellcolor{celllgreen} .281 & \cellcolor{celllgreen} .287 & \cellcolor{celllgreen} .288 \\ 
  & \method{LRP}       & \cellcolor{cellgreen} .533 & \cellcolor{cellgreen} .443 & \cellcolor{celllgreen} .327 & \cellcolor{celllgreen} .292 & \cellcolor{celllred} .281 & \cellcolor{celllred} .281 & \cellcolor{celllred} .279 & \cellcolor{cellred} .277 & \cellcolor{celllred} .276 & \cellcolor{celllgreen} .282 \\ 
  & \method{IG}        & \cellcolor{cellgreen} .533 & \cellcolor{celllgreen} .440 & \cellcolor{celllgreen} .359 & \cellcolor{cellgreen} .309 & \cellcolor{celllgreen} .283 & \cellcolor{celllred} .283 & \cellcolor{celllred} .279 & \cellcolor{celllgreen} .285 & \cellcolor{cellred} .275 & \cellcolor{cellred} .273 \\ 
  & \consensus{}       & \cellcolor{cellgreen} .533 & \cellcolor{celllgreen} .434 & \cellcolor{cellgreen} .373 & \cellcolor{celllgreen} .303 & \cellcolor{cellgreen} .293 & \cellcolor{celllgreen} .287 & \cellcolor{celllgreen} .281 & \cellcolor{celllred} .279 & \cellcolor{celllgreen} .278 & \cellcolor{celllred} .280 \\ 
\midrule
\multirow{7}{*}{LLaMA-3B}
  & \method{Random}    & \cellcolor{cellgreen} .614 & \cellcolor{celllgreen} .298 & \cellcolor{celllred} .289 & \cellcolor{celllgreen} .299 & \cellcolor{celllgreen} .291 & \cellcolor{celllgreen} .307 & \cellcolor{celllgreen} .283 & \cellcolor{celllgreen} .297 & \cellcolor{celllgreen} .311 & \cellcolor{celllgreen} .288 \\ 
  & \method{Magnitude} & \cellcolor{cellgreen} .614 & \cellcolor{celllred} .283 & \cellcolor{celllgreen} .291 & \cellcolor{celllred} .278 & \cellcolor{cellgreen} .292 & \cellcolor{celllgreen} .288 & \cellcolor{cellgreen} .287 & \cellcolor{celllgreen} .302 & \cellcolor{celllgreen} .291 & \cellcolor{celllred} .280 \\ 
  & \method{Wanda}     & \cellcolor{cellgreen} .614 & \cellcolor{cellred} .277 & \cellcolor{celllred} .285 & \cellcolor{celllgreen} .286 & \cellcolor{celllgreen} .286 & \cellcolor{cellgreen} .312 & \cellcolor{celllgreen} .286 & \cellcolor{cellgreen} .309 & \cellcolor{cellgreen} .319 & \cellcolor{celllgreen} .285 \\ 
  & \method{MeanAct.}  & \cellcolor{cellgreen} .614 & \cellcolor{celllred} .282 & \cellcolor{cellred} .280 & \cellcolor{celllred} .278 & \cellcolor{celllred} .284 & \cellcolor{cellred} .278 & \cellcolor{cellred} .276 & \cellcolor{celllred} .283 & \cellcolor{celllred} .281 & \cellcolor{cellgreen} .289 \\ 
  & \method{LRP}       & \cellcolor{cellgreen} .614 & \cellcolor{celllgreen} .443 & \cellcolor{celllgreen} .328 & \cellcolor{celllred} .285 & \cellcolor{cellred} .281 & \cellcolor{celllgreen} .294 & \cellcolor{celllred} .282 & \cellcolor{cellred} .282 & \cellcolor{cellred} .280 & \cellcolor{celllgreen} .284 \\ 
  & \method{IG}        & \cellcolor{cellgreen} .614 & \cellcolor{cellgreen} .526 & \cellcolor{cellgreen} .406 & \cellcolor{celllgreen} .286 & \cellcolor{celllgreen} .286 & \cellcolor{celllred} .286 & \cellcolor{celllgreen} .285 & \cellcolor{celllred} .283 & \cellcolor{celllred} .286 & \cellcolor{celllgreen} .284 \\ 
  & \consensus{}       & \cellcolor{cellgreen} .614 & \cellcolor{celllgreen} .322 & \cellcolor{celllgreen} .307 & \cellcolor{cellgreen} .341 & \cellcolor{cellgreen} .292 & \cellcolor{celllred} .285 & \cellcolor{celllred} .282 & \cellcolor{celllgreen} .290 & \cellcolor{celllgreen} .293 & \cellcolor{cellred} .277 \\ 
\midrule
\multirow{7}{*}{LLaMA-8B}
  & \method{Random}    & \cellcolor{cellgreen} .671 & \cellcolor{celllgreen} .313 & \cellcolor{celllgreen} .310 & \cellcolor{celllgreen} .316 & \cellcolor{celllred} .291 & \cellcolor{celllgreen} .295 & \cellcolor{cellgreen} .308 & \cellcolor{cellgreen} .321 & \cellcolor{celllred} .288 & \cellcolor{celllgreen} .294 \\ 
  & \method{Magnitude} & \cellcolor{cellgreen} .671 & \cellcolor{celllred} .301 & \cellcolor{celllgreen} .316 & \cellcolor{celllgreen} .308 & \cellcolor{celllgreen} .317 & \cellcolor{celllgreen} .300 & \cellcolor{celllgreen} .283 & \cellcolor{celllred} .280 & \cellcolor{cellgreen} .323 & \cellcolor{celllred} .282 \\ 
  & \method{Wanda}     & \cellcolor{cellgreen} .671 & \cellcolor{celllred} .278 & \cellcolor{cellred} .275 & \cellcolor{cellred} .285 & \cellcolor{celllred} .292 & \cellcolor{celllred} .288 & \cellcolor{cellred} .271 & \cellcolor{celllgreen} .311 & \cellcolor{celllgreen} .309 & \cellcolor{celllgreen} .316 \\ 
  & \method{MeanAct.}  & \cellcolor{cellgreen} .671 & \cellcolor{cellred} .277 & \cellcolor{celllred} .278 & \cellcolor{celllred} .299 & \cellcolor{celllgreen} .318 & \cellcolor{cellgreen} .309 & \cellcolor{celllgreen} .295 & \cellcolor{celllgreen} .281 & \cellcolor{celllgreen} .299 & \cellcolor{cellgreen} .322 \\ 
  & \method{LRP}       & \cellcolor{cellgreen} .671 & \cellcolor{celllgreen} .325 & \cellcolor{celllred} .301 & \cellcolor{celllred} .292 & \cellcolor{cellred} .283 & \cellcolor{cellred} .281 & \cellcolor{celllgreen} .284 & \cellcolor{celllred} .276 & \cellcolor{celllgreen} .312 & \cellcolor{celllgreen} .286 \\ 
  & \method{IG}        & \cellcolor{cellgreen} .671 & \cellcolor{cellgreen} .584 & \cellcolor{cellgreen} .480 & \cellcolor{celllgreen} .383 & \cellcolor{celllgreen} .331 & \cellcolor{celllgreen} .290 & \cellcolor{celllred} .279 & \cellcolor{celllgreen} .286 & \cellcolor{cellred} .277 & \cellcolor{celllred} .279 \\ 
  & \consensus{}       & \cellcolor{cellgreen} .671 & \cellcolor{celllgreen} .580 & \cellcolor{celllgreen} .432 & \cellcolor{cellgreen} .392 & \cellcolor{cellgreen} .337 & \cellcolor{celllred} .284 & \cellcolor{celllred} .282 & \cellcolor{celllred} .276 & \cellcolor{celllred} .283 & \cellcolor{cellred} .276 \\ 
\midrule
\multirow{7}{*}{Qwen-8B}
  & \method{Random}    & \cellcolor{cellgreen} .649 & \cellcolor{celllgreen} .316 & \cellcolor{celllred} .282 & \cellcolor{celllgreen} .297 & \cellcolor{cellgreen} .301 & \cellcolor{celllgreen} .299 & \cellcolor{cellgreen} .307 & \cellcolor{cellgreen} .316 & \cellcolor{celllgreen} .298 & \cellcolor{celllgreen} .302 \\ 
  & \method{Magnitude} & \cellcolor{cellgreen} .649 & \cellcolor{celllred} .291 & \cellcolor{celllgreen} .305 & \cellcolor{celllred} .288 & \cellcolor{celllgreen} .292 & \cellcolor{celllgreen} .293 & \cellcolor{celllgreen} .293 & \cellcolor{celllgreen} .297 & \cellcolor{cellgreen} .300 & \cellcolor{cellgreen} .307 \\ 
  & \method{Wanda}     & \cellcolor{cellgreen} .649 & \cellcolor{celllred} .282 & \cellcolor{celllred} .285 & \cellcolor{celllred} .292 & \cellcolor{cellred} .282 & \cellcolor{celllred} .284 & \cellcolor{celllgreen} .285 & \cellcolor{celllgreen} .289 & \cellcolor{celllgreen} .297 & \cellcolor{celllred} .289 \\ 
  & \method{MeanAct.}  & \cellcolor{cellgreen} .649 & \cellcolor{cellred} .278 & \cellcolor{celllred} .282 & \cellcolor{cellred} .279 & \cellcolor{celllgreen} .286 & \cellcolor{cellgreen} .313 & \cellcolor{cellred} .274 & \cellcolor{cellred} .268 & \cellcolor{cellred} .275 & \cellcolor{celllred} .285 \\ 
  & \method{LRP}       & \cellcolor{cellgreen} .649 & \cellcolor{celllgreen} .553 & \cellcolor{celllgreen} .409 & \cellcolor{celllgreen} .332 & \cellcolor{celllred} .285 & \cellcolor{cellred} .281 & \cellcolor{celllred} .284 & \cellcolor{celllred} .282 & \cellcolor{celllred} .281 & \cellcolor{cellred} .269 \\ 
  & \method{IG}        & \cellcolor{cellgreen} .649 & \cellcolor{cellgreen} .589 & \cellcolor{cellgreen} .443 & \cellcolor{cellgreen} .355 & \cellcolor{celllgreen} .295 & \cellcolor{celllred} .284 & \cellcolor{celllgreen} .292 & \cellcolor{celllgreen} .303 & \cellcolor{celllred} .286 & \cellcolor{celllgreen} .302 \\ 
  & \consensus{}       & \cellcolor{cellgreen} .649 & \cellcolor{celllgreen} .583 & \cellcolor{celllgreen} .439 & \cellcolor{celllgreen} .348 & \cellcolor{celllred} .285 & \cellcolor{celllgreen} .294 & \cellcolor{celllred} .283 & \cellcolor{celllred} .283 & \cellcolor{celllgreen} .287 & \cellcolor{celllgreen} .292 \\ 
\midrule
\multirow{7}{*}{Gemma-12B}
  & \method{Random}    & \cellcolor{cellgreen} .692 & \cellcolor{celllred} .375 & \cellcolor{cellred} .286 & \cellcolor{cellred} .284 & \cellcolor{celllgreen} .291 & \cellcolor{celllred} .290 & \cellcolor{celllgreen} .296 & \cellcolor{celllgreen} .309 & \cellcolor{cellgreen} .310 & \cellcolor{celllgreen} .285 \\ 
  & \method{Magnitude} & \cellcolor{cellgreen} .692 & \cellcolor{cellred} .287 & \cellcolor{celllgreen} .317 & \cellcolor{celllred} .289 & \cellcolor{celllred} .288 & \cellcolor{cellgreen} .329 & \cellcolor{cellgreen} .311 & \cellcolor{cellgreen} .313 & \cellcolor{celllgreen} .290 & \cellcolor{celllgreen} .313 \\ 
  & \method{Wanda}     & \cellcolor{cellgreen} .692 & \cellcolor{cellgreen} .692 & \cellcolor{celllred} .292 & \cellcolor{celllgreen} .292 & \cellcolor{celllred} .290 & \cellcolor{celllred} .287 & \cellcolor{celllgreen} .289 & \cellcolor{celllgreen} .298 & \cellcolor{celllred} .286 & \cellcolor{cellgreen} .318 \\ 
  & \method{MeanAct.}  & \cellcolor{cellgreen} .692 & \cellcolor{cellgreen} .692 & \cellcolor{celllred} .292 & \cellcolor{celllred} .287 & \cellcolor{cellred} .283 & \cellcolor{cellred} .280 & \cellcolor{celllred} .287 & \cellcolor{celllgreen} .288 & \cellcolor{celllgreen} .291 & \cellcolor{celllred} .279 \\ 
  & \method{LRP}       & \cellcolor{cellgreen} .692 & \cellcolor{celllred} .585 & \cellcolor{celllgreen} .428 & \cellcolor{celllgreen} .329 & \cellcolor{celllgreen} .311 & \cellcolor{celllgreen} .299 & \cellcolor{celllgreen} .302 & \cellcolor{celllred} .286 & \cellcolor{cellred} .277 & \cellcolor{celllgreen} .282 \\ 
  & \method{IG}        & \cellcolor{cellgreen} .692 & \cellcolor{celllgreen} .652 & \cellcolor{celllgreen} .585 & \cellcolor{celllgreen} .462 & \cellcolor{celllgreen} .362 & \cellcolor{celllgreen} .322 & \cellcolor{cellred} .277 & \cellcolor{celllred} .283 & \cellcolor{celllgreen} .290 & \cellcolor{cellred} .276 \\ 
  & \consensus{}       & \cellcolor{cellgreen} .692 & \cellcolor{celllgreen} .645 & \cellcolor{cellgreen} .588 & \cellcolor{cellgreen} .503 & \cellcolor{cellgreen} .384 & \cellcolor{celllgreen} .327 & \cellcolor{celllred} .286 & \cellcolor{cellred} .279 & \cellcolor{celllred} .282 & \cellcolor{celllred} .279 \\ 
\bottomrule
\end{tabular}
\end{table*}

% ============================================================
\clearpage
\section{Stability, Convergence, and Inter-Method Agreement}
\label{app:stability_full}

\begin{figure}[ht]
  \centering
  \includegraphics[width=0.65\linewidth]{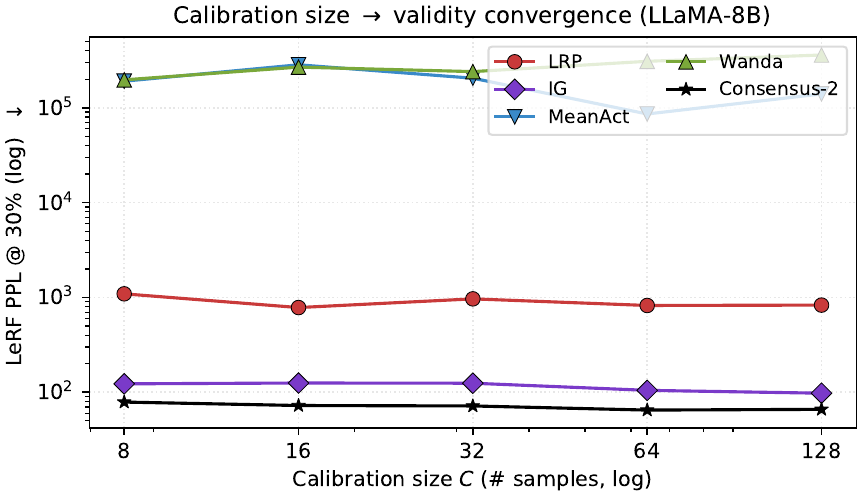}
  \caption{PPL@30\% vs.\ calibration size (C8-C128) for all five
    data-dependent selectors on LLaMA-8B (log $y$). MeanActivation
    is flat (stable) but high (invalid); \consensus{} converges
    smoothly to the lowest PPL.}
  \label{fig:convergence_8b}
\end{figure}

Tables~\ref{tab:convergence_full}--\ref{tab:convergence_full_qwen} report full calibration-size convergence (PPL at all sizes C8-C128 and masking rates); Tables~\ref{tab:jaccard_stability}--\ref{tab:jaccard_stability_qwen} give mask-level Jaccard similarity vs.\ the C128 reference; Tables~\ref{tab:spearman}, \ref{tab:spearman_qwen}, and~\ref{tab:stability_qwen} give Spearman rank correlations and the Qwen-8B stability profile.
Figure~\ref{fig:convergence_8b} plots PPL@30\% vs.\ calibration size for all data-dependent selectors on LLaMA-8B.

\subsection{Calibration-size convergence (LLaMA-8B and Qwen-8B).}

Attribution methods converge smoothly C8$\to$C128. \method{LRP}
PPL@30\% drops from $1{,}092$ to $830$ on LLaMA-8B and from
$1{,}312$ to $724$ on Qwen-8B; \method{IG} drops from $123$ to $98$
on LLaMA-8B; \consensus{} reaches the lowest PPL on both
architectures ($66$ on LLaMA-8B, $187$ on Qwen-8B at C128). All three converge overall with calibration size, indicating they
extract a real data-dependent signal that more samples sharpen.
\method{Wanda} exhibits the opposite: LeRF PPL \emph{increases}
with more calibration data (C8: $197\text{k} \to$ C128: $360\text{k}$
on LLaMA-8B; C8: $5.4\times10^6 \to$ C128: $1.95\times10^7$ on
Qwen-8B), indicating that additional activation statistics push its
scores further from causal relevance.
\method{MeanAct.} shows an architecture-dependent pattern: it stays
high on LLaMA-8B at all calibration sizes (PPL $\approx 140\text{k}$
at C128), but improves substantially on Qwen-8B (C8: $360\text{k}
\to$ C128: $65.9\text{k}$), yet remains orders of magnitude worse
than attribution methods at every calibration size on both
architectures.

The Jaccard mask-stability tables
(Tables~\ref{tab:jaccard_stability}, \ref{tab:jaccard_stability_qwen})
make the stability-validity dissociation explicit. \method{Wanda}'s
mask is essentially fixed across calibration sizes (Jaccard
$0.97$-$1.00$ vs.\ C128); \method{MeanAct.} is also extremely stable
(Jaccard $0.92$-$0.98$). \method{LRP}, \method{IG}, and
\consensus{} have lower Jaccard at small $C$ ($0.51$-$0.90$
depending on rate and model) and grow with calibration size, so
their masks shift as more samples arrive. The selectors that move
least are the ones least anchored in the data, which is what makes
them stable but causally invalid.

The Spearman score-level tables
(Tables~\ref{tab:method_spearman_8b}, \ref{tab:method_spearman_qwen},
and~\ref{tab:method_spearman} for LLaMA-8B, Qwen-8B, and LLaMA-1B respectively) extend this picture
across selectors. The non-attribution cluster
(\method{Magnitude}, \method{Wanda}, \method{MeanAct.}) is
internally coherent ($\rho = 0.71$-$0.88$ on LLaMA-8B), and the
attribution cluster (\method{LRP}, \method{IG}, \consensus{}) is
also internally coherent ($\rho = 0.65$-$0.91$). Notably, the
two clusters \emph{anti-correlate}: cross-cluster $\rho$ ranges
from $-0.30$ to $-0.58$ on LLaMA-8B (and $-0.43$ to $0.04$ on
Qwen-8B). Selectors that rank by data-independent weight or
activation statistics rank rows in roughly the opposite order from
selectors that propagate causal signal, so the two paradigms are
not just imperfectly correlated: they capture systematically
different structure.

\begin{table*}[ht]
\centering
\small
\setlength{\tabcolsep}{3pt}
\caption{PPL at all calibration sizes $\times$ masking rates $\times$ methods (LLaMA-8B).
  5 data-dependent methods, sizes C8-C128, rates 10/20/30/50\%.}
\label{tab:convergence_full}
\begin{tabular}{l l rrrr}
\toprule
Selector & Size & PPL@10\% & PPL@20\% & PPL@30\% & PPL@50\% \\
\midrule
\multirow{5}{*}{\method{LRP}}
  & C8   & \cellcolor{celllgreen} 143.17 & \cellcolor{celllred} 406.42 & \cellcolor{cellred} 1091.54 & \cellcolor{cellred} 3029.93 \\ 
  & C16  & \cellcolor{cellred} 225.72 & \cellcolor{cellred} 565.50 & \cellcolor{cellgreen} 782.35 & \cellcolor{celllred} 2990.52 \\ 
  & C32  & \cellcolor{cellgreen} 124.40 & \cellcolor{celllgreen} 350.56 & \cellcolor{celllred} 966.27 & \cellcolor{celllgreen} 2426.62 \\ 
  & C64  & \cellcolor{celllgreen} 131.79 & \cellcolor{celllgreen} 346.03 & \cellcolor{celllgreen} 822.99 & \cellcolor{celllgreen} 2165.87 \\ 
  & C128 & \cellcolor{celllred} 218.72 & \cellcolor{cellgreen} 340.46 & \cellcolor{celllgreen} 830.38 & \cellcolor{cellgreen} 2108.11 \\ 
\midrule
\multirow{5}{*}{\method{IG}}
  & C8   & \cellcolor{cellred} 19.73 & \cellcolor{celllred} 45.21 & \cellcolor{celllgreen} 122.77 & \cellcolor{celllgreen} 1077.53 \\ 
  & C16  & \cellcolor{celllred} 19.44 & \cellcolor{cellred} 45.58 & \cellcolor{cellred} 125.26 & \cellcolor{cellgreen} 1064.62 \\ 
  & C32  & \cellcolor{celllgreen} 19.27 & \cellcolor{celllgreen} 43.54 & \cellcolor{celllred} 124.65 & \cellcolor{celllred} 1330.44 \\ 
  & C64  & \cellcolor{celllgreen} 18.30 & \cellcolor{celllgreen} 40.00 & \cellcolor{celllgreen} 104.82 & \cellcolor{cellred} 1375.04 \\ 
  & C128 & \cellcolor{cellgreen} 18.25 & \cellcolor{cellgreen} 38.96 & \cellcolor{cellgreen} 97.79 & \cellcolor{celllgreen} 1274.79 \\ 
\midrule
\multirow{5}{*}{\method{MeanAct.}}
  & C8   & \cellcolor{celllgreen} 8524.86 & \cellcolor{cellred} 138676.38 & \cellcolor{celllgreen} 190639.61 & \cellcolor{cellgreen} 891119.46 \\ 
  & C16  & \cellcolor{celllred} 9319.21 & \cellcolor{celllred} 122148.29 & \cellcolor{cellred} 283958.89 & \cellcolor{celllgreen} 3.23e6 \\ 
  & C32  & \cellcolor{cellgreen} 7489.42 & \cellcolor{celllgreen} 71002.56 & \cellcolor{celllred} 204995.58 & \cellcolor{celllgreen} 2.94e6 \\ 
  & C64  & \cellcolor{cellred} 9969.30 & \cellcolor{celllgreen} 70056.76 & \cellcolor{cellgreen} 85998.22 & \cellcolor{celllred} 5.25e6 \\ 
  & C128 & \cellcolor{celllgreen} 8688.86 & \cellcolor{cellgreen} 42793.06 & \cellcolor{celllgreen} 139874.10 & \cellcolor{cellred} 6.28e6 \\ 
\midrule
\multirow{5}{*}{\method{Wanda}}
  & C8   & \cellcolor{celllgreen} 22592.14 & \cellcolor{cellred} 194265.23 & \cellcolor{cellgreen} 196927.15 & \cellcolor{cellred} 263282.16 \\ 
  & C16  & \cellcolor{celllgreen} 21320.19 & \cellcolor{celllred} 176425.26 & \cellcolor{celllgreen} 268372.34 & \cellcolor{celllgreen} 102723.13 \\ 
  & C32  & \cellcolor{celllred} 23135.39 & \cellcolor{cellgreen} 144191.40 & \cellcolor{celllgreen} 240606.88 & \cellcolor{celllred} 150619.53 \\ 
  & C64  & \cellcolor{cellred} 23455.56 & \cellcolor{celllgreen} 171089.39 & \cellcolor{celllred} 308971.59 & \cellcolor{celllgreen} 91331.71 \\ 
  & C128 & \cellcolor{cellgreen} 20027.75 & \cellcolor{celllgreen} 148982.31 & \cellcolor{cellred} 360082.74 & \cellcolor{cellgreen} 89066.16 \\ 
\midrule
\multirow{5}{*}{\consensus{}}
  & C8   & \cellcolor{cellred} 17.21 & \cellcolor{cellred} 33.12 & \cellcolor{cellred} 78.85 & \cellcolor{celllred} 895.01 \\ 
  & C16  & \cellcolor{celllred} 16.97 & \cellcolor{celllred} 32.90 & \cellcolor{celllred} 72.52 & \cellcolor{cellred} 929.28 \\ 
  & C32  & \cellcolor{celllgreen} 16.40 & \cellcolor{celllgreen} 32.30 & \cellcolor{celllgreen} 71.76 & \cellcolor{celllgreen} 754.09 \\ 
  & C64  & \cellcolor{cellgreen} 16.24 & \cellcolor{celllgreen} 30.45 & \cellcolor{cellgreen} 64.92 & \cellcolor{celllgreen} 793.32 \\ 
  & C128 & \cellcolor{cellgreen} 16.24 & \cellcolor{cellgreen} 29.95 & \cellcolor{celllgreen} 66.08 & \cellcolor{cellgreen} 737.56 \\ 
\bottomrule
\end{tabular}
\end{table*}

\begin{table*}[ht]
\centering
\small
\setlength{\tabcolsep}{3pt}
\caption{PPL at all calibration sizes $\times$ masking rates $\times$ methods (Qwen-8B).
  5 data-dependent methods, sizes C8-C128, rates 10/20/30/50\%.}
\label{tab:convergence_full_qwen}
\begin{tabular}{l l rrrr}
\toprule
Selector & Size & PPL@10\% & PPL@20\% & PPL@30\% & PPL@50\% \\
\midrule
\multirow{5}{*}{\method{LRP}}
  & C8   & \cellcolor{cellgreen} 32.66 & \cellcolor{cellred} 312.55 & \cellcolor{cellred} 1311.68 & \cellcolor{cellred} 6275.03 \\ 
  & C16  & \cellcolor{celllgreen} 35.97 & \cellcolor{celllgreen} 219.59 & \cellcolor{celllgreen} 1092.08 & \cellcolor{celllred} 4960.51 \\ 
  & C32  & \cellcolor{cellred} 40.34 & \cellcolor{celllred} 254.33 & \cellcolor{celllred} 1093.23 & \cellcolor{celllgreen} 4181.11 \\ 
  & C64  & \cellcolor{celllred} 40.29 & \cellcolor{celllgreen} 188.70 & \cellcolor{cellgreen} 627.52 & \cellcolor{celllgreen} 4651.75 \\ 
  & C128 & \cellcolor{celllgreen} 39.93 & \cellcolor{cellgreen} 181.12 & \cellcolor{celllgreen} 724.26 & \cellcolor{cellgreen} 3355.82 \\ 
\midrule
\multirow{5}{*}{\method{IG}}
  & C8   & \cellcolor{celllred} 35.31 & \cellcolor{cellred} 82.74 & \cellcolor{celllred} 253.39 & \cellcolor{celllgreen} 2531.52 \\ 
  & C16  & \cellcolor{celllgreen} 34.90 & \cellcolor{celllred} 79.38 & \cellcolor{celllgreen} 252.86 & \cellcolor{cellgreen} 2416.50 \\ 
  & C32  & \cellcolor{cellred} 35.78 & \cellcolor{celllgreen} 77.23 & \cellcolor{cellred} 263.41 & \cellcolor{cellred} 3169.32 \\ 
  & C64  & \cellcolor{celllgreen} 34.34 & \cellcolor{cellgreen} 74.01 & \cellcolor{celllgreen} 241.83 & \cellcolor{celllgreen} 2757.75 \\ 
  & C128 & \cellcolor{cellgreen} 33.03 & \cellcolor{celllgreen} 76.50 & \cellcolor{cellgreen} 228.79 & \cellcolor{celllred} 3074.25 \\ 
\midrule
\multirow{5}{*}{\method{MeanAct.}}
  & C8   & \cellcolor{celllgreen} 15277.74 & \cellcolor{celllred} 60250.96 & \cellcolor{cellred} 359687.43 & \cellcolor{cellred} 6.50e6 \\ 
  & C16  & \cellcolor{celllgreen} 17216.07 & \cellcolor{cellred} 878133.38 & \cellcolor{celllred} 109930.64 & \cellcolor{celllgreen} 1.81e6 \\ 
  & C32  & \cellcolor{cellred} 19595.10 & \cellcolor{cellgreen} 43709.65 & \cellcolor{celllgreen} 89529.09 & \cellcolor{celllred} 2.81e6 \\ 
  & C64  & \cellcolor{cellgreen} 14993.40 & \cellcolor{celllgreen} 55328.04 & \cellcolor{cellgreen} 64808.50 & \cellcolor{cellgreen} 1.69e6 \\ 
  & C128 & \cellcolor{celllred} 18600.53 & \cellcolor{celllgreen} 58346.41 & \cellcolor{celllgreen} 65913.15 & \cellcolor{celllgreen} 1.84e6 \\ 
\midrule
\multirow{5}{*}{\method{Wanda}}
  & C8   & \cellcolor{celllred} 51142.39 & \cellcolor{celllgreen} 238091.20 & \cellcolor{cellgreen} 5.40e6 & \cellcolor{celllgreen} 3.19e6 \\ 
  & C16  & \cellcolor{celllgreen} 49620.06 & \cellcolor{celllgreen} 193918.23 & \cellcolor{celllgreen} 9.61e6 & \cellcolor{cellred} 4.45e6 \\ 
  & C32  & \cellcolor{celllgreen} 48357.82 & \cellcolor{cellred} 344607.18 & \cellcolor{celllred} 1.50e7 & \cellcolor{celllgreen} 3.37e6 \\ 
  & C64  & \cellcolor{cellgreen} 40048.66 & \cellcolor{celllred} 247664.77 & \cellcolor{celllgreen} 1.24e7 & \cellcolor{cellgreen} 3.17e6 \\ 
  & C128 & \cellcolor{cellred} 67494.64 & \cellcolor{cellgreen} 69640.78 & \cellcolor{cellred} 1.95e7 & \cellcolor{celllred} 4.17e6 \\ 
\midrule
\multirow{5}{*}{\consensus{}}
  & C8   & \cellcolor{celllred} 30.41 & \cellcolor{cellred} 79.53 & \cellcolor{cellred} 231.67 & \cellcolor{cellred} 1920.62 \\ 
  & C16  & \cellcolor{celllgreen} 29.48 & \cellcolor{celllgreen} 73.20 & \cellcolor{celllred} 221.16 & \cellcolor{celllred} 1539.67 \\ 
  & C32  & \cellcolor{cellred} 30.42 & \cellcolor{celllred} 75.31 & \cellcolor{celllgreen} 209.32 & \cellcolor{cellgreen} 1343.81 \\ 
  & C64  & \cellcolor{celllgreen} 29.39 & \cellcolor{celllgreen} 69.48 & \cellcolor{celllgreen} 202.85 & \cellcolor{celllgreen} 1410.55 \\ 
  & C128 & \cellcolor{cellgreen} 28.82 & \cellcolor{cellgreen} 69.10 & \cellcolor{cellgreen} 187.24 & \cellcolor{celllgreen} 1378.05 \\ 
\bottomrule
\end{tabular}
\end{table*}

\begin{table}[ht]
\centering
\small
\setlength{\tabcolsep}{3pt}
\caption{Jaccard mask stability vs.\ C128 reference at masking rates
  10/20/30/50\% (LLaMA-8B).}
\label{tab:jaccard_stability}
\begin{tabular}{l l rrrr}
\toprule
Selector & Size & J@10\% & J@20\% & J@30\% & J@50\% \\
\midrule
\multirow{4}{*}{\method{LRP}}
  & C8  & \cellcolor{cellred} 0.5111 & \cellcolor{cellred} 0.5545 & \cellcolor{cellred} 0.6349 & \cellcolor{cellred} 0.7839 \\ 
  & C16 & \cellcolor{celllred} 0.6343 & \cellcolor{celllred} 0.6442 & \cellcolor{celllred} 0.7114 & \cellcolor{celllred} 0.8442 \\ 
  & C32 & \cellcolor{celllgreen} 0.7138 & \cellcolor{celllgreen} 0.7072 & \cellcolor{celllgreen} 0.7675 & \cellcolor{celllgreen} 0.8850 \\ 
  & C64 & \cellcolor{cellgreen} 0.8074 & \cellcolor{cellgreen} 0.8067 & \cellcolor{cellgreen} 0.8555 & \cellcolor{cellgreen} 0.9333 \\ 
\midrule
\multirow{4}{*}{\method{IG}}
  & C8  & \cellcolor{cellred} 0.5735 & \cellcolor{cellred} 0.6540 & \cellcolor{cellred} 0.7558 & \cellcolor{cellred} 0.9026 \\ 
  & C16 & \cellcolor{celllred} 0.6868 & \cellcolor{celllred} 0.7399 & \cellcolor{celllred} 0.8200 & \cellcolor{celllred} 0.9303 \\ 
  & C32 & \cellcolor{celllgreen} 0.7458 & \cellcolor{celllgreen} 0.7927 & \cellcolor{celllgreen} 0.8635 & \cellcolor{celllgreen} 0.9531 \\ 
  & C64 & \cellcolor{cellgreen} 0.8277 & \cellcolor{cellgreen} 0.8641 & \cellcolor{cellgreen} 0.9148 & \cellcolor{cellgreen} 0.9703 \\ 
\midrule
\multirow{4}{*}{\method{MeanAct.}}
  & C8  & \cellcolor{cellred} 0.9421 & \cellcolor{cellred} 0.9584 & \cellcolor{cellred} 0.9337 & \cellcolor{cellred} 0.9220 \\ 
  & C16 & \cellcolor{celllred} 0.9617 & \cellcolor{celllred} 0.9736 & \cellcolor{celllred} 0.9561 & \cellcolor{celllred} 0.9453 \\ 
  & C32 & \cellcolor{celllgreen} 0.9752 & \cellcolor{celllgreen} 0.9841 & \cellcolor{celllgreen} 0.9734 & \cellcolor{celllgreen} 0.9638 \\ 
  & C64 & \cellcolor{cellgreen} 0.9840 & \cellcolor{cellgreen} 0.9895 & \cellcolor{cellgreen} 0.9839 & \cellcolor{cellgreen} 0.9790 \\ 
\midrule
\multirow{4}{*}{\method{Wanda}}
  & C8  & \cellcolor{celllred} 0.9969 & \cellcolor{cellred} 0.9902 & \cellcolor{cellred} 0.9807 & \cellcolor{cellred} 0.9905 \\ 
  & C16 & \cellcolor{celllgreen} 0.9977 & \cellcolor{celllred} 0.9903 & \cellcolor{celllred} 0.9840 & \cellcolor{celllgreen} 0.9964 \\ 
  & C32 & \cellcolor{cellred} 0.9873 & \cellcolor{celllgreen} 0.9976 & \cellcolor{celllgreen} 0.9900 & \cellcolor{celllred} 0.9953 \\ 
  & C64 & \cellcolor{cellgreen} 0.9982 & \cellcolor{cellgreen} 0.9988 & \cellcolor{cellgreen} 0.9990 & \cellcolor{cellgreen} 0.9993 \\ 
\midrule
\multirow{4}{*}{\consensus{}}
  & C8  & \cellcolor{cellred} 0.5299 & \cellcolor{cellred} 0.6129 & \cellcolor{cellred} 0.7268 & \cellcolor{cellred} 0.8305 \\ 
  & C16 & \cellcolor{celllred} 0.6307 & \cellcolor{celllred} 0.6940 & \cellcolor{celllred} 0.7948 & \cellcolor{celllred} 0.8782 \\ 
  & C32 & \cellcolor{celllgreen} 0.6909 & \cellcolor{celllgreen} 0.7488 & \cellcolor{celllgreen} 0.8401 & \cellcolor{celllgreen} 0.9140 \\ 
  & C64 & \cellcolor{cellgreen} 0.7966 & \cellcolor{cellgreen} 0.8389 & \cellcolor{cellgreen} 0.9041 & \cellcolor{cellgreen} 0.9479 \\ 
\bottomrule
\end{tabular}
\end{table}

\begin{table}[ht]
\centering
\small
\setlength{\tabcolsep}{3pt}
\caption{Jaccard mask stability vs.\ C128 reference at masking rates
  10/20/30/50\% (Qwen-8B).}
\label{tab:jaccard_stability_qwen}
\begin{tabular}{l l rrrr}
\toprule
Selector & Size & J@10\% & J@20\% & J@30\% & J@50\% \\
\midrule
\multirow{4}{*}{\method{LRP}}
  & C8  & \cellcolor{cellred} 0.7774 & \cellcolor{cellred} 0.7100 & \cellcolor{cellred} 0.7602 & \cellcolor{cellred} 0.8476 \\ 
  & C16 & \cellcolor{celllred} 0.8204 & \cellcolor{celllred} 0.7568 & \cellcolor{celllred} 0.8142 & \cellcolor{celllred} 0.8908 \\ 
  & C32 & \cellcolor{celllgreen} 0.8835 & \cellcolor{celllgreen} 0.8154 & \cellcolor{celllgreen} 0.8540 & \cellcolor{celllgreen} 0.9216 \\ 
  & C64 & \cellcolor{cellgreen} 0.9246 & \cellcolor{cellgreen} 0.8699 & \cellcolor{cellgreen} 0.9009 & \cellcolor{cellgreen} 0.9502 \\ 
\midrule
\multirow{4}{*}{\method{IG}}
  & C8  & \cellcolor{cellred} 0.7637 & \cellcolor{cellred} 0.8029 & \cellcolor{cellred} 0.8490 & \cellcolor{cellred} 0.8992 \\ 
  & C16 & \cellcolor{celllred} 0.8177 & \cellcolor{celllred} 0.8527 & \cellcolor{celllred} 0.8888 & \cellcolor{celllred} 0.9245 \\ 
  & C32 & \cellcolor{celllgreen} 0.8676 & \cellcolor{celllgreen} 0.8955 & \cellcolor{celllgreen} 0.9250 & \cellcolor{celllgreen} 0.9511 \\ 
  & C64 & \cellcolor{cellgreen} 0.9143 & \cellcolor{cellgreen} 0.9327 & \cellcolor{cellgreen} 0.9524 & \cellcolor{cellgreen} 0.9693 \\ 
\midrule
\multirow{4}{*}{\method{MeanAct.}}
  & C8  & \cellcolor{cellred} 0.9417 & \cellcolor{cellred} 0.9132 & \cellcolor{cellred} 0.8765 & \cellcolor{cellred} 0.9555 \\ 
  & C16 & \cellcolor{celllred} 0.9628 & \cellcolor{celllred} 0.9441 & \cellcolor{celllred} 0.9145 & \cellcolor{celllred} 0.9707 \\ 
  & C32 & \cellcolor{celllgreen} 0.9750 & \cellcolor{celllgreen} 0.9653 & \cellcolor{celllgreen} 0.9461 & \cellcolor{celllgreen} 0.9817 \\ 
  & C64 & \cellcolor{cellgreen} 0.9843 & \cellcolor{cellgreen} 0.9770 & \cellcolor{cellgreen} 0.9660 & \cellcolor{cellgreen} 0.9886 \\ 
\midrule
\multirow{4}{*}{\method{Wanda}}
  & C8  & \cellcolor{cellred} 0.9812 & \cellcolor{celllred} 0.9890 & \cellcolor{cellred} 0.9752 & \cellcolor{cellred} 0.9902 \\ 
  & C16 & \cellcolor{celllred} 0.9847 & \cellcolor{cellred} 0.9880 & \cellcolor{celllred} 0.9838 & \cellcolor{celllred} 0.9928 \\ 
  & C32 & \cellcolor{celllgreen} 0.9849 & \cellcolor{celllgreen} 0.9968 & \cellcolor{cellgreen} 0.9977 & \cellcolor{cellgreen} 0.9979 \\ 
  & C64 & \cellcolor{cellgreen} 0.9935 & \cellcolor{cellgreen} 0.9983 & \cellcolor{celllgreen} 0.9975 & \cellcolor{celllgreen} 0.9972 \\ 
\midrule
\multirow{4}{*}{\consensus{}}
  & C8  & \cellcolor{cellred} 0.7548 & \cellcolor{cellred} 0.7564 & \cellcolor{cellred} 0.8178 & \cellcolor{cellred} 0.8828 \\ 
  & C16 & \cellcolor{celllred} 0.7992 & \cellcolor{celllred} 0.8098 & \cellcolor{celllred} 0.8645 & \cellcolor{celllred} 0.9144 \\ 
  & C32 & \cellcolor{celllgreen} 0.8508 & \cellcolor{celllgreen} 0.8532 & \cellcolor{celllgreen} 0.8986 & \cellcolor{celllgreen} 0.9424 \\ 
  & C64 & \cellcolor{cellgreen} 0.8985 & \cellcolor{cellgreen} 0.9003 & \cellcolor{cellgreen} 0.9331 & \cellcolor{cellgreen} 0.9636 \\ 
\bottomrule
\end{tabular}
\end{table}

\begin{table}[ht]
\centering
\small
\setlength{\tabcolsep}{4pt}
\caption{Spearman rank correlation vs.\ C128 scores (LLaMA-8B).}
\label{tab:spearman}
\begin{tabular}{l rrrr}
\toprule
Selector & C8 & C16 & C32 & C64 \\
\midrule
\method{MeanAct.}  & \cellcolor{celllgreen} 0.9937 & \cellcolor{celllgreen} 0.9970 & \cellcolor{celllgreen} 0.9988 & \cellcolor{celllgreen} 0.9996 \\ 
\method{Wanda}     & \cellcolor{cellgreen} 0.9998 & \cellcolor{cellgreen} 0.9999 & \cellcolor{cellgreen} 0.9999 & \cellcolor{cellgreen} 1.0000 \\ 
\method{IG}        & \cellcolor{celllgreen} 0.9647 & \cellcolor{celllgreen} 0.9823 & \cellcolor{celllgreen} 0.9905 & \cellcolor{celllgreen} 0.9961 \\ 
\method{LRP}       & \cellcolor{cellred} 0.9130 & \cellcolor{cellred} 0.9529 & \cellcolor{cellred} 0.9720 & \cellcolor{cellred} 0.9897 \\ 
\consensus{}       & \cellcolor{celllred} 0.9448 & \cellcolor{celllred} 0.9707 & \cellcolor{celllred} 0.9833 & \cellcolor{celllred} 0.9937 \\ 
\bottomrule
\end{tabular}
\end{table}

\begin{table}[ht]
\centering
\small
\setlength{\tabcolsep}{4pt}
\caption{Spearman rank correlation vs.\ C128 scores (Qwen-8B).}
\label{tab:spearman_qwen}
\begin{tabular}{l rrrr}
\toprule
Selector & C8 & C16 & C32 & C64 \\
\midrule
\method{MeanAct.}  & \cellcolor{celllgreen} 0.9952 & \cellcolor{celllgreen} 0.9979 & \cellcolor{celllgreen} 0.9992 & \cellcolor{celllgreen} 0.9997 \\ 
\method{Wanda}     & \cellcolor{cellgreen} 0.9997 & \cellcolor{cellgreen} 0.9999 & \cellcolor{cellgreen} 0.9999 & \cellcolor{cellgreen} 1.0000 \\ 
\method{IG}        & \cellcolor{celllgreen} 0.9823 & \cellcolor{celllgreen} 0.9904 & \cellcolor{celllgreen} 0.9956 & \cellcolor{celllgreen} 0.9982 \\ 
\method{LRP}       & \cellcolor{cellred} 0.9593 & \cellcolor{cellred} 0.9763 & \cellcolor{cellred} 0.9865 & \cellcolor{cellred} 0.9936 \\ 
\consensus{}       & \cellcolor{celllred} 0.9758 & \cellcolor{celllred} 0.9863 & \cellcolor{celllred} 0.9926 & \cellcolor{celllred} 0.9966 \\ 
\bottomrule
\end{tabular}
\end{table}

\begin{table}[ht]
\centering
\caption{Stability vs.\ validity (Qwen-8B). Same format as
  Table~\ref{tab:stability}. Confirms the dissociation
  replicates across architectures.
  \method{Wanda}'s negative gap ($-9.78\times10^6$) is the most extreme inversion among activation-weighted selectors: its ``important'' set is less important than its ``dispensable'' set by nearly $10\text{M}$ PPL.
  \method{LRP} achieves the best MoRF ($3.74\times10^7$), outperforming \method{IG} ($2.68\times10^7$): the same LRP/IG ordering that appears at the behavior level on Qwen3-8B.}
\label{tab:stability_qwen}
\begin{tabular}{l rr rrr}
\toprule
& \multicolumn{2}{c}{Stability (C8$\to$C128)} & \multicolumn{3}{c}{Validity (C128, @30\%)} \\
\cmidrule(lr){2-3}\cmidrule(lr){4-6}
Selector & Spearman & Jacc & LeRF$\downarrow$ & MoRF$\uparrow$ & Gap$\uparrow$ \\
\midrule
\method{MeanAct.}  & .995 & .877 & \cellcolor{celllred} 65{,}913 & \cellcolor{cellgreen} 3.11e8 & \cellcolor{cellgreen} 3.11e8 \\ 
\method{Wanda}     & .9997 & .975 & \cellcolor{cellred} 1.95e7 & \cellcolor{cellred} 9.67e6 & \cellcolor{cellred} $-$9.78e6 \\ 
\method{IG}        & .982 & .849 & \cellcolor{celllgreen} 228.8 & \cellcolor{celllred} 2.68e7 & \cellcolor{celllred} 2.68e7 \\ 
\method{LRP}       & .959 & .760 & \cellcolor{celllgreen} 724.3 & \cellcolor{celllgreen} 3.74e7 & \cellcolor{celllgreen} 3.74e7 \\ 
\consensus{}       & .976 & .818 & \cellcolor{cellgreen} 187.2 & \cellcolor{celllgreen} 2.83e7 & \cellcolor{celllgreen} 2.83e7 \\ 
\bottomrule
\end{tabular}
\end{table}

\begin{figure}[ht]
  \centering
  \includegraphics[width=0.78\linewidth]{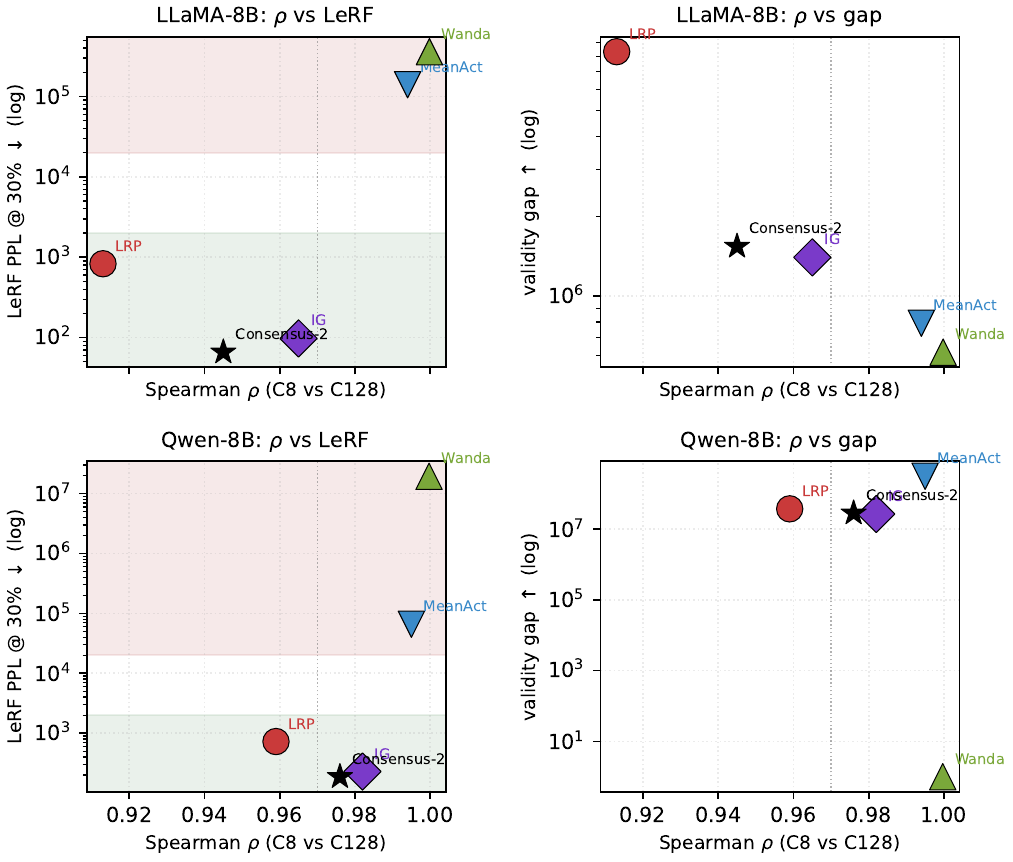}
  \caption{Stability vs.\ validity on LLaMA-3.1-8B (top) and Qwen3-8B
    (bottom). $x$: Spearman $\rho$ (C8 vs C128); $y$: LeRF PPL@30\%
    (log). In the LeRF panels the green band marks the low-LeRF (faithful,
    desirable) region and the red band the high-LeRF (invalid) region:
    MeanActivation and Wanda are ``stable but invalid'' (red), while LRP/IG
    are less stable but land orders of magnitude lower (green).}
  \label{fig:scatter}
\end{figure}

% ============================================================
\subsection{Pairwise Jaccard agreement matrices.}
\label{app:jaccard_full}

\begin{figure}[ht]
  \centering
  \includegraphics[width=0.75\linewidth]{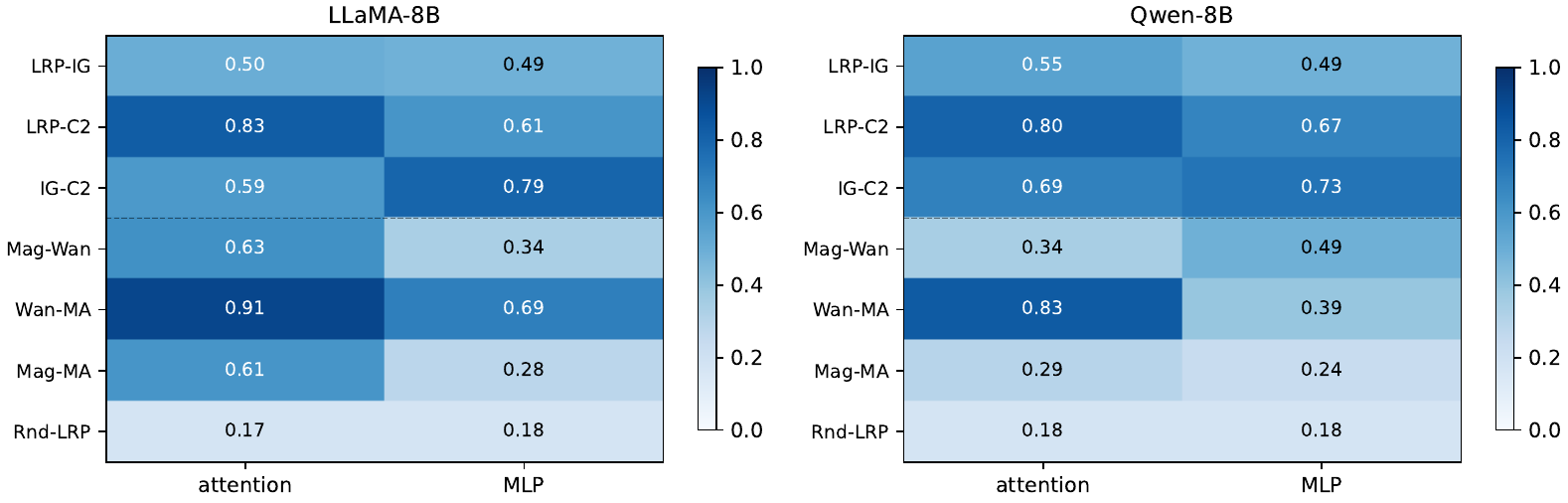}
  \caption{Pairwise Jaccard agreement at $30\%$ masking for seven
    key selector pairs (rows) $\times$ \{attention, MLP\} sublayers
    (columns), on LLaMA-8B (left) and Qwen-8B (right). Darker blue
    indicates higher agreement. Top three rows (above dashed line)
    are attribution-attribution pairs (LRP-IG, LRP-C2, IG-C2);
    these generally exhibit stronger overlap than most baseline
    comparisons, although agreement strength varies across selector
    pairs. Agreement patterns vary between attention and MLP sublayers,
    but are qualitatively similar across model families, suggesting
    selector relationships are largely preserved across architectures.}
  \label{fig:jaccard}
\end{figure}

Figure~\ref{fig:jaccard} visualizes pairwise Jaccard at 30\% masking for LLaMA-8B and Qwen-8B. Tables~\ref{tab:jaccard_20}, \ref{tab:jaccard_30}, and~\ref{tab:jaccard_50} give full matrices at 20\%, 30\%, and 50\% masking (LLaMA-1B); Tables~\ref{tab:jaccard_30_8b}--\ref{tab:jaccard_30_qwen} give the corresponding 30\% matrices for LLaMA-8B and Qwen-8B.

Three patterns hold across all five matrices. (1) Within-cluster
agreement is much higher than cross-cluster: attribution-attribution
(LRP-IG, LRP-C2, IG-C2) reach Jaccard $0.49$-$0.83$ at $30\%$,
while attribution-baseline pairs sit at most $0.29$.
(2) The cross-cluster agreement is essentially at the random floor:
attribution-vs-Random is $0.13$-$0.18$, and several attribution-vs-baseline
pairs (LRP-Wanda, LRP-MeanAct on LLaMA) are \emph{below}
that floor on the MLP sublayer ($0.000$-$0.005$), indicating
attribution and weight-statistic selectors pick essentially disjoint
MLP rows. (3) Agreement is more stable for baselines than for attribution methods
across masking rates: LRP-IG MLP Jaccard on LLaMA-1B stays in the
$0.54$-$0.60$ range from 20\% to 50\%, while attention Jaccard for
the baseline cluster (Wan-MA) remains high ($0.89$-$0.96$) because
the underlying ranking barely depends on the mask threshold.

\begin{table}[ht]
\centering
\small
\setlength{\tabcolsep}{2pt}
\caption{Pairwise Jaccard at 20\% masking (LLaMA-1B). Upper: attention. Lower: MLP.}
\label{tab:jaccard_20}
\begin{tabular}{l ccccccc}
\toprule
 & Rnd & Mag & Wan & MA & LRP & IG & C2 \\
\midrule
Rnd & --- & \cellcolor{cellgreen} .160 & \cellcolor{celllgreen} .158 & \cellcolor{celllgreen} .156 & \cellcolor{celllgreen} .113 & \cellcolor{celllgreen} .085 & \cellcolor{celllgreen} .093 \\ 
Mag & \cellcolor{celllred} .089 & --- & \cellcolor{celllgreen} .591 & \cellcolor{celllgreen} .576 & \cellcolor{celllred} .072 & \cellcolor{cellred} .057 & \cellcolor{celllred} .069 \\ 
Wan & \cellcolor{cellred} .088 & \cellcolor{celllgreen} .107 & --- & \cellcolor{cellgreen} .962 & \cellcolor{celllred} .021 & \cellcolor{celllred} .067 & \cellcolor{celllred} .019 \\ 
MA  & \cellcolor{celllred} .090 & \cellcolor{celllgreen} .061 & \cellcolor{cellgreen} .594 & --- & \cellcolor{cellred} .012 & \cellcolor{celllred} .063 & \cellcolor{cellred} .013 \\ 
LRP & \cellcolor{celllgreen} .112 & \cellcolor{celllred} .033 & \cellcolor{celllred} .023 & \cellcolor{celllred} .022 & --- & \cellcolor{celllgreen} .417 & \cellcolor{cellgreen} .709 \\ 
IG  & \cellcolor{cellgreen} .118 & \cellcolor{cellred} .022 & \cellcolor{cellred} .000 & \cellcolor{celllred} .000 & \cellcolor{celllgreen} .544 & --- & \cellcolor{celllgreen} .563 \\ 
C2  & \cellcolor{celllgreen} .117 & \cellcolor{celllred} .030 & \cellcolor{celllred} .001 & \cellcolor{celllred} .000 & \cellcolor{cellgreen} .731 & \cellcolor{cellgreen} .721 & --- \\
\bottomrule
\end{tabular}
\end{table}

\begin{table}[ht]
\centering
\small
\setlength{\tabcolsep}{2pt}
\caption{Pairwise Jaccard at 30\% masking (LLaMA-1B). Upper: attention. Lower: MLP.}
\label{tab:jaccard_30}
\begin{tabular}{l ccccccc}
\toprule
 & Rnd & Mag & Wan & MA & LRP & IG & C2 \\
\midrule
Rnd & --- & \cellcolor{celllgreen} .225 & \cellcolor{celllgreen} .217 & \cellcolor{celllgreen} .214 & \cellcolor{celllgreen} .151 & \cellcolor{celllred} .141 & \cellcolor{celllgreen} .136 \\ 
Mag & \cellcolor{cellred} .160 & --- & \cellcolor{cellgreen} .655 & \cellcolor{celllgreen} .642 & \cellcolor{celllred} .076 & \cellcolor{cellred} .096 & \cellcolor{celllred} .073 \\ 
Wan & \cellcolor{celllred} .163 & \cellcolor{cellgreen} .253 & --- & \cellcolor{cellgreen} .936 & \cellcolor{celllred} .031 & \cellcolor{celllgreen} .150 & \cellcolor{celllred} .028 \\ 
MA  & \cellcolor{celllred} .165 & \cellcolor{celllgreen} .232 & \cellcolor{celllgreen} .634 & --- & \cellcolor{cellred} .018 & \cellcolor{celllred} .147 & \cellcolor{cellred} .019 \\ 
LRP & \cellcolor{celllgreen} .185 & \cellcolor{celllred} .087 & \cellcolor{celllred} .099 & \cellcolor{celllred} .095 & --- & \cellcolor{celllgreen} .428 & \cellcolor{cellgreen} .808 \\ 
IG  & \cellcolor{celllgreen} .186 & \cellcolor{cellred} .072 & \cellcolor{cellred} .005 & \cellcolor{cellred} .005 & \cellcolor{celllgreen} .543 & --- & \cellcolor{celllgreen} .505 \\ 
C2  & \cellcolor{cellgreen} .188 & \cellcolor{celllred} .096 & \cellcolor{celllred} .013 & \cellcolor{celllred} .015 & \cellcolor{cellgreen} .678 & \cellcolor{cellgreen} .770 & --- \\
\bottomrule
\end{tabular}
\end{table}

\begin{table}[ht]
\centering
\small
\setlength{\tabcolsep}{2pt}
\caption{Pairwise Jaccard at 50\% masking (LLaMA-1B). Upper: attention. Lower: MLP.}
\label{tab:jaccard_50}
\begin{tabular}{l ccccccc}
\toprule
 & Rnd & Mag & Wan & MA & LRP & IG & C2 \\
\midrule
Rnd & --- & \cellcolor{celllgreen} .350 & \cellcolor{celllgreen} .334 & \cellcolor{celllgreen} .333 & \cellcolor{celllgreen} .214 & \cellcolor{celllgreen} .273 & \cellcolor{celllgreen} .233 \\ 
Mag & \cellcolor{cellred} .329 & --- & \cellcolor{cellgreen} .776 & \cellcolor{celllgreen} .784 & \cellcolor{celllred} .087 & \cellcolor{cellred} .236 & \cellcolor{celllred} .132 \\ 
Wan & \cellcolor{celllred} .334 & \cellcolor{cellgreen} .623 & --- & \cellcolor{cellgreen} .890 & \cellcolor{celllred} .051 & \cellcolor{celllred} .254 & \cellcolor{celllred} .118 \\ 
MA  & \cellcolor{celllred} .334 & \cellcolor{celllgreen} .569 & \cellcolor{celllgreen} .757 & --- & \cellcolor{cellred} .036 & \cellcolor{celllred} .257 & \cellcolor{cellred} .105 \\ 
LRP & \cellcolor{cellgreen} .360 & \cellcolor{celllred} .298 & \cellcolor{celllred} .294 & \cellcolor{celllred} .306 & --- & \cellcolor{celllgreen} .448 & \cellcolor{cellgreen} .741 \\ 
IG  & \cellcolor{celllgreen} .348 & \cellcolor{cellred} .271 & \cellcolor{cellred} .151 & \cellcolor{cellred} .166 & \cellcolor{celllgreen} .599 & --- & \cellcolor{celllgreen} .616 \\ 
C2  & \cellcolor{celllgreen} .356 & \cellcolor{celllred} .281 & \cellcolor{celllred} .213 & \cellcolor{celllred} .227 & \cellcolor{cellgreen} .764 & \cellcolor{cellgreen} .794 & --- \\
\bottomrule
\end{tabular}
\end{table}

% (Full Jaccard matrices for LLaMA-8B and Qwen-8B are shown in
%  Tables~\ref{tab:jaccard_30_8b} and~\ref{tab:jaccard_30_qwen} below.)

\begin{table}[ht]
\centering
\small
\setlength{\tabcolsep}{2pt}
\caption{Pairwise Jaccard at 30\% masking (LLaMA-8B). Upper: attention. Lower: MLP.}
\label{tab:jaccard_30_8b}
\begin{tabular}{l ccccccc}
\toprule
 & Rnd & Mag & Wan & MA & LRP & IG & C2 \\
\midrule
Rnd & --- & \cellcolor{celllgreen} .217 & \cellcolor{celllgreen} .222 & \cellcolor{celllgreen} .213 & \cellcolor{celllgreen} .173 & \cellcolor{celllgreen} .152 & \cellcolor{celllgreen} .156 \\ 
Mag & \cellcolor{celllred} .162 & --- & \cellcolor{celllgreen} .626 & \cellcolor{celllgreen} .609 & \cellcolor{celllred} .075 & \cellcolor{cellred} .087 & \cellcolor{celllred} .065 \\ 
Wan & \cellcolor{cellred} .158 & \cellcolor{cellgreen} .337 & --- & \cellcolor{cellgreen} .911 & \cellcolor{celllred} .032 & \cellcolor{celllred} .097 & \cellcolor{celllred} .015 \\ 
MA  & \cellcolor{celllred} .163 & \cellcolor{celllgreen} .284 & \cellcolor{cellgreen} .695 & --- & \cellcolor{cellred} .008 & \cellcolor{celllred} .096 & \cellcolor{cellred} .007 \\ 
LRP & \cellcolor{celllgreen} .177 & \cellcolor{celllred} .075 & \cellcolor{celllred} .129 & \cellcolor{celllred} .123 & --- & \cellcolor{celllgreen} .502 & \cellcolor{cellgreen} .826 \\ 
IG  & \cellcolor{cellgreen} .183 & \cellcolor{cellred} .038 & \cellcolor{cellred} .000 & \cellcolor{cellred} .000 & \cellcolor{celllgreen} .488 & --- & \cellcolor{celllgreen} .592 \\ 
C2  & \cellcolor{celllgreen} .182 & \cellcolor{celllred} .049 & \cellcolor{celllred} .002 & \cellcolor{celllred} .002 & \cellcolor{cellgreen} .607 & \cellcolor{cellgreen} .794 & --- \\
\bottomrule
\end{tabular}
\end{table}

\begin{table}[ht]
\centering
\small
\setlength{\tabcolsep}{2pt}
\caption{Pairwise Jaccard at 30\% masking (Qwen-8B). Upper: attention. Lower: MLP.}
\label{tab:jaccard_30_qwen}
\begin{tabular}{l ccccccc}
\toprule
 & Rnd & Mag & Wan & MA & LRP & IG & C2 \\
\midrule
Rnd & --- & \cellcolor{celllgreen} .205 & \cellcolor{celllgreen} .243 & \cellcolor{celllgreen} .233 & \cellcolor{cellred} .178 & \cellcolor{celllred} .145 & \cellcolor{celllred} .163 \\ 
Mag & \cellcolor{celllred} .165 & --- & \cellcolor{celllgreen} .343 & \cellcolor{celllgreen} .294 & \cellcolor{celllgreen} .291 & \cellcolor{celllgreen} .268 & \cellcolor{celllgreen} .287 \\ 
Wan & \cellcolor{cellred} .139 & \cellcolor{cellgreen} .491 & --- & \cellcolor{cellgreen} .835 & \cellcolor{celllred} .228 & \cellcolor{celllred} .113 & \cellcolor{celllred} .178 \\ 
MA  & \cellcolor{celllred} .147 & \cellcolor{celllgreen} .238 & \cellcolor{cellgreen} .393 & --- & \cellcolor{celllred} .208 & \cellcolor{cellred} .102 & \cellcolor{celllred} .163 \\ 
LRP & \cellcolor{celllgreen} .176 & \cellcolor{celllred} .191 & \cellcolor{celllred} .167 & \cellcolor{celllred} .084 & --- & \cellcolor{celllgreen} .553 & \cellcolor{cellgreen} .802 \\ 
IG  & \cellcolor{cellgreen} .186 & \cellcolor{celllred} .177 & \cellcolor{cellred} .102 & \cellcolor{cellred} .031 & \cellcolor{celllgreen} .488 & --- & \cellcolor{celllgreen} .687 \\ 
C2  & \cellcolor{celllgreen} .181 & \cellcolor{cellred} .176 & \cellcolor{celllred} .119 & \cellcolor{celllred} .045 & \cellcolor{cellgreen} .672 & \cellcolor{cellgreen} .735 & --- \\
\bottomrule
\end{tabular}
\end{table}

\begin{table}[ht]
\centering
\small
\setlength{\tabcolsep}{2pt}
\caption{Spearman correlation between selector importance scores
  (all neurons, LLaMA-8B).}
\label{tab:method_spearman_8b}
\begin{tabular}{l ccccccc}
\toprule
 & Rnd & Mag & Wan & MA & LRP & IG & C2 \\
\midrule
Rnd & \cellcolor{cellgreen} 1.00 & \cellcolor{celllgreen} .00 & \cellcolor{celllgreen} .00 & \cellcolor{celllgreen} .00 & \cellcolor{celllgreen} .00 & \cellcolor{celllgreen} .00 & \cellcolor{celllgreen} .00 \\ 
Mag & \cellcolor{celllgreen} .00 & \cellcolor{cellgreen} 1.00 & \cellcolor{celllgreen} .72 & \cellcolor{celllgreen} .71 & \cellcolor{celllred} $-$.30 & \cellcolor{celllred} $-$.35 & \cellcolor{celllred} $-$.36 \\ 
Wan & \cellcolor{celllgreen} .00 & \cellcolor{celllgreen} .72 & \cellcolor{cellgreen} 1.00 & \cellcolor{celllgreen} .88 & \cellcolor{cellred} $-$.43 & \cellcolor{cellred} $-$.58 & \cellcolor{cellred} $-$.56 \\ 
MA  & \cellcolor{celllgreen} .00 & \cellcolor{celllgreen} .71 & \cellcolor{celllgreen} .88 & \cellcolor{cellgreen} 1.00 & \cellcolor{celllred} $-$.36 & \cellcolor{celllred} $-$.44 & \cellcolor{celllred} $-$.44 \\ 
LRP & \cellcolor{celllgreen} .00 & \cellcolor{celllred} $-$.30 & \cellcolor{celllred} $-$.43 & \cellcolor{celllred} $-$.36 & \cellcolor{cellgreen} 1.00 & \cellcolor{celllgreen} .65 & \cellcolor{celllgreen} .91 \\ 
IG  & \cellcolor{celllgreen} .00 & \cellcolor{celllred} $-$.35 & \cellcolor{cellred} $-$.58 & \cellcolor{celllred} $-$.44 & \cellcolor{celllgreen} .65 & \cellcolor{cellgreen} 1.00 & \cellcolor{celllgreen} .91 \\ 
C2  & \cellcolor{celllgreen} .00 & \cellcolor{cellred} $-$.36 & \cellcolor{celllred} $-$.56 & \cellcolor{celllred} $-$.44 & \cellcolor{celllgreen} .91 & \cellcolor{celllgreen} .91 & \cellcolor{cellgreen} 1.00 \\ 
\bottomrule
\end{tabular}
\end{table}

\begin{table}[ht]
\centering
\small
\setlength{\tabcolsep}{2pt}
\caption{Spearman correlation between selector importance scores
  (all neurons, Qwen-8B).}
\label{tab:method_spearman_qwen}
\begin{tabular}{l ccccccc}
\toprule
 & Rnd & Mag & Wan & MA & LRP & IG & C2 \\
\midrule
Rnd & \cellcolor{cellgreen} 1.00 & \cellcolor{celllred} .00 & \cellcolor{celllgreen} .00 & \cellcolor{celllgreen} .00 & \cellcolor{celllred} .00 & \cellcolor{celllgreen} .00 & \cellcolor{celllred} .00 \\ 
Mag & \cellcolor{celllgreen} .00 & \cellcolor{cellgreen} 1.00 & \cellcolor{celllgreen} .45 & \cellcolor{celllgreen} .32 & \cellcolor{celllgreen} .04 & \cellcolor{celllgreen} .00 & \cellcolor{celllgreen} .01 \\ 
Wan & \cellcolor{celllgreen} .00 & \cellcolor{celllgreen} .45 & \cellcolor{cellgreen} 1.00 & \cellcolor{celllgreen} .77 & \cellcolor{celllred} $-$.06 & \cellcolor{celllred} $-$.26 & \cellcolor{celllred} $-$.18 \\ 
MA  & \cellcolor{celllgreen} .00 & \cellcolor{celllgreen} .32 & \cellcolor{celllgreen} .77 & \cellcolor{cellgreen} 1.00 & \cellcolor{cellred} $-$.26 & \cellcolor{cellred} $-$.43 & \cellcolor{cellred} $-$.37 \\ 
LRP & \cellcolor{celllgreen} .00 & \cellcolor{celllgreen} .04 & \cellcolor{celllred} $-$.06 & \cellcolor{celllred} $-$.26 & \cellcolor{cellgreen} 1.00 & \cellcolor{celllgreen} .75 & \cellcolor{celllgreen} .93 \\ 
IG  & \cellcolor{celllgreen} .00 & \cellcolor{celllred} .00 & \cellcolor{cellred} $-$.26 & \cellcolor{cellred} $-$.43 & \cellcolor{celllgreen} .75 & \cellcolor{cellgreen} 1.00 & \cellcolor{celllgreen} .93 \\ 
C2  & \cellcolor{celllgreen} .00 & \cellcolor{celllred} .01 & \cellcolor{celllred} $-$.18 & \cellcolor{celllred} $-$.37 & \cellcolor{celllgreen} .93 & \cellcolor{celllgreen} .93 & \cellcolor{cellgreen} 1.00 \\ 
\bottomrule
\end{tabular}
\end{table}

% ============================================================
\subsection{Depth profiles and selector-score correlations.}
\label{app:depth_full}

\begin{figure}[ht]
  \centering
  \includegraphics[width=0.9\linewidth]{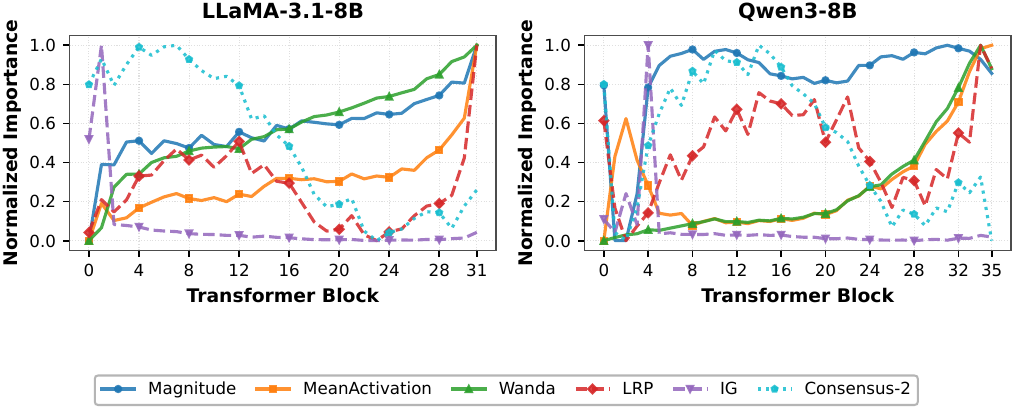}
  \caption{Layer-wise importance depth profiles for LLaMA-3.1-8B (left) and
    Qwen3-8B (right). Each line shows a selector's normalized mean importance
    per transformer block. Attribution-based selectors (LRP, IG, and
    \consensus{}) tend to emphasize earlier and intermediate layers,
    whereas magnitude-based selectors (Magnitude, Wanda) place increasing
    importance on later layers near the model output. MeanActivation
    exhibits a comparatively flatter profile on LLaMA-3.1-8B, consistent
    with its observed stability characteristics (Table~\ref{tab:stability}).
    These differing layer allocation patterns help explain the need for
    layer-matched controls (Table~\ref{tab:consensus_8b}).}
  \label{fig:depth_profile}
\end{figure}

\begin{figure}[ht]
  \centering
  \includegraphics[width=0.8\linewidth]{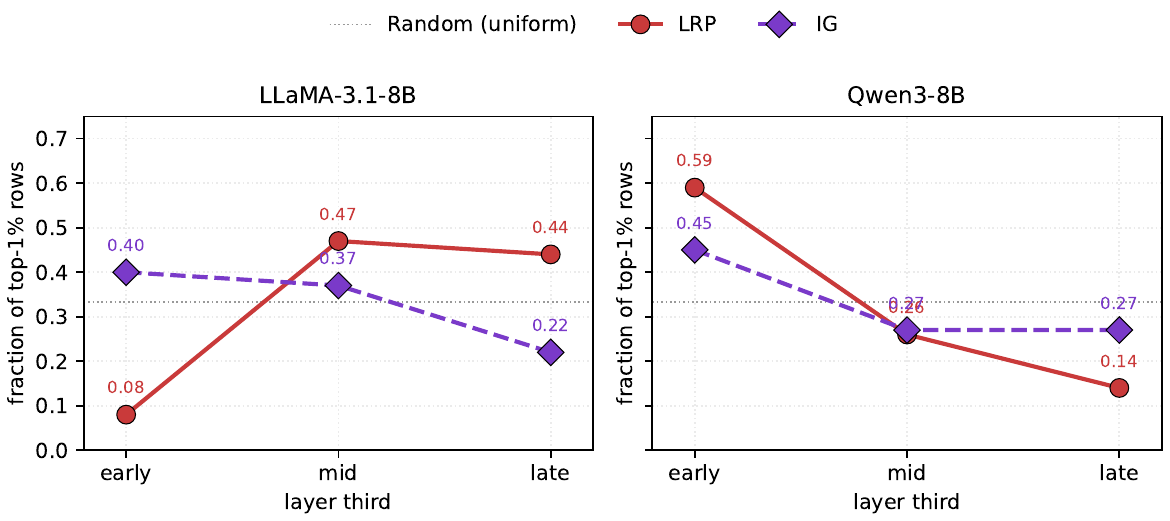}
  \caption{Layer distribution of top-1\% $|C_{\text{comply}}|$ rows
    (hate domain) for LRP and IG on LLaMA-3.1-8B and Qwen3-8B, split
    by early / mid / late layer thirds. On LLaMA, LRP is late-heavy
    ($44\%$ late) and IG is early-heavy ($40\%$ early); on Qwen3,
    both methods route to early layers but LRP is sharper ($59\%$
    early). Random would be uniform $0.33/0.33/0.33$.}
  \label{fig:layer_dist}
\end{figure}

Figure~\ref{fig:depth_profile} shows per-block importance profiles for the general LM-level audit, revealing that attribution methods (LRP, IG, \consensus{}) concentrate importance in early-to-mid layers while magnitude-based methods concentrate in late layers, explaining why layer-matched controls are necessary (Table~\ref{tab:consensus_8b}). Figure~\ref{fig:layer_dist} shows how top-1\% $|C_{\text{comply}}|$ rows are distributed across early/mid/late layer thirds for LRP and IG on LLaMA-3.1-8B and Qwen3-8B, explaining the cross-architecture method reversal (§\ref{sec:cross}).

Tables~\ref{tab:method_spearman_8b}, \ref{tab:method_spearman_qwen}, and~\ref{tab:method_spearman} give Spearman rank correlations between all selector pairs across LLaMA-8B, Qwen-8B, and LLaMA-1B respectively.
Attribution methods anti-correlate with magnitude and activation baselines across all three models (e.g.\ LRP-Wanda $= -0.43$, IG-Wanda $= -0.58$ on LLaMA-8B): the two families rank neurons in opposite order, mechanistically explaining why they select disjoint row sets and why layer-matched controls are necessary.

% Spearman correlation between methods (all neurons)
\begin{table}[ht]
\centering
\small
\setlength{\tabcolsep}{2pt}
\caption{Spearman correlation between selector importance scores
  (all neurons, LLaMA-1B). Measures how similarly two selectors rank
  the full neuron population.}
\label{tab:method_spearman}
\begin{tabular}{l ccccccc}
\toprule
 & Rnd & Mag & Wan & MA & LRP & IG & C2 \\
\midrule
Rnd & \cellcolor{cellgreen} 1.00 & \cellcolor{celllgreen} .00 & \cellcolor{celllgreen} .00 & \cellcolor{celllgreen} .00 & \cellcolor{celllgreen} .00 & \cellcolor{celllgreen} .00 & \cellcolor{celllgreen} .00 \\ 
Mag & \cellcolor{celllgreen} .00 & \cellcolor{cellgreen} 1.00 & \cellcolor{celllgreen} .73 & \cellcolor{celllgreen} .66 & \cellcolor{celllred} $-$.22 & \cellcolor{celllred} $-$.20 & \cellcolor{celllred} $-$.23 \\ 
Wan & \cellcolor{celllgreen} .00 & \cellcolor{celllgreen} .73 & \cellcolor{cellgreen} 1.00 & \cellcolor{celllgreen} .89 & \cellcolor{cellred} $-$.37 & \cellcolor{cellred} $-$.45 & \cellcolor{cellred} $-$.44 \\ 
MA  & \cellcolor{celllgreen} .00 & \cellcolor{celllgreen} .66 & \cellcolor{celllgreen} .89 & \cellcolor{cellgreen} 1.00 & \cellcolor{celllred} $-$.29 & \cellcolor{celllred} $-$.35 & \cellcolor{celllred} $-$.35 \\ 
LRP & \cellcolor{celllgreen} .00 & \cellcolor{celllred} $-$.22 & \cellcolor{celllred} $-$.37 & \cellcolor{celllred} $-$.29 & \cellcolor{cellgreen} 1.00 & \cellcolor{celllgreen} .70 & \cellcolor{celllgreen} .92 \\ 
IG  & \cellcolor{celllgreen} .00 & \cellcolor{celllred} $-$.20 & \cellcolor{cellred} $-$.45 & \cellcolor{celllred} $-$.35 & \cellcolor{celllgreen} .70 & \cellcolor{cellgreen} 1.00 & \cellcolor{celllgreen} .92 \\ 
C2  & \cellcolor{celllgreen} .00 & \cellcolor{cellred} $-$.23 & \cellcolor{celllred} $-$.44 & \cellcolor{celllred} $-$.35 & \cellcolor{celllgreen} .92 & \cellcolor{celllgreen} .92 & \cellcolor{cellgreen} 1.00 \\ 
\bottomrule
\end{tabular}
\end{table}

% ============================================================
\subsection{Consensus controls (all models).}
\label{app:consensus_full}

\begin{figure}[ht]
  \centering
  \includegraphics[width=0.85\linewidth]{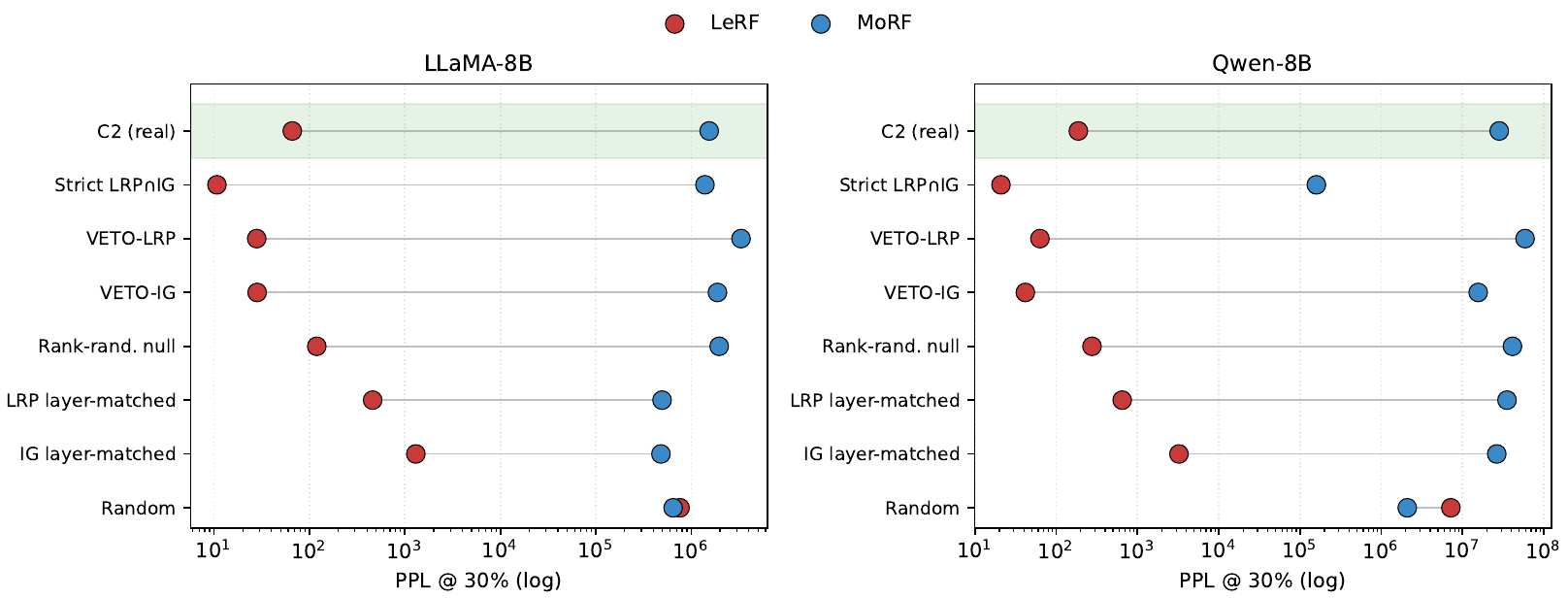}
  \caption{Consensus controls across families at $30\%$ masking
      as a dumbbell plot. Each row is one mask variant; the LeRF PPL
      (red) and MoRF PPL (blue) are connected by a gray line. Wider
      span indicates stronger separation between least- and most-relevant
      rows. Real \consensus{} performs strongly, but veto and
      rank-randomized variants show that the benefit is not explained
      by depth allocation or by generic rank averaging alone; disagreement-
      aware variants can outperform real \consensus{} on some metrics.}
  \label{fig:consensus_controls}
\end{figure}

Figure~\ref{fig:consensus_controls} visualizes the cross-family dumbbell; Tables~\ref{tab:consensus_8b}--\ref{tab:consensus_qwen} give full results for LLaMA-8B and Qwen-8B.

The control suite probes four candidate explanations for the
\consensus{} advantage. \emph{Layer-matched} controls copy each
method's per-layer mask count but draw row identities uniformly,
testing whether depth allocation alone explains the win. The
\emph{strict intersection} (LRP$\cap$IG) keeps rows ranked
bottom-$k\%$ by \emph{both} methods, testing whether
agreement, rather than rank averaging, is the source of safety.
The \emph{rank-randomized null} fixes LRP and permutes IG's rank
list before Borda averaging ($5$ seeds), testing whether the
mechanical act of combining two rank lists drives the gain. The
\emph{veto} variants take rows in one method's top-$k$ but not the
other's: \method{VETO-LRP} keeps LRP's exclusive picks (rows IG
ranks as still-needed but LRP overrides), and \method{VETO-IG}
keeps IG's exclusive picks, testing whether the rows where the two
methods disagree carry causal signal.

On LLaMA-8B, \method{VETO-LRP} achieves LeRF$=28.0$ and MoRF$=3.32\times10^6$, outperforming real \consensus{} on both metrics ($66.1$ / $1.54\times10^6$); rows where the two methods disagree carry stronger causal signal than the agreed set, pointing to veto-based aggregation as a direction for future work.
On Qwen3-8B, the strict LRP$\cap$IG intersection collapses catastrophically (MoRF$=159\text{k}$ vs.\ $28.3\text{M}$ for real \consensus{}); the rank-randomized null beats real \consensus{} on MoRF ($4.12\times10^7$ vs.\ $2.83\times10^7$), indicating that on Qwen the Borda rank-averaging step is not the source of the aggregate's gain; \method{VETO-LRP} achieves $5.89\times10^7$, the strongest variant on Qwen.

% \begin{table}[ht]
% \centering
% \caption{Consensus controls on LLaMA-3.1-8B at $30\%$ masking.
%   Each control isolates one potential confound.}
% \label{tab:consensus_8b_main}
% \begin{tabular}{l rrrr}
% \toprule
% Mask & LeRF$\downarrow$ & MoRF$\uparrow$ & Gap$\uparrow$ & Tests \\
% \midrule
% \consensus{} (real)       & \cellcolor{celllgreen} 66.1 & \cellcolor{celllgreen} 1.54e6 & \cellcolor{celllgreen} 1.54e6 & agreement \\
% LRP layer-matched         & \cellcolor{celllred} 459.3 & \cellcolor{celllred} 494{,}193 & \cellcolor{celllred} 493{,}734 & layer alloc. \\
% IG layer-matched          & \cellcolor{celllred} 1{,}301 & \cellcolor{cellred} 481{,}082 & \cellcolor{celllred} 479{,}781 & layer alloc. \\
% Strict LRP$\cap$IG        & \cellcolor{cellgreen} 10.7 & \cellcolor{celllgreen} 1.39e6 & \cellcolor{celllgreen} 1.39e6 & intersection \\
% Rank-rand.\ null          & \cellcolor{celllgreen} 119.2 & \cellcolor{cellgreen} 1.96e6 & \cellcolor{cellgreen} 1.96e6 & rank avg. \\
% \method{Random}           & \cellcolor{cellred} 762{,}286 & \cellcolor{celllred} 645{,}432 & \cellcolor{cellred} $-$116{,}854 & no-info null \\
% \bottomrule
% \end{tabular}
% \end{table}

\begin{table}[ht]
\centering
\caption{Consensus controls at 30\% masking (LLaMA-8B).}
\label{tab:consensus_8b}
\begin{tabular}{l rrrr}
\toprule
Mask & LeRF$\downarrow$ & MoRF$\uparrow$ & Gap$\uparrow$ & Mechanism \\
\midrule
\consensus{} (real)    & \cellcolor{celllgreen} 66.08 & \cellcolor{celllgreen} 1.54e6 & \cellcolor{celllgreen} 1.54e6 & agreement \\
LRP layer-matched      & \cellcolor{celllred} 459.26 & \cellcolor{celllred} 494{,}193 & \cellcolor{celllred} 493{,}734 & layer alloc. \\
IG layer-matched       & \cellcolor{celllred} 1{,}301 & \cellcolor{cellred} 481{,}082 & \cellcolor{celllred} 479{,}781 & layer alloc. \\
Strict LRP$\cap$IG     & \cellcolor{cellgreen} 10.73 & \cellcolor{celllred} 1.39e6 & \cellcolor{celllred} 1.39e6 & intersection \\
VETO-LRP               & \cellcolor{celllgreen} 28.02 & \cellcolor{cellgreen} 3.32e6 & \cellcolor{cellgreen} 3.32e6 & veto \\
VETO-IG                & \cellcolor{celllgreen} 28.29 & \cellcolor{celllgreen} 1.88e6 & \cellcolor{celllgreen} 1.88e6 & veto \\
Rank-rand.\ null       & \cellcolor{celllred} 119.23 & \cellcolor{celllgreen} 1.96e6 & \cellcolor{celllgreen} 1.96e6 & rank avg. \\
\method{Random}        & \cellcolor{cellred} 762{,}286 & \cellcolor{celllred} 645{,}432 & \cellcolor{cellred} $-$116{,}854 & no-info null \\
\bottomrule
\end{tabular}
\end{table}

\begin{table}[ht]
\centering
\caption{Consensus controls at 30\% masking (Qwen-8B).}
\label{tab:consensus_qwen}
\begin{tabular}{l rrrr}
\toprule
Mask & LeRF$\downarrow$ & MoRF$\uparrow$ & Gap$\uparrow$ & Mechanism \\
\midrule
\consensus{} (real)    & \cellcolor{celllgreen} 187.2 & \cellcolor{celllgreen} 2.83e7 & \cellcolor{celllgreen} 2.83e7 & agreement \\
LRP layer-matched      & \cellcolor{celllred} 648.7 & \cellcolor{celllgreen} 3.53e7 & \cellcolor{celllgreen} 3.53e7 & layer alloc. \\
IG layer-matched       & \cellcolor{celllred} 3{,}239 & \cellcolor{celllred} 2.64e7 & \cellcolor{celllred} 2.64e7 & layer alloc. \\
Strict LRP$\cap$IG     & \cellcolor{cellgreen} 20.9 & \cellcolor{cellred} 159{,}199 & \cellcolor{celllred} 159{,}178 & intersection \\
VETO-LRP               & \cellcolor{celllgreen} 63.1 & \cellcolor{cellgreen} 5.89e7 & \cellcolor{cellgreen} 5.89e7 & veto \\
VETO-IG                & \cellcolor{celllgreen} 41.7 & \cellcolor{celllred} 1.56e7 & \cellcolor{celllred} 1.56e7 & veto \\
Rank-rand.\ null       & \cellcolor{celllred} 275.5 & \cellcolor{celllgreen} 4.12e7 & \cellcolor{celllgreen} 4.12e7 & rank avg. \\
\method{Random}        & \cellcolor{cellred} 7.19e6 & \cellcolor{celllred} 2.09e6 & \cellcolor{cellred} $-$5.10e6 & no-info null \\
\bottomrule
\end{tabular}
\end{table}

% ============================================================
\textbf{\consensus{} rank distribution.}
\label{app:rank_dist}

Table~\ref{tab:rank_dist_c2} reports, for each masking rate, the fraction of \consensus{}-selected neurons appearing in the bottom-$k$\% of both LRP and IG simultaneously versus only one method.

\begin{table}[ht]
\centering
\small
\setlength{\tabcolsep}{4pt}
\caption{Consensus-2 rank distribution (LLaMA-8B). For neurons pruned by
  \consensus{} at rate $k$: fraction also in bottom-$k$\% of both LRP
  and IG simultaneously (``Both''), exactly one of the two (``Only-1''),
  or neither (``Neither'': neurons outside either single-method's
  bottom-$k$\% individually but elevated into the Borda bottom-$k$\%
  by combined rank averaging).}
\label{tab:rank_dist_c2}
\begin{tabular}{l rrr}
\toprule
Rate & Both agree & Only-1 & Neither \\
\midrule
10\% & \cellcolor{cellred} 52.2\% & \cellcolor{celllgreen} 38.2\% & \cellcolor{cellgreen} 9.6\% \\ 
20\% & \cellcolor{celllred} 57.0\% & \cellcolor{cellgreen} 39.5\% & \cellcolor{celllgreen} 3.5\% \\ 
30\% & \cellcolor{celllgreen} 65.9\% & \cellcolor{celllred} 32.8\% & \cellcolor{celllred} 1.3\% \\ 
50\% & \cellcolor{cellgreen} 75.8\% & \cellcolor{cellred} 24.2\% & \cellcolor{cellred} 0.0\% \\ 
\bottomrule
\end{tabular}
\end{table}

% (Consensus rank distribution for Qwen-8B follows the same format.)

At $30\%$ masking, $65.9\%$ of \consensus{}-selected neurons fall in the bottom-$30\%$ of \emph{both} LRP and IG independently, confirming the aggregate's advantage comes from genuine cross-method agreement rather than Borda arithmetic elevating neurons neither method individually identifies.

% ============================================================
\clearpage
\section{Diagnostic Analyses}
\label{app:diagnostics}

\textbf{Dangerous false negatives : }
\label{app:false_neg_full}
Table~\ref{tab:false_neg_full} reports the dangerous false-negative rate (fraction of neurons scoring above the $2\times$ dense PPL danger threshold that are not selected by the mask) at 30\% masking for each selector.

\method{IG} and \consensus{} achieve zero dangerous false negatives.
\method{LRP} misses one dangerous group ($1.0\%$ of dangerous neurons),
while \method{Wanda} and \method{MeanAct.} miss four dangerous groups,
corresponding to $41.4\%$ and $35.2\%$ of dangerous neurons respectively.

\begin{table}[ht]
\centering
\small
\setlength{\tabcolsep}{3pt}
\caption{Dangerous false-negative rate at 30\% masking (LLaMA-1B).
  Danger threshold: PPL $>$ 33.0 (2$\times$ dense).
  Lower = fewer dangerous false negatives.}
\label{tab:false_neg_full}
\begin{tabular}{l rrrrr}
\toprule
Selector & \#Danger & \#Total & \%Groups & \#Neurons & \%Neurons \\
\midrule
\method{LRP}       & \cellcolor{celllgreen} 1 & 32 & \cellcolor{celllgreen} 3.1\% & \cellcolor{celllgreen} 1{,}113 & \cellcolor{celllgreen} 1.0\% \\ 
\method{IG}        & \cellcolor{cellgreen} 0 & 30 & \cellcolor{cellgreen} 0.0\% & \cellcolor{cellgreen} 0 & \cellcolor{cellgreen} 0.0\% \\ 
\method{Wanda}     & \cellcolor{celllred} 4 & 32 & \cellcolor{celllred} 12.5\% & \cellcolor{cellred} 46{,}800 & \cellcolor{cellred} 41.4\% \\ 
\method{MeanAct.}  & \cellcolor{celllred} 4 & 32 & \cellcolor{celllred} 12.5\% & \cellcolor{celllred} 39{,}782 & \cellcolor{celllred} 35.2\% \\ 
\consensus{}       & \cellcolor{cellgreen} 0 & 32 & \cellcolor{cellgreen} 0.0\% & \cellcolor{cellgreen} 0 & \cellcolor{cellgreen} 0.0\% \\ 
\bottomrule
\end{tabular}
\end{table}

% ============================================================
\textbf{Domain sensitivity: WikiText-2 vs.\ C4 calibration : }
\label{app:domain_full}
Table~\ref{tab:domain_8b} measures sensitivity to calibration-corpus
choice (WikiText-2 vs.\ C4) via Spearman and Jaccard agreement,
assessing robustness to dataset shift. The table reveals a refinement
of the stability-validity dissociation: \method{Wanda} and
\method{MeanAct.} are extremely stable across corpora (Jaccard
$0.94$-$0.99$ at $30\%$) precisely because their scores are
dominated by data-independent weight statistics or bulk activation
magnitudes that barely shift between WikiText-2 and C4. Attribution
methods are less stable across corpora at the mask level (LRP $0.65$,
IG $0.77$, \consensus{} $0.75$ at $30\%$) because they extract a
calibration-specific causal signal, yet the LeRF PPL stays low
under both corpora (LRP $830 \to 732$, IG $98 \to 206$,
\consensus{} $66 \to 85$ on the C4-calibrated mask evaluated on
WikiText-2 test), so the underlying signal generalizes despite
mask-level mobility.

\begin{table}[ht]
\centering
\small
\setlength{\tabcolsep}{3pt}
\caption{Domain sensitivity (LLaMA-8B). Spearman $\rho$ and Jaccard
  between WikiText-2 and C4 calibrated scores/masks at 20\% and 30\%.
  LeRF PPL with each calibration source (evaluated on WikiText-2 test).
  Dense PPL = 10.58.}
\label{tab:domain_8b}
\begin{tabular}{l rrr rr rr}
\toprule
& & & & \multicolumn{2}{c}{LeRF@20\%} & \multicolumn{2}{c}{LeRF@30\%} \\
\cmidrule(lr){5-6}\cmidrule(lr){7-8}
Selector & Spearman & J@20 & J@30 & WT2 & C4 & WT2 & C4 \\
\midrule
\method{LRP}       & 0.9233 & 0.5751 & 0.6458 & \cellcolor{celllgreen} 340.46 & \cellcolor{celllgreen} 466.44 & \cellcolor{celllgreen} 830.38 & \cellcolor{celllgreen} 731.83 \\ 
\method{IG}        & 0.9704 & 0.6940 & 0.7725 & \cellcolor{celllgreen} 38.96 & \cellcolor{celllgreen} 69.48 & \cellcolor{celllgreen} 97.79 & \cellcolor{celllgreen} 205.51 \\ 
\consensus{}       & 0.9496 & 0.6420 & 0.7501 & \cellcolor{cellgreen} 29.95 & \cellcolor{cellgreen} 40.90 & \cellcolor{cellgreen} 66.08 & \cellcolor{cellgreen} 85.13 \\ 
\method{MeanAct.}  & 0.9933 & 0.9614 & 0.9415 & \cellcolor{celllred} 42{,}793 & \cellcolor{celllred} 96{,}180 & \cellcolor{celllred} 139{,}874 & \cellcolor{celllred} 264{,}304 \\ 
\method{Wanda}     & 0.9998 & 0.9857 & 0.9932 & \cellcolor{cellred} 148{,}982 & \cellcolor{cellred} 168{,}699 & \cellcolor{cellred} 360{,}083 & \cellcolor{cellred} 334{,}291 \\ 
\bottomrule
\end{tabular}
\end{table}

\textbf{Random baseline seed variance : }
\label{app:random_seeds}
Table~\ref{tab:random_seeds} confirms that the \method{Random}
selector's LeRF and MoRF PPL is stable across three random seeds.
Despite a seed-to-seed std of $\approx 30\%$ around the mean, the
Random baseline sits at LeRF $548\text{k}$ on LLaMA-1B against
\method{LRP}'s $288$, a separation of three orders of magnitude
that no plausible seed variance could close, so the seed-averaging
is conservative for the headline conclusion.

\begin{table}[ht]
\centering
\small
\setlength{\tabcolsep}{4pt}
\caption{Random baseline PPL across 3 LeRF seeds and 3 MoRF seeds
  (LLaMA-1B, 30\% masking). Reports mean $\pm$ std.}
\label{tab:random_seeds}
\begin{tabular}{l rr}
\toprule
& LeRF@30 & MoRF@30 \\
\midrule
Seed 42  & \cellcolor{cellred} 707{,}057 & \cellcolor{celllred} 278{,}836 \\ 
Seed 43  & \cellcolor{celllred} 578{,}339 & \cellcolor{cellred} 231{,}310 \\ 
Seed 44  & \cellcolor{cellgreen} 359{,}560 & \cellcolor{cellgreen} 522{,}863 \\ 
\midrule
Mean $\pm$ std & \cellcolor{celllgreen} 548{,}318 $\pm$ 175{,}683 & \cellcolor{celllgreen} 344{,}336 $\pm$ 156{,}424 \\ 
\bottomrule
\end{tabular}
\end{table}

\begin{figure}[pt]
  \centering
  \includegraphics[width=0.62\linewidth]{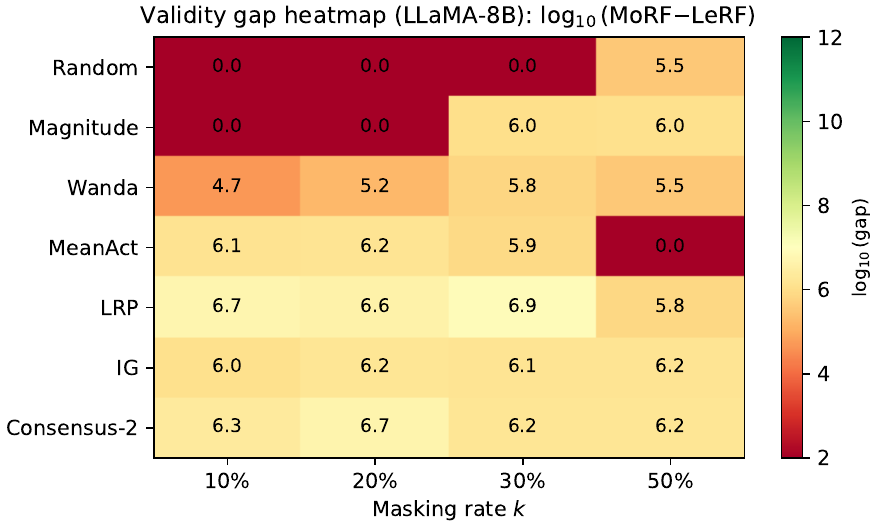}
  \caption{Validity gap heatmap, LLaMA-8B. Rows: selectors. Columns:
    masking rates (10/20/30/50\%). Color: $\log_{10}(\text{MoRF PPL}
    - \text{LeRF PPL})$; higher (greener) means the selector
    separates safe from important neurons more strongly at that
    rate. Attribution selectors (bottom three rows) maintain a
    large positive gap across rates; non-attribution baselines flip
    or saturate.}
  \label{fig:gap_heatmap}
\end{figure}

% ============================================================
\newpage
\textbf{Validity gap heatmap : }
\label{app:gap_heatmap}
Figure~\ref{fig:gap_heatmap} shows the validity gap ($\log_{10}(\text{MoRF PPL} - \text{LeRF PPL})$) for LLaMA-8B across all selectors at four representative masking rates (10/20/30/50\%).

% ============================================================
\clearpage
\section{Contrastive Refusal Editing: Full Numerics and Analyses}
\label{app:contrastive_full}

\textbf{Evaluation protocol.}
\label{app:eval_protocol}

\textbf{Refusal classifier.}
Every model response is scored by
\texttt{ProtectAI/distilroberta-base-rejection-v1}, a DistilRoBERTa
classifier fine-tuned to label responses as \emph{refusal} or
\emph{compliance}. The model generates greedily (max 200 tokens) to a
zero-shot prompt; the classifier produces a binary label from the full
response string. No sampling is used; evaluation is deterministic.

\rev{\textbf{Sampling confidence intervals.}
All refusal rates are proportions over fixed prompt sets (CAST: $500$
malign / $500$ benign; OR-Bench-Hard: $1{,}000$), so sampling
uncertainty is computable in closed form. We report Wilson
$95\%$ intervals in Table~\ref{tab:wilson}; the maximum half-width is
$\pm0.044$ at $n{=}500$ (and $\pm0.031$ at $n{=}1000$), an order of
magnitude smaller than the attribution-vs-control gaps in
Table~\ref{tab:faithfulness}. These quantify sampling uncertainty
only, not judge bias or run-to-run variance.}

\textbf{CAST-malign and CAST-benign.}
The CAST test set~\cite{lee2025cast} contains 500
held-out \emph{risk} prompts (harmful requests across 5 domains:
hate, crime, adult, medical, legal) and 500 matched \emph{base}
prompts (semantically similar but benign). CAST-malign = fraction of
risk prompts where the classifier outputs \emph{refusal};
CAST-benign = fraction of base prompts where the classifier outputs
\emph{refusal} (i.e., over-refusal on benign content). Each is a
proportion in $[0,1]$.

\textbf{SorryBench.}
SorryBench~\cite{xie2024sorrybench} contains 450 test questions
spanning 45 harm categories. A model ``passes'' if it refuses;
we report the fraction of prompts refused. Unedited instruction-tuned
models vary from $\approx 0.20$ (medical on LLaMA-3.2-1B) to
$1.00$ (legal on LLaMA-3.2-3B, a ceiling effect). Higher = stronger
refusal on harder adversarial prompts.

\textbf{OR-Bench-Hard.}
OR-Bench-Hard~\cite{cui2024orbench} contains 1{,}000 prompts that
\emph{appear} harmful but are benign (e.g., questions about chemistry
phrased provocatively). A model ``fails'' if it refuses these
unnecessarily; we report the over-refusal fraction. Baseline
instruction-tuned models sit near $0.30$; aggressive row-zeroing
raises this to $0.57$-$0.88$ for attribution selectors, indicating
the specificity cost at strong operating points.

\textbf{Downstream utility.}
We run the pre-edited and post-edited checkpoints through
\texttt{lm-eval-harness}~\cite{eval-harness} in zero-shot mode on
MMLU ($57$ tasks), GSM8K, and IFEval. We report the absolute
post-minus-pre difference ($\Delta$MMLU, $\Delta$GSM8K, $\Delta$IFEval
in accuracy / solve-rate / instruction-following points). Scores near
$0$ indicate the edit preserved general-purpose capability; large
negative values indicate capability damage from over-aggressive masking.

\textbf{Per-selector OOD and utility profile.}

\begin{figure}[ht]
  \centering
  \includegraphics[width=0.75\linewidth]{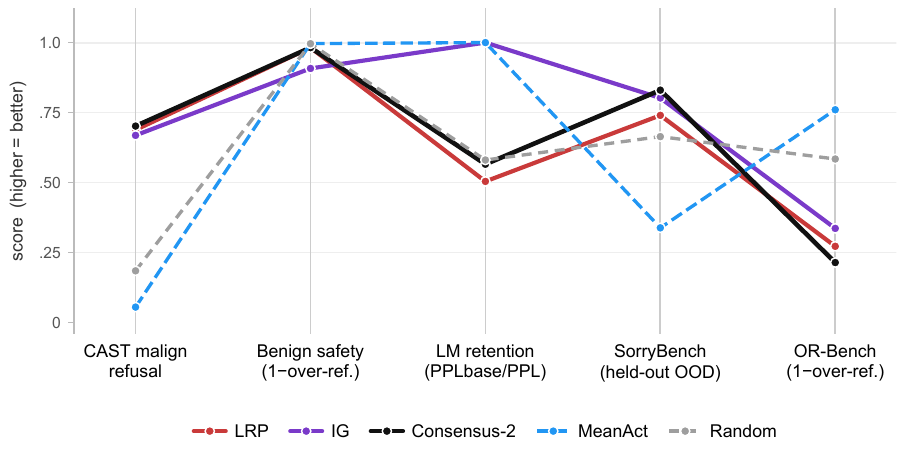}
  \caption{Per-selector OOD profile on LLaMA-3.1-8B-Instruct
    (means across 5 harm domains). Parallel coordinates; every axis
    in $[0,1]$, higher = better. LRP, IG, and \consensus{} score
    high on malign refusal and SorryBench (held-out OOD) while
    maintaining benign safety; \method{MeanAct.} starts near zero
    on malign refusal (fails to install refusal) but scores
    perfectly on benign and LM retention; \method{Random} falls
    between both. $\Delta$MMLU / $\Delta$GSM8K / $\Delta$IFEval
    numerics in Table~\ref{tab:utility_full}.}
  \label{fig:ood_utility}
\end{figure}

\begin{figure}[ht]
  \centering
  \includegraphics[width=0.85\linewidth]{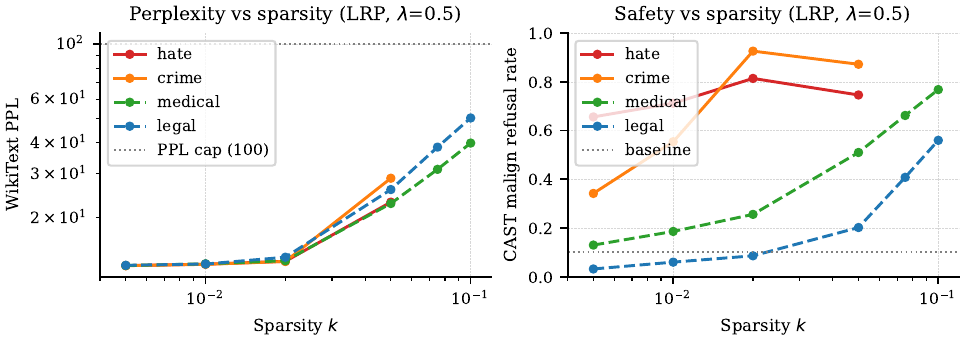}
  \caption{Left: WikiText PPL vs.\ sparsity $k$ for hate, crime,
    medical, legal (LRP, $\lambda = 0.5$, LLaMA-3.1-8B). Medical
    and legal PPL diverges sharply past $k = 0.05$ (40 and 50 at
    $k = 0.10$) while hate and crime stay below 24 at $k = 0.05$.
    Right: CAST malign refusal vs.\ $k$. Medical/legal need $k \ge
    0.05$ to cross the useful-refusal threshold, the same region
    where PPL diverges.}
  \label{fig:ppl_vs_k}
\end{figure}

\textbf{Medical/legal collapse: sparsity-driven.}
\rev{Medical and legal require $k \ge 0.05$-$0.10$ to install refusal,
but WikiText PPL diverges sharply past $k = 0.05$
(Figure~\ref{fig:ppl_vs_k}); the collapse is driven by the sparsity
these low-signal-density domains demand, not by the selector, and it
appears for LRP, \consensus{}, and Random alike.}

\textbf{Harm-cluster geometry.}
\label{app:transfer}

\begin{figure}[ht]
  \centering
  \includegraphics[width=0.39\linewidth]{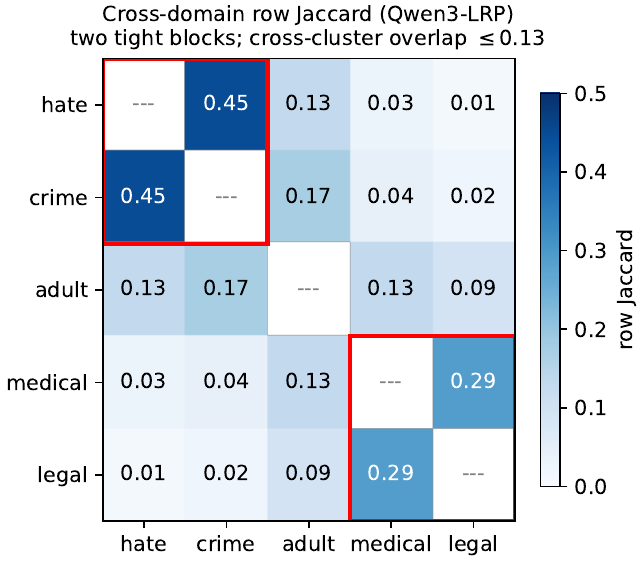}
  \caption{Cross-domain redundancy. Top-1\% row Jaccard
    (Qwen3-8B-LRP) as a heatmap; red boxes mark the two tight
    blocks \{hate, crime\} ($0.445$) and \{medical, legal\}
    ($0.287$). Cross-cluster overlap is $\le 0.13$, evidencing
    latent harm-category structure.}
  \label{fig:harm_clusters_heat}
\end{figure}

The cross-domain Jaccard overlap of top-1\% $|C_{\text{comply}}|$ rows
on Qwen3-LRP is shown in Figure~\ref{fig:harm_clusters_heat}. The
block structure is clean: \{hate, crime\} and \{medical, legal\} are
internally coupled, cross-cluster overlap is small. The contrastive
subtraction recovers the model's latent harm-category geometry:
per-domain edits are projections onto a lower-rank cluster structure
the model already has, which predicts that a single composed mask
over \{hate $\cup$ crime\} rows would simultaneously install refusal
for both domains. \rev{LLaMA-3.1-8B-LRP shows the same two-cluster pattern at attenuated
magnitudes, with within-cluster overlap exceeding cross-cluster
overlap throughout.}

Table~\ref{tab:jaccard_lrp_ig} complements the cross-domain picture with
cross-method redundancy: LRP and IG select nearly disjoint row subsets
(3-6\% overlap) yet both achieve high CAST-malign refusal on the same
model. This confirms that refusal is distributed across a redundant
subspace, not concentrated in a unique set of rows.

\begin{table}[ht]
\centering
\small
\setlength{\tabcolsep}{4pt}
\caption{Top-1\% row Jaccard overlap between LRP and IG masks (same
  model, same domain). Values of 3-6\% confirm that the two methods
  identify nearly disjoint subsets within the same redundant refusal
  subspace. For reference, LRP vs.\ a seeded-Random mask yields
  $\approx 0.005$ Jaccard ($\sim$10$\times$ lower), ruling out that
  the LRP-IG overlap is coincidental.}
\label{tab:jaccard_lrp_ig}
\begin{tabular}{l rrrrr}
\toprule
Model & hate & crime & adult & medical & legal \\
\midrule
LLaMA-1B & \cellcolor{celllred} 0.051 & \cellcolor{cellred} 0.042 & \cellcolor{cellred} 0.042 & \cellcolor{celllred} 0.039 & \cellcolor{celllred} 0.035 \\ 
LLaMA-3B & \cellcolor{celllgreen} 0.058 & \cellcolor{celllgreen} 0.055 & \cellcolor{cellgreen} 0.047 & \cellcolor{cellgreen} 0.048 & \cellcolor{cellgreen} 0.049 \\ 
LLaMA-8B & \cellcolor{cellgreen} 0.059 & \cellcolor{cellgreen} 0.057 & \cellcolor{celllgreen} 0.046 & \cellcolor{celllgreen} 0.047 & \cellcolor{celllgreen} 0.041 \\ 
Qwen3-8B & \cellcolor{cellred} 0.049 & \cellcolor{celllred} 0.046 & \cellcolor{celllred} 0.045 & \cellcolor{cellred} 0.032 & \cellcolor{cellred} 0.030 \\ 
\bottomrule
\end{tabular}
\end{table}

Layer distribution of top-1\% $|C_{\text{comply}}|$ rows
(Table~\ref{tab:layer_dist}) substantiates the architecture-reversal
finding (Section~\ref{sec:cross}): on LLaMA-3.1-8B LRP is late-heavy
(44\% late) while IG is early-heavy (40\% early); on Qwen3-8B both
methods route to early layers, but LRP picks sharper rows there.

\begin{table}[ht]
\centering
\small
\setlength{\tabcolsep}{4pt}
\caption{Distribution of top-1\% $|C_{\text{comply}}|$ rows over layer
  thirds (early / mid / late), hate domain. Random would be uniform
  $0.33 / 0.33 / 0.33$.}
\label{tab:layer_dist}
\begin{tabular}{l l ccc}
\toprule
Model & Method & early & mid & late \\
\midrule
\multirow{2}{*}{LLaMA-3.1-8B} & LRP & \cellcolor{cellred} 0.08 & \cellcolor{cellgreen} 0.47 & \cellcolor{cellgreen} 0.44 \\ 
                              & IG  & \cellcolor{cellgreen} 0.40 & \cellcolor{cellred} 0.37 & \cellcolor{cellred} 0.22 \\ 
\midrule
\multirow{2}{*}{Qwen3-8B}     & LRP & \cellcolor{cellgreen} 0.59 & \cellcolor{cellred} 0.26 & \cellcolor{cellred} 0.14 \\ 
                              & IG  & \cellcolor{cellred} 0.45 & \cellcolor{cellgreen} 0.27 & \cellcolor{cellgreen} 0.27 \\ 
\bottomrule
\end{tabular}
\end{table}

% ============================================================
\textbf{Selector rescue procedure (LLaMA-3.1-8B-IG-adult).}
\label{app:rescue}

For some $(\text{model}, \text{method}, \text{domain})$ cells the
default operating-point constraint set is empty: no $(\lambda, k)$
satisfies benign $\le 0.10$ with non-trivial refusal at PPL $\le 100$.
Among the $70$ cells in our grid this fires once: LLaMA-3.1-8B IG-adult (Table~\ref{tab:rescue_sweep}).

Every cell with non-trivial malign refusal on this triple has benign
$\ge 0.18$, and increasing $\lambda$ does not break benign below
$0.20$. IG's early-layer rows for adult-refusal are also activation-positive
on adult-benign prompts, so zeroing them spills onto benign queries.

\begin{table}[ht]
\centering
\small
\setlength{\tabcolsep}{4pt}
\caption{LLaMA-3.1-8B IG-adult $(\lambda, k)$ sweep. Every cell with
  non-trivial malign violates the benign $\le 0.10$ cap. The rescue
  picks the tightest-benign cell with non-trivial refusal.}
\label{tab:rescue_sweep}\begin{tabular}{l rrr l}
\toprule
$(\lambda, k)$ & malign & benign & PPL & passes? \\
\midrule
$(0.3, 0.005)$ & \cellcolor{cellred} 0.358 & \cellcolor{celllgreen} 0.238 & \cellcolor{cellgreen} 13 & ben $\times$ \\
$(0.5, 0.005)$ & \cellcolor{celllred} 0.400 & \cellcolor{celllred} 0.250 & \cellcolor{cellgreen} 13 & ben $\times$ \\
$(0.6, 0.005)$ & \cellcolor{celllred} 0.422 & \cellcolor{celllgreen} 0.216 & \cellcolor{cellgreen} 13 & ben $\times$ \\
$(0.7, 0.005)$ & \cellcolor{celllred} 0.408 & \cellcolor{celllgreen} 0.216 & \cellcolor{cellgreen} 13 & ben $\times$ \\
$(0.8, 0.005)$ & \cellcolor{celllgreen} 0.544 & \cellcolor{cellgreen} 0.180 & \cellcolor{cellgreen} 13 & rescue \\
$(0.5, 0.02)$  & \cellcolor{celllgreen} 0.524 & \cellcolor{cellred} 0.412 & \cellcolor{celllred} 14 & ben $\times$ \\
$(0.7, 0.05)$  & \cellcolor{celllgreen} 0.606 & \cellcolor{celllred} 0.262 & \cellcolor{cellred} 17 & ben $\times$ \\
$(0.8, 0.05)$  & \cellcolor{cellgreen} 0.626 & \cellcolor{celllred} 0.374 & \cellcolor{celllred} 16 & ben $\times$ \\
\bottomrule
\end{tabular}
\end{table}

We apply a deterministic relaxation: for benign caps $c \in
\{0.05, 0.10, 0.15, 0.20, 0.25\}$, build $\mathcal{V}_c$ and pick the
smallest $c$ that contains at least one cell with malign $> 5\times$
baseline (here $> 0.29$). For 8B-IG-adult this is $c^\star = 0.20$ and
the rescue cell is $(\lambda, k) = (0.8, 0.005)$ with malign $0.544$,
benign $0.180$, PPL $13$. The cell is marked $\star$ in
Figure~\ref{fig:overrefuse_8b} and Table~\ref{tab:overrefuse_cross}, and
will be noted on its model card upon release.

% ============================================================
\textbf{$\lambda$ penalty sweep.}
\label{app:lambda_sweep}

Table~\ref{tab:lambda_sweep_8b_lrp_hate} shows representative
$(\lambda, k)$ cells for LLaMA-3.1-8B LRP-hate. Without $\lambda$
($\lambda = 0$) the top compliance rows overlap with general-LM rows
and PPL collapses; at the published value $\lambda = 0.5$ ($k = 0.02$)
the operating point reaches the highest CAST-malign ($0.814$) in this
evaluation run. Excessive $\lambda$ over-penalizes and trims
refusal-specific rows from the candidate set.

\begin{table}[ht]
\centering
\small
\setlength{\tabcolsep}{4pt}
\caption{$\lambda$ sweep on LLaMA-3.1-8B LRP-hate.}
\label{tab:lambda_sweep_8b_lrp_hate}
\begin{tabular}{l c rrr}
\toprule
$\lambda$ & $k$ & malign & benign & PPL \\
\midrule
0.3 & 0.005 & \cellcolor{celllred} 0.658 & \cellcolor{celllgreen} 0.002 & \cellcolor{cellgreen} 13 \\ 
0.3 & 0.02  & \cellcolor{celllred} 0.704 & \cellcolor{cellgreen} 0.000 & \cellcolor{cellred} 21 \\ 
0.5 & 0.02 & \cellcolor{cellgreen} 0.814 & \cellcolor{celllgreen} 0.004 & \cellcolor{cellgreen} 13 \\ 
0.7 & 0.005 & \cellcolor{cellred} 0.610 & \cellcolor{celllgreen} 0.004 & \cellcolor{cellgreen} 13 \\ 
0.7 & 0.02  & \cellcolor{celllgreen} 0.800 & \cellcolor{celllred} 0.006 & \cellcolor{cellgreen} 13 \\ 
0.8 & 0.02  & \cellcolor{celllgreen} 0.800 & \cellcolor{cellred} 0.008 & \cellcolor{cellgreen} 13 \\ 
\bottomrule
\end{tabular}
\end{table}

% ============================================================
\textbf{Other models: LLaMA-3.2-1B / 3B and Qwen3-8B.}
\label{app:contrastive_other_models}

Table~\ref{tab:overrefuse_cross} provides the numerical summary
visualized in Figure~\ref{fig:reversal}.

On LLaMA-3.2-1B, \method{IG} dominates $4$ of $5$ domains with lower $k$ requirements (adult: IG $k=0.01$, PPL$=27$ vs.\ LRP $k=0.05$, PPL$=56$); LRP's medical win is Pyrrhic ($k=0.15$, PPL$=99$).
On LLaMA-3.2-3B, \method{IG} dominates all $5$ domains; IG-adult at $k=0.01$ achieves PPL$=19$ (identical to dense) with malign$=0.774$ and no observed PPL cost; IG-crime reaches $0.980$.
On Qwen3-8B, \method{LRP} wins $3$ of $5$ domains (with hate
$0.962$, crime $0.974$, and medical narrowly at $0.666$ vs.\ $0.662$); Qwen baseline refusal and OR-Bench over-refusal are already high (hate$=0.654$, OR-Bench$=0.60$); LRP-legal causes SorryBench to drop ($0.90\to0.70$) while IG-legal is unaffected ($0.90\to1.00$).

\begin{table}[ht]
\centering
\small
\setlength{\tabcolsep}{3pt}
\caption{LRP vs.\ IG operating-point CAST malign across all four
  models (best $(\lambda, k)$ per cell, benign $\le 0.10$,
  PPL $\le 100$); numerical companion to Figure~\ref{fig:reversal}.}
\label{tab:overrefuse_cross}
\begin{tabular}{l l cc}
\toprule
Model & Domain & LRP & IG \\
\midrule
\multirow{5}{*}{LLaMA-3.2-1B}
  & hate    & \cellcolor{cellgreen} 0.752 & \cellcolor{cellgreen} 0.924 \\ 
  & crime   & \cellcolor{celllgreen} 0.694 & \cellcolor{celllgreen} 0.882 \\ 
  & adult   & \cellcolor{celllred} 0.414 & \cellcolor{celllgreen} 0.806 \\ 
  & medical & \cellcolor{celllgreen} 0.586 & \cellcolor{celllred} 0.516 \\ 
  & legal   & \cellcolor{cellred} 0.200 & \cellcolor{cellred} 0.456 \\ 
\midrule
\multirow{5}{*}{LLaMA-3.2-3B}
  & hate    & \cellcolor{celllgreen} 0.796 & \cellcolor{celllgreen} 0.936 \\ 
  & crime   & \cellcolor{cellgreen} 0.810 & \cellcolor{cellgreen} 0.980 \\ 
  & adult   & \cellcolor{celllred} 0.414 & \cellcolor{celllgreen} 0.774 \\ 
  & medical & \cellcolor{celllgreen} 0.514 & \cellcolor{celllred} 0.720 \\ 
  & legal   & \cellcolor{cellred} 0.332 & \cellcolor{cellred} 0.620 \\ 
\midrule
\multirow{5}{*}{LLaMA-3.1-8B}
  & hate    & \cellcolor{celllgreen} 0.824 & \cellcolor{cellgreen} 0.844 \\ 
  & crime   & \cellcolor{cellgreen} 0.908 & \cellcolor{celllgreen} 0.838 \\ 
  & adult   & \cellcolor{celllred} 0.486 & \cellcolor{celllred} 0.544$^\star$ \\ 
  & medical & \cellcolor{celllgreen} 0.742 & \cellcolor{celllgreen} 0.576 \\ 
  & legal   & \cellcolor{cellred} 0.480 & \cellcolor{cellred} 0.540 \\ 
\midrule
\multirow{5}{*}{Qwen3-8B}
  & hate    & \cellcolor{celllgreen} 0.962 & \cellcolor{celllgreen} 0.906 \\ 
  & crime   & \cellcolor{cellgreen} 0.974 & \cellcolor{celllgreen} 0.906 \\ 
  & adult   & \cellcolor{celllgreen} 0.878 & \cellcolor{cellgreen} 0.952 \\ 
  & medical & \cellcolor{celllred} 0.666 & \cellcolor{celllred} 0.662 \\ 
  & legal   & \cellcolor{cellred} 0.500 & \cellcolor{cellred} 0.548 \\ 
\bottomrule
\end{tabular}
\end{table}

\begin{table*}[ht]
\centering
\small
\setlength{\tabcolsep}{4pt}
\caption{Selector comparison on LLaMA-3.2-1B-Instruct: LRP vs.\ IG vs.\
  Random at best per-domain operating points.}
\label{tab:overrefuse_1b}
\begin{tabular}{l l l ccccc}
\toprule
Method & Domain & $(\lambda, k)$ & CAST malign$\uparrow$ & CAST benign$\downarrow$ & PPL$\downarrow$ & SorryBench$\uparrow$ & OR-Bench$\downarrow$ \\
\midrule
\method{LRP} & hate & $(0.3, 0.02)$ & \cellcolor{celllgreen} 0.440 $\to$ 0.752 & \cellcolor{cellgreen} 0.000 $\to$ 0.004 & \cellcolor{cellred} 25 $\to$ 36 & \cellcolor{celllgreen} 0.54 $\to$ 0.88 & \cellcolor{celllgreen} 0.26 $\to$ 0.44 \\ 
\method{IG}  & hate & $(0.3, 0.01)$ & \cellcolor{cellgreen} 0.438 $\to$ 0.924 & \cellcolor{cellred} 0.000 $\to$ 0.018 & \cellcolor{cellgreen} 25 $\to$ 29 & \cellcolor{cellgreen} 0.52 $\to$ 0.98 & \cellcolor{cellred} 0.26 $\to$ 0.68 \\ 
\method{Random} & hate & $(-, 0.005)$ & \cellcolor{cellred} 0.436 $\to$ 0.496 & \cellcolor{cellgreen} 0.000 $\to$ 0.004 & \cellcolor{cellgreen} 25 $\to$ 29 & \cellcolor{cellred} 0.52 $\to$ 0.76 & \cellcolor{cellgreen} 0.26 $\to$ 0.38 \\ 
\midrule
\method{LRP} & crime & $(0.3, 0.02)$ & \cellcolor{celllgreen} 0.312 $\to$ 0.694 & \cellcolor{cellgreen} 0.002 $\to$ 0.004 & \cellcolor{cellred} 25 $\to$ 37 & \cellcolor{cellred} 0.59 $\to$ 0.68 & \cellcolor{celllgreen} 0.26 $\to$ 0.64 \\ 
\method{IG}  & crime & $(0.5, 0.01)$ & \cellcolor{cellgreen} 0.314 $\to$ 0.882 & \cellcolor{cellred} 0.002 $\to$ 0.022 & \cellcolor{cellgreen} 25 $\to$ 27 & \cellcolor{cellgreen} 0.59 $\to$ 0.84 & \cellcolor{cellred} 0.27 $\to$ 0.80 \\ 
\method{Random} & crime & $(-, 0.005)$ & \cellcolor{cellred} 0.314 $\to$ 0.440 & \cellcolor{cellgreen} 0.002 $\to$ 0.004 & \cellcolor{celllgreen} 25 $\to$ 29 & \cellcolor{celllgreen} 0.59 $\to$ 0.71 & \cellcolor{cellgreen} 0.26 $\to$ 0.37 \\ 
\midrule
\method{LRP} & adult & $(0.5, 0.05)$ & \cellcolor{celllgreen} 0.060 $\to$ 0.414 & \cellcolor{cellred} 0.000 $\to$ 0.046 & \cellcolor{cellred} 25 $\to$ 56 & \cellcolor{cellgreen} 0.40 $\to$ 0.65 & \cellcolor{celllgreen} 0.27 $\to$ 0.61 \\ 
\method{IG}  & adult & $(0.5, 0.01)$ & \cellcolor{cellgreen} 0.060 $\to$ 0.806 & \cellcolor{cellgreen} 0.002 $\to$ 0.008 & \cellcolor{cellgreen} 25 $\to$ 27 & \cellcolor{celllgreen} 0.40 $\to$ 0.60 & \cellcolor{cellred} 0.26 $\to$ 0.77 \\ 
\method{Random} & adult & $(-, 0.01)$ & \cellcolor{cellred} 0.062 $\to$ 0.186 & \cellcolor{celllgreen} 0.002 $\to$ 0.024 & \cellcolor{celllgreen} 25 $\to$ 49 & \cellcolor{cellred} 0.40 $\to$ 0.35 & \cellcolor{cellgreen} 0.25 $\to$ 0.41 \\ 
\midrule
\method{LRP} & medical & $(0.8, 0.15)$ & \cellcolor{cellgreen} 0.090 $\to$ 0.586 & \cellcolor{cellred} 0.000 $\to$ 0.088 & \cellcolor{cellred} 25 $\to$ 99 & \cellcolor{cellred} 0.20 $\to$ 0.20 & \cellcolor{celllgreen} 0.25 $\to$ 0.55 \\ 
\method{IG}  & medical & $(0.5, 0.01)$ & \cellcolor{celllgreen} 0.090 $\to$ 0.516 & \cellcolor{cellgreen} 0.000 $\to$ 0.018 & \cellcolor{cellgreen} 25 $\to$ 27 & \cellcolor{cellgreen} 0.20 $\to$ 0.50 & \cellcolor{cellred} 0.27 $\to$ 0.69 \\ 
\method{Random} & medical & $(-, 0.01)$ & \cellcolor{cellred} 0.090 $\to$ 0.154 & \cellcolor{celllgreen} 0.002 $\to$ 0.024 & \cellcolor{celllgreen} 25 $\to$ 49 & \cellcolor{celllgreen} 0.20 $\to$ 0.30 & \cellcolor{cellgreen} 0.26 $\to$ 0.41 \\ 
\midrule
\method{LRP} & legal & $(0.8, 0.015)$ & \cellcolor{celllgreen} 0.022 $\to$ 0.200 & \cellcolor{celllgreen} 0.002 $\to$ 0.088 & \cellcolor{cellgreen} 25 $\to$ 27 & \cellcolor{cellgreen} 0.30 $\to$ 0.90 & \cellcolor{celllgreen} 0.24 $\to$ 0.58 \\ 
\method{IG}  & legal & $(0.3, 0.005)$ & \cellcolor{cellgreen} 0.024 $\to$ 0.456 & \cellcolor{celllgreen} 0.002 $\to$ 0.088 & \cellcolor{celllgreen} 25 $\to$ 32 & \cellcolor{cellgreen} 0.30 $\to$ 0.90 & \cellcolor{cellred} 0.27 $\to$ 0.78 \\ 
\method{Random} & legal & $(-, 0.01)$ & \cellcolor{cellred} 0.024 $\to$ 0.074 & \cellcolor{cellgreen} 0.002 $\to$ 0.022 & \cellcolor{cellred} 25 $\to$ 49 & \cellcolor{cellred} 0.30 $\to$ 0.60 & \cellcolor{cellgreen} 0.27 $\to$ 0.40 \\ 
\bottomrule
\end{tabular}
\end{table*}

\begin{table*}[ht]
\centering
\small
\setlength{\tabcolsep}{4pt}
\caption{Contrastive refusal editing: LLaMA-3.2-3B-Instruct.}
\label{tab:overrefuse_3b}
\begin{tabular}{l l l ccccc}
\toprule
Method & Domain & $(\lambda, k)$ & CAST malign$\uparrow$ & CAST benign$\downarrow$ & PPL$\downarrow$ & SorryBench$\uparrow$ & OR-Bench$\downarrow$ \\
\midrule
\method{LRP} & hate    & $(0.5, 0.05)$ & \cellcolor{celllgreen} 0.416 $\to$ 0.796 & \cellcolor{cellgreen} 0.000 $\to$ 0.004 & \cellcolor{celllgreen} 19 $\to$ 29 & \cellcolor{celllgreen} 0.70 $\to$ 0.90 & \cellcolor{celllgreen} 0.32 $\to$ 0.80 \\ 
\method{LRP} & crime   & $(0.7, 0.05)$ & \cellcolor{cellgreen} 0.294 $\to$ 0.810 & \cellcolor{celllgreen} 0.000 $\to$ 0.010 & \cellcolor{cellgreen} 19 $\to$ 23 & \cellcolor{celllred} 0.72 $\to$ 0.72 & \cellcolor{celllgreen} 0.32 $\to$ 0.80 \\ 
\method{LRP} & adult   & $(0.8, 0.05)$ & \cellcolor{celllred} 0.034 $\to$ 0.414 & \cellcolor{celllgreen} 0.000 $\to$ 0.006 & \cellcolor{cellgreen} 19 $\to$ 23 & \cellcolor{celllgreen} 0.30 $\to$ 0.75 & \cellcolor{celllred} 0.32 $\to$ 0.84 \\ 
\method{LRP} & medical & $(0.3, 0.05)$ & \cellcolor{celllgreen} 0.108 $\to$ 0.514 & \cellcolor{celllred} 0.000 $\to$ 0.016 & \cellcolor{cellred} 19 $\to$ 69 & \cellcolor{cellred} 0.20 $\to$ 0.30 & \cellcolor{cellgreen} 0.32 $\to$ 0.40 \\ 
\method{LRP} & legal$^\dagger$   & $(0.3, 0.01)$ & \cellcolor{cellred} 0.038 $\to$ 0.332 & \cellcolor{cellred} 0.000 $\to$ 0.032 & \cellcolor{celllred} 19 $\to$ 32 & \cellcolor{cellgreen} 1.00 $\to$ 1.00 & \cellcolor{cellred} 0.32 $\to$ 0.88 \\ 
\midrule
\method{IG}  & hate    & $(0.5, 0.05)$ & \cellcolor{celllgreen} 0.416 $\to$ 0.936 & \cellcolor{cellgreen} 0.000 $\to$ 0.020 & \cellcolor{cellred} 19 $\to$ 33 & \cellcolor{celllgreen} 0.70 $\to$ 0.96 & \cellcolor{celllgreen} 0.32 $\to$ 0.82 \\ 
\method{IG}  & crime   & $(0.8, 0.05)$ & \cellcolor{cellgreen} 0.294 $\to$ 0.980 & \cellcolor{cellred} 0.000 $\to$ 0.092 & \cellcolor{celllgreen} 19 $\to$ 24 & \cellcolor{celllred} 0.72 $\to$ 0.87 & \cellcolor{cellred} 0.32 $\to$ 0.94 \\ 
\method{IG}  & adult   & $(0.5, 0.01)$ & \cellcolor{celllgreen} 0.034 $\to$ 0.774 & \cellcolor{celllgreen} 0.000 $\to$ 0.060 & \cellcolor{cellgreen} 19 $\to$ 19 & \cellcolor{celllgreen} 0.30 $\to$ 0.95 & \cellcolor{celllred} 0.32 $\to$ 0.92 \\ 
\method{IG}  & medical & $(0.3, 0.01)$ & \cellcolor{celllred} 0.108 $\to$ 0.720 & \cellcolor{celllgreen} 0.000 $\to$ 0.060 & \cellcolor{celllred} 19 $\to$ 27 & \cellcolor{cellred} 0.20 $\to$ 0.80 & \cellcolor{cellgreen} 0.32 $\to$ 0.81 \\ 
\method{IG}  & legal$^\dagger$   & $(0.7, 0.02)$ & \cellcolor{cellred} 0.038 $\to$ 0.620 & \cellcolor{celllred} 0.000 $\to$ 0.070 & \cellcolor{celllgreen} 19 $\to$ 20 & \cellcolor{cellgreen} 1.00 $\to$ 1.00 & \cellcolor{celllgreen} 0.32 $\to$ 0.83 \\ 
\midrule
\method{Random} & hate    & $(-, 0.005)$ & \cellcolor{cellgreen} 0.416 $\to$ 0.376 & \cellcolor{cellgreen} 0.000 $\to$ 0.000 & \cellcolor{cellgreen} 19 $\to$ 23 & \cellcolor{celllgreen} 0.70 $\to$ 0.58 & \cellcolor{celllgreen} 0.32 $\to$ 0.27 \\ 
\method{Random} & crime   & $(-, 0.005)$ & \cellcolor{celllgreen} 0.294 $\to$ 0.362 & \cellcolor{cellgreen} 0.000 $\to$ 0.000 & \cellcolor{cellgreen} 19 $\to$ 23 & \cellcolor{celllgreen} 0.72 $\to$ 0.72 & \cellcolor{cellgreen} 0.32 $\to$ 0.26 \\ 
\method{Random} & adult   & $(-, 0.020)$ & \cellcolor{celllred} 0.034 $\to$ 0.030 & \cellcolor{celllred} 0.000 $\to$ 0.006 & \cellcolor{celllred} 19 $\to$ 40 & \cellcolor{celllred} 0.30 $\to$ 0.45 & \cellcolor{celllgreen} 0.32 $\to$ 0.27 \\ 
\method{Random} & medical & $(-, 0.005)$ & \cellcolor{celllgreen} 0.108 $\to$ 0.068 & \cellcolor{cellgreen} 0.000 $\to$ 0.000 & \cellcolor{cellgreen} 19 $\to$ 23 & \cellcolor{cellred} 0.20 $\to$ 0.20 & \cellcolor{celllgreen} 0.32 $\to$ 0.27 \\ 
\method{Random} & legal$^\dagger$   & $(-, 0.020)$ & \cellcolor{cellred} 0.038 $\to$ 0.022 & \cellcolor{celllred} 0.000 $\to$ 0.006 & \cellcolor{celllred} 19 $\to$ 40 & \cellcolor{cellgreen} 0.90 $\to$ 0.90 & \cellcolor{celllgreen} 0.32 $\to$ 0.27 \\ 
\midrule
\multicolumn{8}{l}{\footnotesize $\dagger$ SorryBench ceiling: unedited 3B already refuses nearly all SorryBench legal prompts; CAST-malign} \\
\multicolumn{8}{l}{\footnotesize \phantom{$\dagger$} improvement (0.038$\to$0.332--0.620) reflects harder adversarial prompts not captured by SorryBench.} \\
\bottomrule
\end{tabular}
\end{table*}

\clearpage

\begin{table*}[ht]
\centering
\small
\setlength{\tabcolsep}{4pt}
\caption{Contrastive refusal editing: Qwen3-8B.}
\label{tab:overrefuse_qwen}
\begin{tabular}{l l l ccccc}
\toprule
Method & Domain & $(\lambda, k)$ & CAST malign$\uparrow$ & CAST benign$\downarrow$ & PPL$\downarrow$ & SorryBench$\uparrow$ & OR-Bench$\downarrow$ \\
\midrule
\method{LRP} & hate    & $(0.8, 0.05)$  & \cellcolor{celllgreen} 0.654 $\to$ 0.962 & \cellcolor{cellred} 0.000 $\to$ 0.084 & \cellcolor{celllgreen} 19 $\to$ 22 & \cellcolor{cellgreen} 0.86 $\to$ 1.00 & \cellcolor{celllgreen} 0.60 $\to$ 0.99 \\ 
\method{LRP} & crime   & $(0.7, 0.02)$  & \cellcolor{cellgreen} 0.362 $\to$ 0.974 & \cellcolor{celllred} 0.000 $\to$ 0.078 & \cellcolor{cellgreen} 19 $\to$ 20 & \cellcolor{celllgreen} 0.83 $\to$ 0.96 & \cellcolor{celllred} 0.60 $\to$ 1.00 \\ 
\method{LRP} & adult   & $(0.5, 0.01)$  & \cellcolor{celllgreen} 0.086 $\to$ 0.878 & \cellcolor{cellgreen} 0.000 $\to$ 0.026 & \cellcolor{cellgreen} 19 $\to$ 20 & \cellcolor{cellgreen} 0.60 $\to$ 1.00 & \cellcolor{celllred} 0.60 $\to$ 1.00 \\ 
\method{LRP} & medical & $(0.3, 0.02)$  & \cellcolor{celllred} 0.044 $\to$ 0.666 & \cellcolor{celllgreen} 0.000 $\to$ 0.056 & \cellcolor{cellred} 19 $\to$ 28 & \cellcolor{celllred} 0.40 $\to$ 0.80 & \cellcolor{celllgreen} 0.60 $\to$ 0.96 \\ 
\method{LRP} & legal$^\ddagger$   & $(0.7, 0.075)$ & \cellcolor{cellred} 0.012 $\to$ 0.500 & \cellcolor{celllgreen} 0.000 $\to$ 0.044 & \cellcolor{celllred} 19 $\to$ 24 & \cellcolor{cellred} 0.90 $\to$ 0.70 & \cellcolor{cellgreen} 0.60 $\to$ 0.61 \\ 
\midrule
\method{IG}  & hate    & $(0.7, 0.02)$  & \cellcolor{celllgreen} 0.654 $\to$ 0.906 & \cellcolor{cellgreen} 0.000 $\to$ 0.014 & \cellcolor{celllgreen} 19 $\to$ 21 & \cellcolor{celllgreen} 0.86 $\to$ 0.98 & \cellcolor{celllred} 0.60 $\to$ 0.94 \\ 
\method{IG}  & crime   & $(0.7, 0.01)$  & \cellcolor{celllgreen} 0.362 $\to$ 0.906 & \cellcolor{celllred} 0.000 $\to$ 0.086 & \cellcolor{cellgreen} 19 $\to$ 20 & \cellcolor{celllred} 0.83 $\to$ 0.87 & \cellcolor{celllgreen} 0.60 $\to$ 0.87 \\ 
\method{IG}  & adult   & $(0.8, 0.10)$  & \cellcolor{cellgreen} 0.086 $\to$ 0.952 & \cellcolor{cellred} 0.000 $\to$ 0.092 & \cellcolor{cellred} 19 $\to$ 35 & \cellcolor{cellgreen} 0.60 $\to$ 1.00 & \cellcolor{cellgreen} 0.60 $\to$ 0.47 \\ 
\method{IG}  & medical & $(0.7, 0.01)$  & \cellcolor{celllred} 0.044 $\to$ 0.662 & \cellcolor{celllgreen} 0.000 $\to$ 0.054 & \cellcolor{cellgreen} 19 $\to$ 20 & \cellcolor{cellred} 0.40 $\to$ 0.40 & \cellcolor{celllgreen} 0.60 $\to$ 0.93 \\ 
\method{IG}  & legal   & $(0.8, 0.02)$  & \cellcolor{cellred} 0.012 $\to$ 0.548 & \cellcolor{celllgreen} 0.000 $\to$ 0.046 & \cellcolor{celllgreen} 19 $\to$ 21 & \cellcolor{cellgreen} 0.90 $\to$ 1.00 & \cellcolor{cellred} 0.60 $\to$ 0.95 \\ 
\midrule
\method{Random} & hate    & $(-, 0.005)$ & \cellcolor{cellgreen} 0.654 $\to$ 0.622 & \cellcolor{cellgreen} 0.000 $\to$ 0.002 & \cellcolor{cellgreen} 19 $\to$ 20 & \cellcolor{celllgreen} 0.86 $\to$ 0.78 & \cellcolor{celllgreen} 0.60 $\to$ 0.54 \\ 
\method{Random} & crime   & $(-, 0.020)$ & \cellcolor{celllgreen} 0.362 $\to$ 0.378 & \cellcolor{cellgreen} 0.000 $\to$ 0.002 & \cellcolor{celllred} 19 $\to$ 24 & \cellcolor{cellgreen} 0.83 $\to$ 0.80 & \cellcolor{celllgreen} 0.60 $\to$ 0.53 \\ 
\method{Random} & adult   & $(-, 0.005)$ & \cellcolor{celllgreen} 0.086 $\to$ 0.064 & \cellcolor{cellgreen} 0.000 $\to$ 0.002 & \cellcolor{cellgreen} 19 $\to$ 20 & \cellcolor{celllred} 0.60 $\to$ 0.55 & \cellcolor{celllgreen} 0.60 $\to$ 0.54 \\ 
\method{Random} & medical & $(-, 0.005)$ & \cellcolor{celllred} 0.044 $\to$ 0.038 & \cellcolor{cellgreen} 0.000 $\to$ 0.002 & \cellcolor{cellgreen} 19 $\to$ 20 & \cellcolor{cellred} 0.40 $\to$ 0.30 & \cellcolor{celllgreen} 0.60 $\to$ 0.54 \\ 
\method{Random} & legal$^\ddagger$   & $(-, 0.050)$ & \cellcolor{cellred} 0.012 $\to$ 0.008 & \cellcolor{cellgreen} 0.000 $\to$ 0.002 & \cellcolor{cellred} 19 $\to$ 34 & \cellcolor{celllgreen} 0.90 $\to$ 0.70 & \cellcolor{cellgreen} 0.60 $\to$ 0.41 \\ 
\midrule
\multicolumn{8}{l}{\footnotesize $\ddagger$ Qwen3 natively refuses 90\% of SorryBench legal prompts; LRP/Random edits targeting} \\
\multicolumn{8}{l}{\footnotesize \phantom{$\ddagger$} adversarial CAST legal prompts (0.012$\to$0.500 / 0.008) slightly reduce natural refusal on} \\
\multicolumn{8}{l}{\footnotesize \phantom{$\ddagger$} softer SorryBench queries. IG-legal is unaffected (0.90$\to$1.00).} \\
\bottomrule
\end{tabular}
\end{table*}

% ============================================================
\textbf{Calibration signal density.}
\label{app:signal_density}

Table~\ref{tab:signal_density} reports the fraction of neuron-rows with
$|C_{\text{comply}}| > 0.5$ after per-layer rank normalization for each
domain and selector; the ordering and its implications for operating-point $k$ are discussed in §\ref{sec:headline}.

\begin{table*}[ht]
\centering
\small
\setlength{\tabcolsep}{4pt}
\caption{Calibration signal density across all four models (fraction of
  neuron-rows with $|C_{\text{comply}}| > 0.5$ after per-layer
  rank normalization; values in \%). MA = MeanActivation; C2 =
  \consensus{}; both run on LLaMA-3.1-8B only. Signal density grows
  with LLaMA model size (hate LRP: 1B 4.2\% $\to$ 3B 5.5\% $\to$ 8B
  5.8\%); Qwen3-8B is 3$\times$ denser than LLaMA-8B on hate LRP
  (17.4\% vs.\ 5.8\%), tracking its early-layer intent-classification
  architecture. The hate $>$ crime $>$ adult $>$ medical $>$ legal
  ordering holds across most model-method combinations and predicts the operating-point $k$:
  hate/crime converge at $k=0.01$-$0.02$; legal requires
  $k=0.05$-$0.15$.}
\label{tab:signal_density}
\begin{tabular}{l rr rr rrrr rr}
\toprule
& \multicolumn{2}{c}{LLaMA-1B} & \multicolumn{2}{c}{LLaMA-3B}
& \multicolumn{4}{c}{LLaMA-8B} & \multicolumn{2}{c}{Qwen3-8B} \\
\cmidrule(lr){2-3}\cmidrule(lr){4-5}\cmidrule(lr){6-9}\cmidrule(lr){10-11}
Domain & LRP & IG & LRP & IG & LRP & IG & MA & C2 & LRP & IG \\
\midrule
hate    & \cellcolor{cellgreen} 4.16 &  \cellcolor{celllgreen} 4.48 & \cellcolor{cellgreen} 5.50 &  \cellcolor{celllgreen} 7.37 & \cellcolor{cellgreen} 5.77 &  \cellcolor{cellgreen} 8.52 & \cellcolor{cellgreen} 0.01 & \cellcolor{cellgreen} 5.71 & \cellcolor{cellgreen} 17.39 & \cellcolor{cellgreen} 13.78 \\ 
crime   & \cellcolor{celllgreen} 4.12 &  \cellcolor{cellgreen} 5.53 & \cellcolor{celllgreen} 5.18 &  \cellcolor{cellgreen} 7.70 & \cellcolor{celllgreen} 4.82 &  \cellcolor{celllgreen} 7.56 & \cellcolor{cellgreen} 0.01 & \cellcolor{celllgreen} 4.86 & \cellcolor{celllgreen} 12.42 &  \cellcolor{celllgreen} 9.35 \\ 
adult   & \cellcolor{celllred} 2.24 &  \cellcolor{celllgreen} 3.09 & \cellcolor{celllgreen} 3.18 &  \cellcolor{celllgreen} 5.07 & \cellcolor{celllgreen} 2.78 &  \cellcolor{celllgreen} 4.54 & \cellcolor{cellgreen} 0.01 & \cellcolor{celllgreen} 2.96 &  \cellcolor{celllgreen} 5.73 &  \cellcolor{celllgreen} 4.75 \\ 
medical & \cellcolor{celllgreen} 2.31 &  \cellcolor{celllred} 2.77 & \cellcolor{celllred} 2.65 &  \cellcolor{celllred} 3.74 & \cellcolor{celllred} 2.44 &  \cellcolor{celllred} 3.59 & \cellcolor{cellgreen} 0.01 & \cellcolor{celllred} 2.68 &  \cellcolor{celllred} 2.76 &  \cellcolor{celllred} 3.31 \\ 
legal   & \cellcolor{cellred} 1.71 &  \cellcolor{cellred} 1.89 & \cellcolor{cellred} 2.24 &  \cellcolor{cellred} 3.14 & \cellcolor{cellred} 1.93 &  \cellcolor{cellred} 3.07 & \cellcolor{cellred} 0.00 & \cellcolor{cellred} 2.13 &  \cellcolor{cellred} 2.39 &  \cellcolor{cellred} 2.51 \\ 
\bottomrule
\end{tabular}
\end{table*}

% ============================================================
\textbf{OOD safety and utility (full numerics).}
\label{app:utility}

Table~\ref{tab:utility_full} reports the held-out OOD refusal rates
(SorryBench, OR-Bench-Hard) and utility deltas (MMLU, GSM8K, IFEval)
for all 25 LLaMA-3.1-8B cells (5 methods $\times$ 5 domains) at their
operating points. The first three columns repeat the main-body
Figure~\ref{fig:overrefuse_8b} numerics for cross-reference; the held-out OOD
columns are visualized in Figure~\ref{fig:ood_utility}. Baselines:
MMLU $0.680$, GSM8K $0.510$, IFEval $0.818$; baseline SorryBench
$\approx 0.30$, OR-Bench-Hard $\approx 0.30$.
Hate/crime/adult cells with $k \le 0.02$ preserve most utility;
medical/legal cells at $k = 0.10$ collapse on reasoning tasks. The
collapse is sparsity-driven and fires for LRP, \consensus{}, and Random
alike.

\begin{table*}[ht]
\centering
\footnotesize
\setlength{\tabcolsep}{3.5pt}
\caption{Wilson $95\%$ intervals for the LLaMA-3.1-8B refusal-editing
rates of Table~\ref{tab:utility_full}, from the printed proportions and
prompt counts (CAST $n{=}500$ per split, OR-Bench-Hard $n{=}1000$); no
artifacts required. CAST intervals are exact; OR-Bench rates are
two-decimal, so those intervals are accurate to about $\pm0.01$. Every
headline attribution-vs-control separation is disjoint at this level.}
\label{tab:wilson}
\begin{tabular}{lllll}
\toprule
Method & Domain & mal$\uparrow$ [95\% CI] & ben$\downarrow$ [95\% CI] & OR$\downarrow$ [95\% CI] \\
\midrule
\method{LRP} & hate & .824 [.788, .855] & .004 [.001, .014] & .780 [.753, .805] \\
\method{LRP} & crime & .908 [.879, .930] & .008 [.003, .020] & .880 [.858, .899] \\
\method{LRP} & adult & .486 [.442, .530] & .008 [.003, .020] & .680 [.650, .708] \\
\method{LRP} & medical & .742 [.702, .778] & .028 [.017, .046] & .620 [.590, .650] \\
\method{LRP} & legal & .480 [.437, .524] & .042 [.028, .063] & .680 [.650, .708] \\
\addlinespace
\method{IG} & hate & .844 [.810, .873] & .016 [.008, .031] & .810 [.785, .833] \\
\method{IG} & crime & .838 [.803, .868] & .080 [.059, .107] & .660 [.630, .689] \\
\method{IG} & adult$^{\star}$ & .544 [.500, .587] & .180 [.149, .216] & .570 [.539, .600] \\
\method{IG} & medical & .576 [.532, .619] & .090 [.068, .118] & .640 [.610, .669] \\
\method{IG} & legal & .540 [.496, .583] & .096 [.073, .125] & .640 [.610, .669] \\
\addlinespace
\method{MeanAct.} & hate & .124 [.098, .156] & .004 [.001, .014] & .190 [.167, .215] \\
\method{MeanAct.} & crime & .080 [.059, .107] & .004 [.001, .014] & .160 [.139, .184] \\
\method{MeanAct.} & adult & .008 [.003, .020] & .002 [.000, .011] & .280 [.253, .309] \\
\method{MeanAct.} & medical & .026 [.015, .044] & .004 [.001, .014] & .230 [.205, .257] \\
\method{MeanAct.} & legal & .036 [.023, .056] & .004 [.001, .014] & .340 [.311, .370] \\
\addlinespace
\consensus{} & hate & .846 [.812, .875] & .002 [.000, .011] & .830 [.805, .852] \\
\consensus{} & crime & .948 [.925, .964] & .008 [.003, .020] & .870 [.848, .889] \\
\consensus{} & adult & .534 [.490, .577] & .014 [.007, .029] & .770 [.743, .795] \\
\consensus{} & medical & .798 [.761, .831] & .052 [.036, .075] & .700 [.671, .728] \\
\consensus{} & legal & .382 [.340, .425] & .014 [.007, .029] & .760 [.733, .785] \\
\addlinespace
\method{Random} & hate & .384 [.342, .427] & .002 [.000, .011] & .460 [.429, .491] \\
\method{Random} & crime & .298 [.260, .340] & .000 [.000, .008] & .460 [.429, .491] \\
\method{Random} & adult & .054 [.037, .077] & .002 [.000, .011] & .460 [.429, .491] \\
\method{Random} & medical & .120 [.094, .151] & .008 [.003, .020] & .350 [.321, .380] \\
\method{Random} & legal & .066 [.047, .091] & .008 [.003, .020] & .350 [.321, .380] \\
\bottomrule
\end{tabular}
\end{table*}

\begin{table*}[ht]
\centering
\scriptsize
\setlength{\tabcolsep}{3pt}
\caption{Full per-cell numerics for contrastive refusal editing on
  LLaMA-3.1-8B-Instruct. mal/ben = CAST malign / benign refusal;
  Sorry / OR = held-out post-edit rates on SorryBench / OR-Bench-Hard;
  $\Delta$MMLU/GSM8K/IFEval = post-minus-base difference (base MMLU $0.680$,
  GSM8K $0.510$, IFEval $0.818$; negative = degradation). $^\star$ =
  rescue cell with benign cap relaxed to $0.20$ (achieved benign $0.18$). Medical/legal collapse
  at $k=0.10$ is sparsity-driven, not selector-driven.}
\label{tab:utility_full}
\begin{tabular}{l l c rr r rr rrr}
\toprule
Method & Domain & $(\lambda, k)$ & mal$\uparrow$ & ben$\downarrow$ & PPL$\downarrow$ & Sorry$\uparrow$ & OR$\downarrow$ & $\Delta$MMLU & $\Delta$GSM & $\Delta$IF \\
\midrule
\method{LRP} & hate    & $(0.5, 0.02)$  & \cellcolor{celllgreen} 0.824 & \cellcolor{cellgreen} 0.004 & \cellcolor{cellgreen} 13 & \cellcolor{cellgreen} 0.94 & \cellcolor{celllred} 0.78 & \cellcolor{celllgreen} $-$0.075 & \cellcolor{cellgreen} $-$0.218 & \cellcolor{cellgreen} $-$0.030 \\ 
\method{LRP} & crime   & $(0.5, 0.02)$  & \cellcolor{cellgreen} 0.908 & \cellcolor{celllgreen} 0.008 & \cellcolor{cellgreen} 13 & \cellcolor{celllgreen} 0.86 & \cellcolor{cellred} 0.88 & \cellcolor{celllgreen} $-$0.081 & \cellcolor{celllgreen} $-$0.322 & \cellcolor{celllgreen} $-$0.067 \\ 
\method{LRP} & adult   & $(0.7, 0.01)$  & \cellcolor{celllred} 0.486 & \cellcolor{celllgreen} 0.008 & \cellcolor{cellgreen} 13 & \cellcolor{celllred} 0.70 & \cellcolor{celllgreen} 0.68 & \cellcolor{cellgreen} $-$0.057 & \cellcolor{celllgreen} $-$0.283 & \cellcolor{celllgreen} $-$0.033 \\ 
\method{LRP} & medical & $(0.5, 0.10)$  & \cellcolor{celllgreen} 0.742 & \cellcolor{celllred} 0.028 & \cellcolor{celllred} 40 & \cellcolor{cellred} 0.30 & \cellcolor{cellgreen} 0.62 & \cellcolor{cellred} $-$0.451 & \cellcolor{celllred} $-$0.510 & \cellcolor{celllred} $-$0.565 \\ 
\method{LRP} & legal   & $(0.5, 0.10)$  & \cellcolor{cellred} 0.480 & \cellcolor{cellred} 0.042 & \cellcolor{cellred} 50 & \cellcolor{celllgreen} 0.90 & \cellcolor{celllgreen} 0.68 & \cellcolor{celllred} $-$0.449 & \cellcolor{celllred} $-$0.510 & \cellcolor{cellred} $-$0.614 \\ 
\midrule
\method{IG}  & hate    & $(0.8, 0.005)$ & \cellcolor{cellgreen} 0.844 & \cellcolor{cellgreen} 0.016 & \cellcolor{cellgreen} 13 & \cellcolor{celllgreen} 0.80 & \cellcolor{cellred} 0.81 & \cellcolor{cellgreen} $-$0.056 & \cellcolor{cellgreen} $-$0.274 & \cellcolor{cellgreen} $-$0.024 \\ 
\method{IG}  & crime   & $(0.8, 0.005)$ & \cellcolor{celllgreen} 0.838 & \cellcolor{celllgreen} 0.080 & \cellcolor{cellgreen} 13 & \cellcolor{celllgreen} 0.76 & \cellcolor{celllred} 0.66 & \cellcolor{celllgreen} $-$0.086 & \cellcolor{celllgreen} $-$0.339 & \cellcolor{celllgreen} $-$0.174 \\ 
\method{IG}  & adult   & $(0.8, 0.005)^\star$ & \cellcolor{celllred} 0.544 & \cellcolor{cellred} 0.180 & \cellcolor{cellgreen} 13 & \cellcolor{celllred} 0.75 & \cellcolor{cellgreen} 0.57 & \cellcolor{celllred} $-$0.148 & \cellcolor{cellred} $-$0.451 & \cellcolor{celllred} $-$0.287 \\ 
\method{IG}  & medical & $(0.5, 0.01)$  & \cellcolor{celllgreen} 0.576 & \cellcolor{celllgreen} 0.090 & \cellcolor{cellgreen} 13 & \cellcolor{cellred} 0.70 & \cellcolor{celllgreen} 0.64 & \cellcolor{celllgreen} $-$0.089 & \cellcolor{celllgreen} $-$0.331 & \cellcolor{celllgreen} $-$0.154 \\ 
\method{IG}  & legal   & $(0.8, 0.01)$  & \cellcolor{cellred} 0.540 & \cellcolor{celllred} 0.096 & \cellcolor{cellgreen} 13 & \cellcolor{cellgreen} 1.00 & \cellcolor{celllgreen} 0.64 & \cellcolor{cellred} $-$0.186 & \cellcolor{celllred} $-$0.449 & \cellcolor{cellred} $-$0.346 \\ 
\midrule
\method{MeanAct.} & hate    & $(0.5, 0.005)$ & \cellcolor{cellgreen} 0.124 & \cellcolor{celllgreen} 0.004 & \cellcolor{cellgreen} 13 & \cellcolor{celllred} 0.12 & \cellcolor{celllgreen} 0.19 & \cellcolor{celllred} $-$0.064 & \cellcolor{cellred} $-$0.375 & \cellcolor{cellgreen} $+$0.006 \\ 
\method{MeanAct.} & crime   & $(0.7, 0.005)$ & \cellcolor{celllgreen} 0.080 & \cellcolor{celllgreen} 0.004 & \cellcolor{cellgreen} 13 & \cellcolor{celllgreen} 0.42 & \cellcolor{cellgreen} 0.16 & \cellcolor{cellred} $-$0.070 & \cellcolor{cellgreen} $-$0.207 & \cellcolor{celllgreen} $\pm$0.000 \\ 
\method{MeanAct.} & adult   & $(0.3, 0.005)$ & \cellcolor{cellred} 0.008 & \cellcolor{cellgreen} 0.002 & \cellcolor{cellgreen} 13 & \cellcolor{cellred} 0.05 & \cellcolor{celllred} 0.28 & \cellcolor{cellgreen} $-$0.053 & \cellcolor{celllgreen} $-$0.361 & \cellcolor{cellred} $-$0.018 \\ 
\method{MeanAct.} & medical & $(0.8, 0.005)$ & \cellcolor{celllred} 0.026 & \cellcolor{celllgreen} 0.004 & \cellcolor{cellgreen} 13 & \cellcolor{celllgreen} 0.30 & \cellcolor{celllgreen} 0.23 & \cellcolor{celllgreen} $-$0.060 & \cellcolor{celllgreen} $-$0.275 & \cellcolor{celllgreen} $-$0.012 \\ 
\method{MeanAct.} & legal   & $(0.5, 0.01)$  & \cellcolor{celllgreen} 0.036 & \cellcolor{celllgreen} 0.004 & \cellcolor{cellgreen} 13 & \cellcolor{cellgreen} 0.80 & \cellcolor{cellred} 0.34 & \cellcolor{celllgreen} $-$0.056 & \cellcolor{celllgreen} $-$0.361 & \cellcolor{celllred} $-$0.017 \\ 
\midrule
\consensus{} & hate    & $(0.3, 0.01)$  & \cellcolor{celllgreen} 0.846 & \cellcolor{cellgreen} 0.002 & \cellcolor{celllgreen} 15 & \cellcolor{cellgreen} 0.98 & \cellcolor{celllred} 0.83 & \cellcolor{celllgreen} $-$0.071 & \cellcolor{cellgreen} $-$0.255 & \cellcolor{cellgreen} $-$0.025 \\ 
\consensus{} & crime   & $(0.8, 0.02)$  & \cellcolor{cellgreen} 0.948 & \cellcolor{celllgreen} 0.008 & \cellcolor{celllgreen} 14 & \cellcolor{celllgreen} 0.82 & \cellcolor{cellred} 0.87 & \cellcolor{celllgreen} $-$0.073 & \cellcolor{celllgreen} $-$0.309 & \cellcolor{celllgreen} $-$0.040 \\ 
\consensus{} & adult   & $(0.5, 0.01)$  & \cellcolor{celllred} 0.534 & \cellcolor{celllgreen} 0.014 & \cellcolor{cellgreen} 13 & \cellcolor{celllred} 0.75 & \cellcolor{celllgreen} 0.77 & \cellcolor{cellgreen} $-$0.062 & \cellcolor{celllgreen} $-$0.276 & \cellcolor{celllgreen} $-$0.054 \\ 
\consensus{} & medical & $(0.3, 0.10)$  & \cellcolor{celllgreen} 0.798 & \cellcolor{cellred} 0.052 & \cellcolor{cellred} 52 & \cellcolor{cellred} 0.70 & \cellcolor{cellgreen} 0.70 & \cellcolor{cellred} $-$0.450 & \cellcolor{cellred} $-$0.510 & \cellcolor{cellred} $-$0.599 \\ 
\consensus{} & legal   & $(0.7, 0.075)$ & \cellcolor{cellred} 0.382 & \cellcolor{celllgreen} 0.014 & \cellcolor{celllred} 21 & \cellcolor{celllgreen} 0.90 & \cellcolor{celllgreen} 0.76 & \cellcolor{celllred} $-$0.327 & \cellcolor{celllred} $-$0.481 & \cellcolor{celllred} $-$0.233 \\ 
\midrule
\method{Random} & hate    & $(-, 0.010)$ & \cellcolor{cellgreen} 0.384 & \cellcolor{celllgreen} 0.002 & \cellcolor{cellgreen} 16 & \cellcolor{celllgreen} 0.78 & \cellcolor{celllgreen} 0.46 & \cellcolor{cellgreen} $-$0.127 & \cellcolor{cellgreen} $-$0.433 & \cellcolor{celllgreen} $-$0.073 \\ 
\method{Random} & crime   & $(-, 0.005)$ & \cellcolor{celllgreen} 0.298 & \cellcolor{cellgreen} 0.000 & \cellcolor{cellgreen} 16 & \cellcolor{cellgreen} 0.84 & \cellcolor{celllgreen} 0.46 & \cellcolor{celllgreen} $-$0.147 & \cellcolor{celllgreen} $-$0.500 & \cellcolor{cellgreen} $-$0.028 \\ 
\method{Random} & adult   & $(-, 0.010)$ & \cellcolor{cellred} 0.054 & \cellcolor{celllgreen} 0.002 & \cellcolor{cellgreen} 16 & \cellcolor{celllgreen} 0.60 & \cellcolor{celllgreen} 0.46 & \cellcolor{celllgreen} $-$0.147 & \cellcolor{celllgreen} $-$0.500 & \cellcolor{celllgreen} $-$0.073 \\ 
\method{Random} & medical & $(-, 0.020)$ & \cellcolor{celllgreen} 0.120 & \cellcolor{celllred} 0.008 & \cellcolor{celllred} 32 & \cellcolor{cellred} 0.50 & \cellcolor{cellgreen} 0.35 & \cellcolor{celllred} $-$0.426 & \cellcolor{celllred} $-$0.510 & \cellcolor{cellred} $-$0.324 \\ 
\method{Random} & legal   & $(-, 0.020)$ & \cellcolor{celllred} 0.066 & \cellcolor{celllred} 0.008 & \cellcolor{celllred} 32 & \cellcolor{celllgreen} 0.60 & \cellcolor{cellgreen} 0.35 & \cellcolor{celllred} $-$0.426 & \cellcolor{celllred} $-$0.510 & \cellcolor{celllred} $-$0.323 \\ 
\bottomrule
\end{tabular}
\end{table*}

% ============================================================
\clearpage
\section{Extended Discussion}
\label{app:extended_discussion}

\textbf{Implications for attribution-guided pruning.}
The structural analyses qualify the attribution advantage in three ways.
First, rank stability is a misleading proxy: Wanda is the most
stable selector at Spearman $0.9998$ and is not the most faithful,
and MeanActivation's near-perfect stability hides catastrophic
causal invalidity. Second, the Wanda inversion at LLaMA-3.2-3B and
Qwen3-8B (negative validity gap) shows that activation-weighted
criteria can invert importance ordering on specific architectures.
Third, false-negative rates confirm that \method{IG} and
\consensus{} both reach zero dangerous false negatives while
\method{LRP} alone misses one dangerous group, \method{Wanda}
misses four ($41.4\%$ of its masked neurons), and \method{MeanAct.}
misses four ($35.2\%$); domain sensitivity shows the rankings are
robust to calibration corpus choice for the attribution cluster,
while the high cross-corpus stability of the non-attribution
baselines reflects their weak reliance on calibration content
rather than genuine robustness. Rank stability and inter-method agreement are not sufficient proxies;
LeRF/MoRF curves, or equivalent direct interventions, are what expose
the failures documented here.

\textbf{A protocol for behavior localization.}
The four-step protocol described in §\ref{sec:conclusion} (sufficiency, specificity, utility cost, cross-architecture replication) is motivated by the IG/LRP reversal
(Section~\ref{sec:cross}): a
method that wins on LLaMA may lose on Qwen3, and the reversal is
not noise but a function of where each method routes relevance
relative to where the model computes the target.

\textbf{Attribution and circuit discovery.}
Our method replaces the activation-patching inner loop of
mechanistic interpretability with attribution (one forward and
backward pass per prompt), trading some causal precision for speed.
The layer-matched audit (Section~\ref{sec:exp_b}) shows that the
approximation is precise enough to localize behavior-specific
circuitry at 70-cell scale, where full activation patching would
be prohibitive. The redundant-subspace finding
(Section~\ref{sec:redundant}) suggests circuit discovery and
attribution answer different questions: a circuit is a candidate
necessary mechanism, while an attribution-derived mask is one of
many sufficient ablations in a redundant subspace. Reporting both
gives different information.

\textbf{Consensus as error control.}
\consensus{} is not an attribution method in its own right; it is a
way of reducing the false-negative rate by requiring agreement
between two attribution paradigms with incompatible mechanisms. It
reduces false negatives at the LM level and wins on $3$ of $5$
LLaMA-3.1-8B refusal domains at the behavior level (hate, crime,
medical), with adult within $0.01$ of the IG-rescue cell and legal
the only clear loss, despite an LRP/IG
top-$1\%$ Jaccard of about $6\%$ (Table~\ref{tab:jaccard_lrp_ig}).
The agreed rows concentrate on the same MLP \texttt{gate\_proj} and
  \texttt{up\_proj} subspace identified in Section~\ref{sec:redundant}:
  \consensus{} is not finding new structure but confirming the
  gated-linear refusal subspace each
method already identifies. In the remaining $94\%$ of rows the two methods see
different partial views, and Borda aggregation combines them into a
more robust mask without forcing them to agree on row identities.

\end{document}